\documentclass[pdflatex,sn-apa,iicol]{sn-jnl}% Default with double column layout
\usepackage{microtype}
\usepackage{graphicx}

\usepackage{graphics} % for pdf, bitmapped graphics files
\usepackage{epsfig} % for postscript graphics files

\usepackage{booktabs}
\usepackage{multirow}
\usepackage{float}
\usepackage{adjustbox}
\usepackage{array}
\usepackage{tabulary}
\usepackage[table]{xcolor}

\usepackage{caption}

\usepackage{graphicx}%
\usepackage{amsmath,amssymb,amsfonts}%
\usepackage{amsthm}%
\usepackage{mathrsfs}%
\usepackage[title]{appendix}%
\usepackage{textcomp}%
\usepackage{manyfoot}%
\usepackage{algorithm}%
\usepackage{algorithmicx}%
\usepackage{algpseudocode}%
\usepackage{listings}%

\usepackage{minted}
\definecolor{LightGray}{gray}{0.9}
\definecolor{maroon}{rgb}{0.76, 0.13, 0.28}

\RequirePackage{xspace}
\makeatletter
\DeclareRobustCommand\onedot{\futurelet\@let@token\@onedot}
\def\@onedot{\ifx\@let@token.\else.\null\fi\xspace}

\def\eg{\emph{e.g}\onedot}

\def\etc{\emph{etc}\onedot}

\makeatother

\AtEndPreamble{
    \usepackage[capitalize]{cleveref}
    \crefname{section}{Sec.}{Secs.}
    \Crefname{section}{Section}{Sections}
    \crefname{table}{Tab.}{Tabs.}
    \Crefname{table}{Table}{Tables}
}

\usepackage{pifont}
\newcommand{\cmark}{\ding{51}}%
\newcommand{\xmark}{\ding{55}}%

\newcolumntype{H}{>{\setbox0=\hbox\bgroup}c<{\egroup}@{}}

\usepackage{anyfontsize}

\begin{document}

\title[Catch Me If You Can Describe Me: Open-Vocabulary Camouflaged Instance Segmentation with Diffusion]{Catch Me If You Can Describe Me: Open-Vocabulary Camouflaged Instance Segmentation with Diffusion}

%%=============================================================%%
%% Prefix	-> \pfx{Dr}
%% GivenName	-> \fnm{Joergen W.}
%% Particle	-> \spfx{van der} -> surname prefix
%% FamilyName	-> \sur{Ploeg}
%% Suffix	-> \sfx{IV}
%% NatureName	-> \tanm{Poet Laureate} -> Title after name
%% Degrees	-> \dgr{MSc, PhD}
%% \author*[1,2]{\pfx{Dr} \fnm{Joergen W.} \spfx{van der} \sur{Ploeg} \sfx{IV} \tanm{Poet Laureate} 
%%                 \dgr{MSc, PhD}}\email{iauthor@gmail.com}
%%=============================================================%%

\author*[1,2]{\fnm{Tuan-Anh} \sur{Vu}}\email{tavu@connect.ust.hk}
\author*[3]{\fnm{Duc Thanh} \sur{Nguyen}}\email{duc.nguyen@deakin.edu.au}
\author*[4]{\fnm{Qing} \sur{Guo}}\email{tsingqguo@ieee.org}
\author[2,5]{\fnm{Nhat} \sur{Chung}}\email{nhatcm@u.nus.edu} 
\author[6]{\fnm{Binh-Son} \sur{Hua}}\email{binhson.hua@tcd.ie}
\author[2]{\fnm{Ivor W.} \sur{Tsang}}\email{ivor\_tsang@a-star.edu.sg}
\author[1]{\fnm{Sai-Kit} \sur{Yeung}}\email{saikit@ust.hk}
% \equalcont{These authors contributed equally to this work.}

\affil[1]{\orgname{The Hong Kong University of Science and Technology}, \orgaddress{\country{Hong Kong}}}
% \affil[2]{\orgname{Centre for Frontier AI Research (CFAR) \& Institute of High Performance Computing (IHPC), A*STAR}, \orgaddress{\country{Singapore}}}
\affil[2]{\orgname{CFAR \& IHPC, A*STAR}, \orgaddress{\country{Singapore}}}
% \affil[3]{\orgname{Institute of High Performance Computing (IHPC), A*STAR}, \orgaddress{\country{Singapore}}}
\affil[3]{\orgname{Deakin University}, \orgaddress{\country{Australia}}}
\affil[4]{\orgname{Nankai University}, \orgaddress{\country{China}}}
\affil[5]{\orgname{National University of Singapore}, \orgaddress{\country{Singapore}}}
\affil[6]{\orgname{Trinity College Dublin}, \orgaddress{\country{Ireland}}}

%%==================================%%
%% sample for unstructured abstract %%
%%==================================%%

\abstract{Text-to-image diffusion techniques have shown exceptional capabilities in producing high-quality, dense visual predictions from open-vocabulary text. This indicates a strong correlation between visual and textual domains in open concepts and that diffusion-based text-to-image models can capture rich and diverse information for computer vision tasks. However, we found that those advantages do not hold for learning of features of camouflaged individuals because of the significant blending between their visual boundaries and their surroundings. In this paper, while leveraging the benefits of diffusion-based techniques and text-image models in open-vocabulary settings, we aim to address a challenging problem in computer vision: open-vocabulary camouflaged instance segmentation (OVCIS). Specifically, we propose a method built upon state-of-the-art diffusion empowered by open-vocabulary to learn multi-scale textual-visual features for camouflaged object representation learning. Such cross-domain representations are desirable in segmenting camouflaged objects where visual cues subtly distinguish the objects from the background, and in segmenting novel object classes which are not seen in training. To enable such powerful representations, we devise complementary modules to effectively fuse cross-domain features, and to engage relevant features towards respective foreground objects. We validate and compare our method with existing ones on several benchmark datasets of camouflaged and generic open-vocabulary instance segmentation. The experimental results confirm the advances of our method over existing ones. We believe that our proposed method would open a new avenue for handling camouflages such as computer vision-based surveillance systems, wildlife monitoring, and military reconnaissance. }


%%\pacs[JEL Classification]{D8, H51}

%%\pacs[MSC Classification]{35A01, 65L10, 65L12, 65L20, 65L70}

% Title/abstract in one-column mode
\onecolumn

\maketitle

% Now switch to two-column mode
\twocolumn

\section{Introduction}
\label{sec:intro}

\begin{figure*}[!t]
    \centering
    \includegraphics[width=0.99\linewidth]{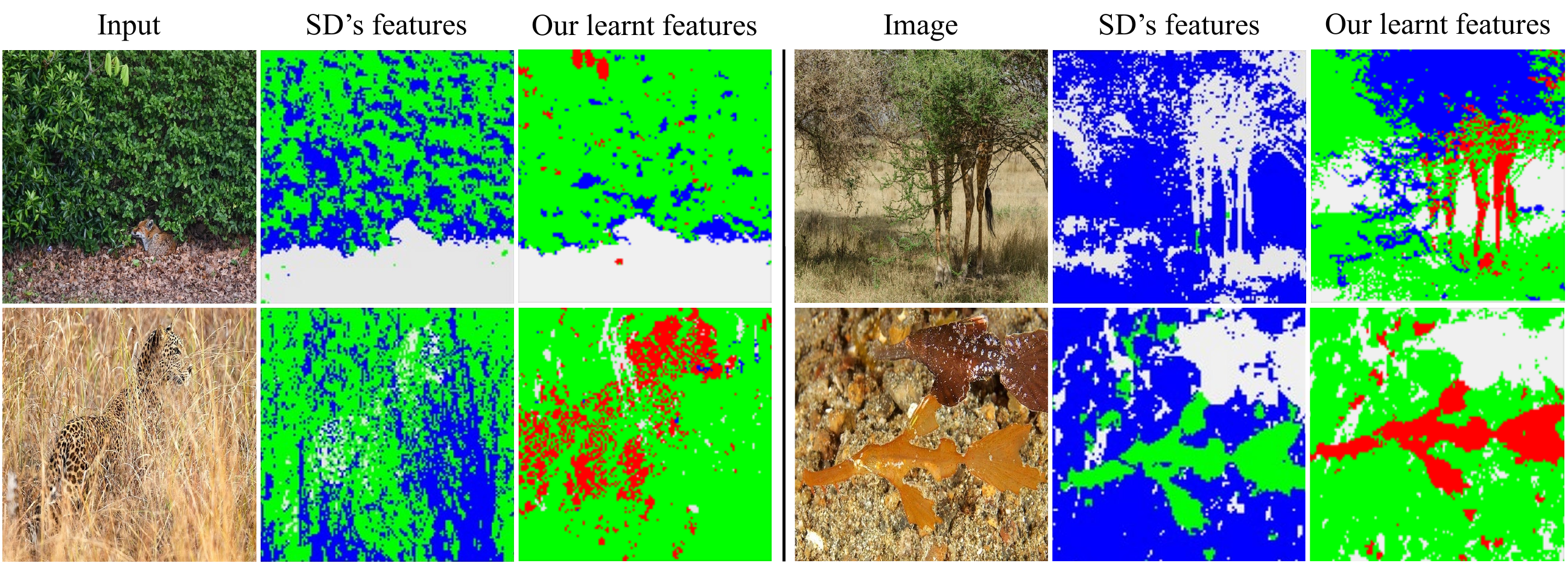}
    \vspace{2mm}
    \caption{Illustration of textual-visual features of off-the-shelf Stable Diffusion when dealing with CIS and our learnt features. Given an input image, textual-visual features are extracted and clustered using a $K$-means clustering algorithm ($K=4$). As shown, camouflaged animals can be localised based on the clustering results. We leverage these rich features to perform instance segmentation of camouflaged objects. This figure is best viewed in colour.}
    \label{fig:teaser}
\end{figure*}

Camouflage is a powerful biological mechanism for avoiding detection and identification. In nature, camouflaged tactics are employed to deceive the sensory and cognitive processes of both prey and predators. Wild animals utilise these tactics in various ways, ranging from blending themselves into the surrounding environment to employing disruptive patterns and colouration~\citep{Nguyen_MTAP_2023}. Thus, identifying camouflages is pivotal in many wildlife surveillance applications~\citep{Fleming2014CameraTW,9371667}, as it helps locate hidden individuals for monitoring and conservation.

In fact, localisation of camouflaged objects~\citep{Fan_CVPR_2020,He_CVPR_2023}, such as Camouflaged Object Detection (COD) and Camouflaged Instance Segmentation (CIS), has been an important research topic in computer vision, whose main challenge lies in the need to learn discriminative features that for discerning camouflaged target objects from their surroundings. Existing COD techniques can be utilised to roughly identify camouflaged objects at regional scales (\eg, bounding boxes), but they are not designed to distinguish individual instances at finer scales like pixel level. CIS, on the other hand, operates under the assumption that individual instances' features closely resemble one another and aims to provide class-independent segmentation masks~\citep{pei2022osformer}. However, the diversity of camouflages within a single scene can lead to complex intertwining patterns, making the CIS task more challenging in severe environmental conditions, \eg, terrestrial and aquatic environments, under poor imaging quality, \eg, occlusions, image blur, and low-light conditions in underwater applications. These challenges also hinder the collection and annotation of high-quality data for training and testing CIS algorithms.

Meanwhile, while humans can recognise an unlimited number of target categories, and open-vocabulary recognition has been developed to mimic human intelligence with unbounded understanding, current endeavours focus only on generic objects and individuals~\citep{OV-DETR,OpenSeg,du2022learning,gao2022open,minderer2022simple,kuo2023f, xu2023odise}. For example, while~\citep{xu2023odise} suggested that Internet-scale text-to-image diffusion models can be utilised to create a state-of-the-art open-vocabulary segmenter for many concepts, our investigations show that they demonstrate inconsistent segmentation results when working with camouflages, as indicated by their pixel-wise embeddings in~\Cref{fig:teaser}. Although pretrained generative features offer strong potential for open-vocabulary generalization, our findings highlight their limitations in capturing fine-grained visual ambiguities such as camouflage. Notably, existing open-vocabulary segmentation methods~\citep{MaskCLIP, xu2023masqclip, xu2023odise, xDecoder, zou2023segment, OpenSeeD} share this limitation, as camouflage detection is not central to their design. 

To overcome the aforementioned hurdles, we propose a method that leverages text-to-image diffusion to address the problem of OVCIS. Our method is inspired by the advanced object representation learning capabilities of diffusion techniques and the language-vision transferability of text-image models. Text-to-image diffusion models, \eg, the stable diffusion model by~\citep{rombach2022high}, are designed to learn essential object features in the presence of noise, making them useful for extracting features relevant to target objects in a noisy and cluttered background. While we observed that features learnt solely from the visual domain are weak to distinguish camouflaged objects from their surroundings, the features learnt by text-image discriminative models, \eg, CLIP~\citep{CLIP}, contain rich information about the real world thanks to the variety of concepts in open-vocabulary training data~\citep{OVSurvey}. We hypothesize that an effective combination of features learnt from both the textual and visual domains would benefit the representation learning of camouflaged objects. We illustrate the effectiveness of textual-visual representations for CIS in~\Cref{fig:teaser}. To the best of our knowledge, such a cross-domain combination with open-vocabulary for CIS is \textit{novel}, and ours is the \textit{first framework} to localise camouflaged object instances at this scale.

To effectively learn textual-visual representations of camouflaged objects, our method assimilates an input image and a text prompt about objects included in the input image, so the input image and its implicit caption (generated by a captioner) are integrated into a text-to-image diffusion model to extract visual features. These features are processed at multiple scales and fused into a visual feature map, which is then used to generate object masks. Simultaneously, textual features are extracted from the text prompt using a text encoder. These textual features are enriched from open-vocabulary category labels and proven to improve the discriminative power of camouflaged objects' representations against the background. Our proposed pipeline aggregates textual and visual features in a mask-out manner to recognise the masks of the target objects. The diffusion model utilises a cross-attention mechanism to link textual features with visual features and condition the feature learning process. Hence, the learnt features are likely to be distinct and connected to high/mid-level semantic notions that may be expressed in the language part. While our method somewhat shares a similar approach with the works by~\citep{xu2023odise,vpd} at a high-level perspective, our pipeline is more specialised to CIS by designing camouflage-specialised modules.

\textbf{COD vs.\ CIS vs.\ OVCIS.}
Camouflaged Object Detection (COD) aims to separate camouflaged regions from background and typically produces a \emph{binary} camouflage mask without requiring instance separation.
Camouflaged Instance Segmentation (CIS) extends COD by requiring \emph{instance-level} separation of multiple camouflaged objects, but prior CIS formulations are often class-agnostic or focus primarily on instance delineation rather than open-vocabulary semantic generalization~\citep{pei2022osformer}.
In contrast, Open-Vocabulary Camouflaged Instance Segmentation (OVCIS) requires \emph{both} (i) robust instance separation under camouflage and (ii) \emph{open-vocabulary} category assignment at inference via textual category prompts, where training categories $\mathcal{C}_{\text{train}}$ and test categories $\mathcal{C}_{\text{test}}$ may be disjoint.
While open-vocabulary segmentation has been explored in general-domain settings~\citep{MaskCLIP, xu2023masqclip, xu2023odise, xDecoder, zou2023segment, OpenSeeD}, existing methods are not designed to address the boundary ambiguity and low-contrast appearance intrinsic to camouflage.
Accordingly, OVCIS lies at the intersection of camouflage understanding, instance segmentation, and open-vocabulary recognition (\Cref{tab:task_taxonomy}).

% -------------------------------
% Task taxonomy table (COD vs CIS vs OVCIS)
% -------------------------------
\begin{table*}[t]
\centering
\caption{Conceptual distinctions among Camouflaged Object Detection (COD), Camouflaged Instance Segmentation (CIS), and Open-Vocabulary Camouflaged Instance Segmentation (OVCIS). OVCIS combines camouflaged instance separation with open-vocabulary category assignment at inference.}
\label{tab:task_taxonomy}
\setlength{\tabcolsep}{6pt}
\renewcommand{\arraystretch}{1.4}
\resizebox{0.99\textwidth}{!}{%
% \begin{tabular}{lcccccc}
\begin{tabular}{p{1.2cm} >{\centering\arraybackslash}p{3.7cm} >{\centering\arraybackslash}p{1.9cm} >{\centering\arraybackslash}p{3.3cm} >{\centering\arraybackslash}p{3.5cm}}
\toprule
\textbf{Task} &
\textbf{Primary output} &
\textbf{Instance separation} &
\textbf{Vocabulary regime} &
\textbf{Typical supervision} \\
\midrule \midrule
\textbf{COD} &
Binary mask (camouflage vs.\ background) &
\texttimes &
N/A &
Binary mask annotation \\
\midrule
\textbf{CIS} &
Instance masks (often class-agnostic) &
\checkmark &
Closed or class-agnostic &
Instance masks (w/ or w/o categories)\\
\midrule
\textbf{OVCIS (Ours)} &
Instance masks \& open-vocab category labels &
\checkmark &
Open-vocabulary ($\mathcal{C}_{\text{train}} \rightarrow \mathcal{C}_{\text{test}}$) &
Instance masks + category names/text \\
\bottomrule
\end{tabular}
}
\end{table*}

In summary, we make the following contributions to our work:
\begin{itemize}
    \item We address a new and challenging task: open-vocabulary camouflaged instance segmentation (OVCIS), which would enhance the capability of many critical applications such as computer vision-based surveillance systems, wildlife monitoring, and military reconnaissance. 
    
    \item We propose a method for OVCIS built upon text-to-image diffusion and text-image transfer techniques, advanced with open-vocabulary utilisation.
    
    \item We propose an object representation learning paradigm specialised for camouflages. Our camouflage-specialised components include a Multi-scale Features Fusion (MSFF) module to encapsulate visual features from diffusion, a Textual-Visual Aggregation (TVA) module to utilise textual information that pronounces visual features, 

    and a Camouflaged Instance Normalisation (CIN) module to adaptively capture textual-visual information that enhances camouflaged object representations.
    
    \item We conduct extensive experiments and ablation studies that demonstrate the advantages of our method over existing works.
\end{itemize}

%%===========================================================================================%%

\section{Related Work}
\label{sec:related}

We start our review of related work with an overview of deep learning-based advances for camouflaged object understanding. Following it, we delve into contemporary research in text-to-image diffusion, thereby discussing their role in facilitating open-vocabulary computer vision tasks. Then, we review prior research on generative models and their applications to visual segmentation. 

\subsection{Camouflaged Object Understanding}

The main aim of camouflaged object understanding lies in learning object representations that are difficult to dissimilate from their background. Existing research has attempted to address various tasks in camouflaged object understanding from images. For instance, \citep{sun2023ioc} counted objects that blended seamlessly into backgrounds. Following closely, ~\citep{yunqiu_cod21} identified salient image regions of hidden objects that align with the nuances of human perception. COD was studied by~\citep{He_CVPR_2023}, in which the authors decomposed learnt features into different frequency bands using learnable wavelets to identify the most informative features to differentiate target objects and backgrounds. In addition, an auxiliary reconstruction network was built to boost up further the discriminative power of the foreground's features against the background's ones. In the work by~\citep{fan2022concealed}, a method for segmenting camouflaged objects was proposed to segment obscured objects without pinpointing specific categories for the objects. 

CIS was brought forth by~\citep{pei2022osformer} to emphasise the learning of object-vs-background-discriminative representations, which is different from general instance segmentation~\citep{xie2021trans} that aims to maximise inter-object distances. 
Although this goal is common in existing camouflaged object understanding methods and various attempts have been made to address it in the literature, learning such representations from solely imagery data is challenging as it is the nature of visual camouflages. Our research differs from existing ones by exploring the potential of diffusion-based representations and textual data as additional cues to drive the open-vocabulary learning of CIS, thereby utilising them to make camouflaged object representations adaptive to camouflages that are never seen in training. 

Thanks to the variety of concepts, textual features learnt from text prompts about objects included in an input image can help to find visual features relevant to the objects. In addition, an effective combination of both textual and visual features would further enhance the robustness of camouflaged object representations, where visual features solely are not robust enough to distinguish camouflaged objects from their surroundings. To the best of our knowledge, our study is the first of such work.

\subsection{Text-to-Image Diffusion}

Significant progress has been made in Artificial Intelligence (AI)-empowered picture creation with recent advances in large-scale text-to-image diffusion models, including Stable Diffusion~\citep{rombach2022high}, DALL-E 2~\citep{ramesh2022hierarchical}, and Imagen~\citep{saharia2022photorealistic}. These models have demonstrated photo-realistic quality image generation by being trained on text-image datasets of substantial scale sourced from the Internet. They also have shown the ability to be conditioned on unrestricted text prompts in order to produce visuals that closely resemble real-life photographs. 

The application of text-to-image diffusion models has facilitated the creation and manipulation of visual contents in an ever-easy and convenient manner via language-based interactions (\eg, text prompts). This has enabled a wide spectrum of applications such as content-personalised customisation~\citep{kumari2023multi}, zero-shot translation~\citep{parmar2023zero}, content editing~\citep{hertz2023prompt}, and image generation~\citep{gal2023designing}. 

In this paper, we do not apply text-to-image diffusion technique to image creation and/or image manipulation. Instead, we explore its capability of cross-domain feature learning. 
Most related to our work, ~\citep{xu2023odise} showed that pre-trained representations in diffusion models can be utilised for open-vocabulary segmentation. However, we found that their method performs poorly and inconsistently on camouflaged datasets, due to a lack of ability to identify object boundaries in camouflages. To address this limitation, we devise a feature fusion strategy based on a state-of-the-art text-to-image diffusion architecture to fuse image features with implicit caption features at multiple scales. Our experiments show that such a fusion facilitates the learning of object-vs-background discriminative features, which are crucial for CIS.

\subsection{Generative Models for Segmentation}

Many studies are related to our work in terms of applying image generative models, such as Generative Adversarial Networks (GANs)~\citep{esser2021taming, karras2020styleganv2} or diffusion models~\citep{ho2020denoising, song2021denoising, dhariwal2021diffusion}, to semantic segmentation~\citep{li2022bigdatasetgan,baranchuk2022ddpmseg, tritrong2022repurposing}. For GANs, a straightforward approach is to synthesise images and their corresponding semantic maps to train a segmentation network~\citep{li2022bigdatasetgan}. ~\citep{tritrong2022repurposing}, segmentation is performed by training a generative model on datasets with a limited vocabulary.
For example, ~\citep{baranchuk2022ddpmseg} proposed a diffusion-based framework, named DDPMSeg, based on the denoising diffusion probabilistic model (DDPM)~\citep{ho2020denoising} to learn a feature map for an input image. The feature map was then passed to a pixel classifier to perform semantic or part segmentation. A few hand-annotated examples per category are then utilised to classify learnt representations into semantic regions.
Similarly, \cite{xu2023odise} showed that pre-trained representations in diffusion models can be utilised for open-vocabulary segmentation in the wild. Their insights suggest that internal representations learnt by diffusion models can well correlate with high- and mid-level semantic concepts that can be described in language, addressing the lack of spatial and relational understanding in traditional open-vocabulary segmentation. 
Therefore, their approach introduces a new capacity for generative models, \eg, image generation-driven representation learning. 
However, while promising as a practical tool, we found that diffusion-based pre-trained representations are not designed to tackle camouflaging effects, even though the intermediate representations of a generative model can be trained to capture high-level semantic concepts (\eg, the presence of an object in an input image) under specific feature constraints.

\subsection{Open-Vocabulary Detection and Segmentation}

Numerous studies have been proposed to incorporate vision-language models (VLMs) into open-vocabulary detection and segmentation~\citep{zhong2022regionclip,OV-DETR,OpenSeg,du2022learning,gao2022open,minderer2022simple,rasheed2022bridging,kuo2023f}. 
This has enabled the detection and classification of novel objects from a vast conceptual domain with the help of pre-trained VLMs~\citep{zhang2023vision,OVSurvey}. 
OVR-CNN was the first open-vocabulary object detection introduced by~\citep{zareian2021open}, which underwent pre-training with image-caption data in order to learn and identify unknown objects, followed by fine-tuning for zero-shot detection. 

Following recent advances in VLMs~\citep{CLIP,ALIGN}, ViLD~\citep{ViLD} pioneered the incorporation of extensive representations of pre-trained CLIP~\citep{CLIP} into an object detector, and many works~\citep{du2022learning, kuo2023f, detic} have followed the similar framework.
~\citep{du2022learning} proposed DetPro, a sophisticated automated prompt learning method, to learn the presence of an object in a background via prompt training. F-VLM~\citep{kuo2023f} adopted a frozen VLM to generate new object categories based on cropped CLIP features. ~\citep{detic} extended the ability of the well-known object detector, Faster R-CNN~\citep{FasterRCNN} to newly introduced object categories by replacing the classification weights (in the classification head) by fixed language embeddings learnt from open-vocabulary. 

Despite the successes achieved, existing methods have limited capabilities against camouflaged objects due to the utilisation of small closed vocabularies and/or the incorporation of VLMs for generic object classes, which are often distinguishable from the background. 
It is because the pre-trained representations learnt on general object classes are not designed for discerning object boundaries between camouflaged individuals~\citep{MaskCLIP, xu2023masqclip, xu2023odise, xDecoder, zou2023segment, OpenSeeD}.
While exploiting insights and advantages from prior studies, our work stands out in a specifically focused direction: tackling the challenge of open-vocabulary instance segmentation of camouflaged targets, yet without losing much representation localisation capability on general objects.
Our proposed method extend towards segmentation of novel object categories with concealed appearances in natural environments using an open-vocabulary set.

%%===========================================================================================%%

\section{Proposed Method}
\label{sec:method}

\subsection{Problem Definition} 

We aim to build and train an instance segmentation model with a set of pre-defined object categories, referred to as $\mathbf{C}_{\text{train}}$. The instance segmentation model can work on a new domain with $\mathbf{C}_{\text{test}}$ object categories, where $\mathbf{C}_{\text{test}}$ and $\mathbf{C}_{\text{train}}$ may or may not share common object categories. In other words, $\mathbf{C}_{\text{test}}$ may include object categories previously unseen during the training of the instance segmentation model. 

Throughout the training process, it is presumed that binary mask annotations for target objects in each training image are available. Moreover, each mask is either associated with a category name or a caption presented in the text form. During the testing phase, however, neither the category label nor the caption is accessible for any test image. Note that, only the names of the test categories in $\mathbf{C}_{\text{test}}$ are provided.

\subsection{Overview}

\subsubsection{Preliminaries}

We build our method upon two technical advances: text-to-image diffusion and text-image transfer. We first briefly summarise those techniques and then describe how they can be applied to our method. 

\vspace{2mm}
\noindent\textbf{Text-to-Image Diffusion} facilitates the creation of high-quality images guided by text prompts. A text-to-image diffusion model is trained on a massive corpus of image-text pairs amassed through web crawling, as indicated in the literature~\citep{nichol22glide,saharia2022photorealistic,xu2023odise}. Text inputs are encoded into embeddings using an established text encoder, e.g., T5~\citep{colin2022t5}. An image is initially perturbed by introducing Gaussian noise at a controlled intensity before being fed into the diffusion network. The network is fine-tuned to reverse the noise application, utilising noisy images and associated text embeddings to diminish the distortion. In the inference phase, the model synthesises an image from inputs, including pure Gaussian noise shaped to the image's dimensions and a user-provided description's text embedding. Through successive inference iterations, the model iteratively denoises the input and finally results in a photo-realistic image of the user-provided text description. 

In our work, we adopt the Stable Diffusion (SD) model developed by~\citep{rombach2022high} and pre-trained on the LAION-5B dataset~\citep{schuhmann2022laion}. SD is chosen for two reasons. First, SD is well known for its ability in effective fusion of textual and visual information, which we found useful for camouflaged instance segmentation where visual features only can be indistinguishable. Second, thanks to the denoising process, SD is able to manage noisy and subtle visual distinctions effectively, making them particularly suitable for camouflage segmentation where visual boundaries blend significantly with the background. 

The SD model is composed of a trio of elements: \ding{172} a captioner (realised by a pre-trained text encoder) that generates a text embedding (implicit caption) for an input image; \ding{173} a pre-trained variational auto-encoder for learning of image representations; and \ding{174} a denoising time-conditional U-Net~$\epsilon_{\theta}(\cdot)$, which applies progressive convolution operations to downsample and upsample feature maps of an input image with skip connections. Within the U-Net, textual-visual interactions are enabled by cross-attention. In detail, the captioner projects a text input $y$ into an embedding, which is then transformed into \texttt{Key} and \texttt{Value} pairs. At the same time, a feature map of a noisy image undergoes a linear projection to form a \texttt{Query}. This design allows for iterative updates of input images conditioned on accompanying text descriptions.

\begin{figure*}[t]
    \centering
    \includegraphics[width=0.99\linewidth]{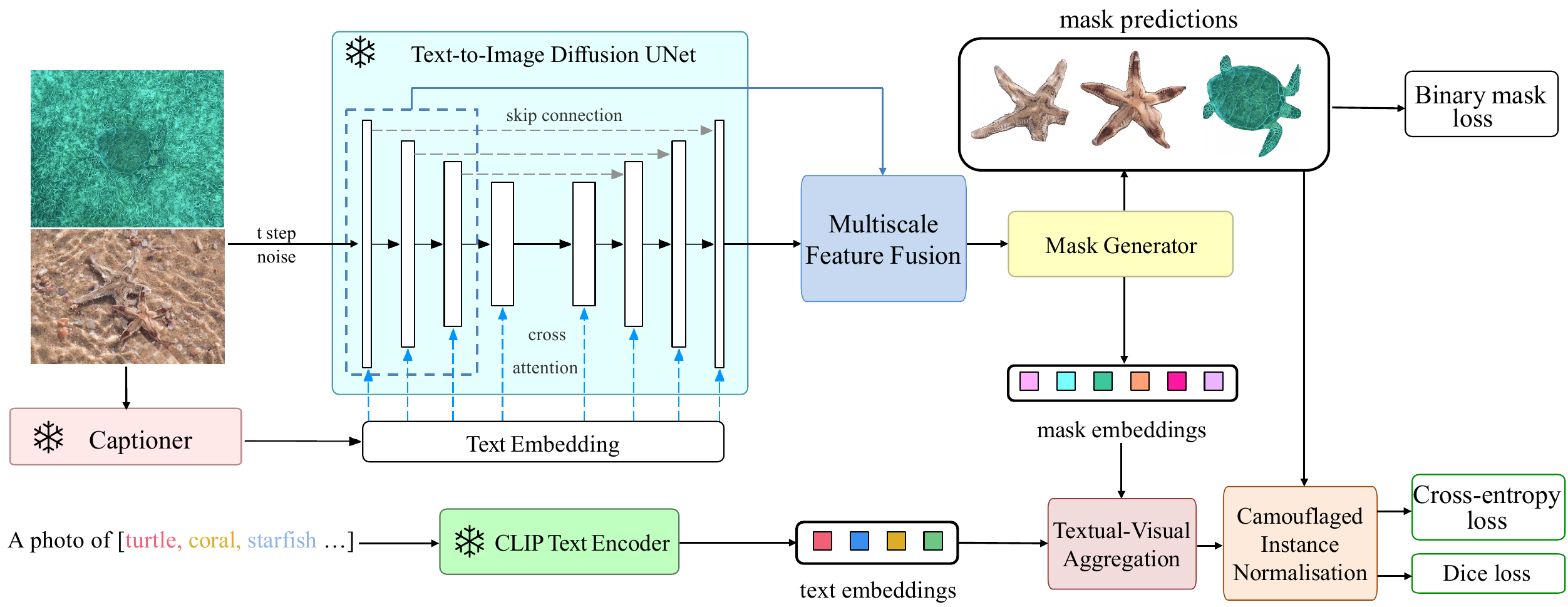}
    \vspace{2mm}
    \caption{Pipeline of our proposed method for Open-vocabulary Camouflaged Instance Segmentation (OVCIS). Inputs include an image and a text prompt about target objects in the input image. Outputs include instance masks of the target objects. The target objects can be novel and have never been seen in the training data. We leverage state-of-the-art text-to-image diffusion and text-image transfer techniques to learn textual-visual features that facilitate the object representation learning for segmenting camouflaged objects. }
    \label{fig:camo_pipeline}
    %\vspace{-0.3cm}
\end{figure*}

The training process of the SD model is outlined as follows. For a given pair $(\mathcal{I}, y)$ in a training dataset, the image $\mathcal{I}$ is encoded into a latent representation $z$ and then subjected to noise, resulting in a noised vector $z^t:= \alpha^t z + \sigma^t \epsilon$, where $\epsilon \sim \mathcal{N}(0,1)$ is a noise variable, and $\alpha^t, \sigma^t$ are parameters that manage the noise level and the fidelity of each sample. The training aims to fine-tune the time-conditional U-Net~$\epsilon_{\theta}(\cdot)$ to anticipate the noise vector $\epsilon$ and to accurately reconstruct the initial latent vector $z$, while being conditioned on the text input $y$. The fine-tuning is performed by using a loss function that minimises the mean squared error of noise prediction as follows:
\begin{equation}
    \mathcal{L}_{\text{diffusion}} = \mathbb{E}_{z, \epsilon \sim \mathcal{N}(0,1),t,y} \left[||\epsilon - \epsilon_{\theta}(z^{t},t,y)||_{2}^{2} \right]
\end{equation}
\noindent where the time variable $t$ is randomly selected from the set $\{1,\dots,T\}$.

During the inference phase, the SD model synthesises an image by sequentially refining a latent vector $z^T \sim \mathcal{N}(0,I)$, with the process being contingent on a text input $y$. Specifically, for each time step $t = 1,\dots, T$ of the denoising sequence, $z^{t-1}$ is derived from the current $z^t$ and the U-Net's noise prediction, which in turn takes $z^t$ and the text prompt $y$ as inputs. After the final denoising stage, the latent vector $z^0$ is transformed back to produce a final output image $\mathcal{I'}$.

\vspace{2mm}
\noindent\textbf{Text-Image Transfer} originally aims to learn directly from raw text about images. This technique leverages rich textual representations learnt from the textual domain to scale up representation learning in the visual domain. As shown in the literature, natural language can be used to supervise a wide set of visual concepts through its generality~\citep{Sariyildiz_ECCV_2020,Desai_CVPR_2021,Zhang_PMLR_2022}. Recently, CLIP proposed by~\citep{CLIP} offers text-image transferibility in both directions, i.e., text-to-image and image-to-text. 

In our work, we adopt a CLIP model, pre-trained on 400 million image-text pairs crawled from the Internet. This model is used to generate text embeddings for implicit captions of input images and text embeddings for text prompts associated with input images. Due to learning from large-scale and diverse training data, we observed that these text embeddings greatly aid in improving camouflaged objects' representation.

\subsubsection{Our Pipeline}

\Cref{fig:camo_pipeline} illustrates the pipeline of our method. At an abstract level, our method takes an image and a text prompt about target objects as inputs and produces instance masks with object categories for the target objects as outputs. 

The input image is first passed to the SD model~\citep{rombach2022high}, which is pre-trained and frozen (no training), to extract latent features. The input image is also fed to the pre-trained and frozen CLIP model~\citep{CLIP} to calculate its implicit caption embedding. The caption embedding is inserted into the SD model at various scales (layers) and fused with the SD model's last layer to form image-guided features. We call these features ``image-guided features'' though they somewhat include textual information. This is because the input image drives the textual features from the implicit caption embedding. The image-guided features, coupled with annotated training masks, serve as inputs to train a mask generator capable of producing instance masks for all potential categories within the input image. The instance masks are then used to locate object-relevant features in a mask-out manner. This step results in mask embeddings (i.e., features extracted within masked regions).

The input text prompt is concurrently processed by the CLIP~\citep{CLIP}, independently of the input image, and its corresponding text embeddings are calculated. These text embeddings are transferable to visual features yet extracted from the textual input, hence considered as ``text-guided features''. The text embeddings (text-guided features) and mask embeddings (image-guided features) are aggregated by a textual-visual aggregation module, which aims to emphasise the learnt features towards foreground objects defined in the input text prompt. This module results in a textual-visual representation for the input image and text prompt. 

Next, the textual-visual representation is normalised regarding the instance masks segmented by the mask generator and finally classified by a mask classifier into object categories. 

The entire pipeline is trained with object categories in $\mathbf{C}_{\text{train}}$. Note that, since the SD and CLIP models have been pre-trained and frozen, the training of the entire pipeline is equivalent to learning of parameters in modules specialised for CIS (multi-scale feature fusion, mask generator, textual-visual aggregation, camouflaged instance normalisation). Once the training is completed, the inference process carries out open-vocabulary instance segmentation, i.e., the pipeline can perform instance segmentation of object categories in $\mathbf{C}_{\text{test}}$. 

To make our pipeline specialised to CIS, we develop several technical components to facilitate camouflaged object representation learning (see~\Cref{sec:camouflage_object_representation}) and camouflaged instance normalisation (see~\Cref{sec:camourlaged_instance_normalisation}). 

\subsection{Camouflaged Object Representation Learning}
\label{sec:camouflage_object_representation}

Given the features learnt by the SD model from the input image and the text embeddings produced by the CLIP from the input text prompt, we perform camouflaged object representation learning via three modules: multi-scale feature fusion, mask generator, and textual-visual aggregation. These modules are described below.

\begin{figure}[t]
    \centering
    \includegraphics[width=0.99\linewidth]{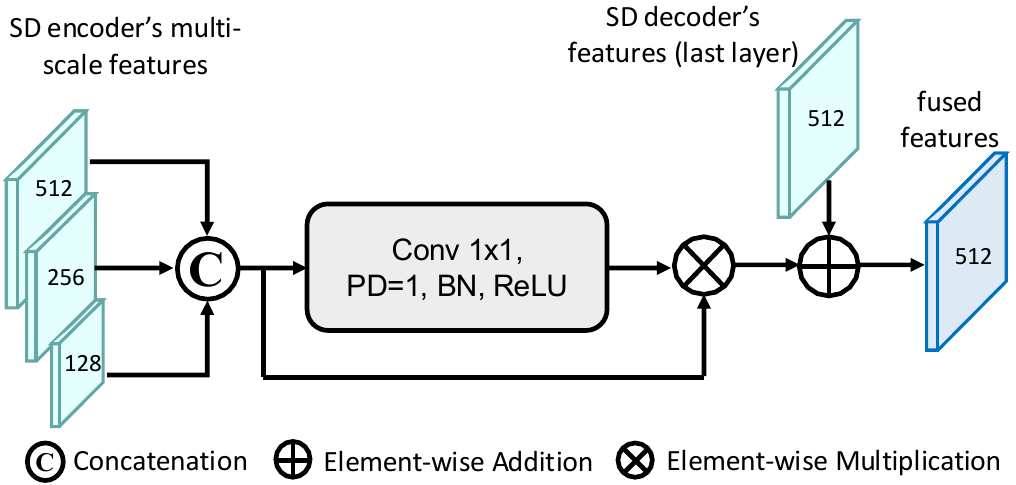}
    \vspace{1mm}
    \caption{Architecture of the multi-scale features fusion (MSFF) module.}
    \label{fig:msff}
\end{figure}

\subsubsection{Multi-scale Features Fusion} 

The MSFF module fuses the multi-scale features from the encoder part of the SD model and the features from the last layer of the decoder part of the SD model. We present the architecture of the MSFF module in~\Cref{fig:msff}.

The fusion process concatenates multi-scale SD encoder features and applies the $1\times1$ convolution on the concatenated features. The resulting features are then combined with the concatenated features via element-wise multiplication, and the modulated output is added to the SD decoder’s final-layer features. 

\begin{figure}
    \centering
    \includegraphics[width=0.9\linewidth]{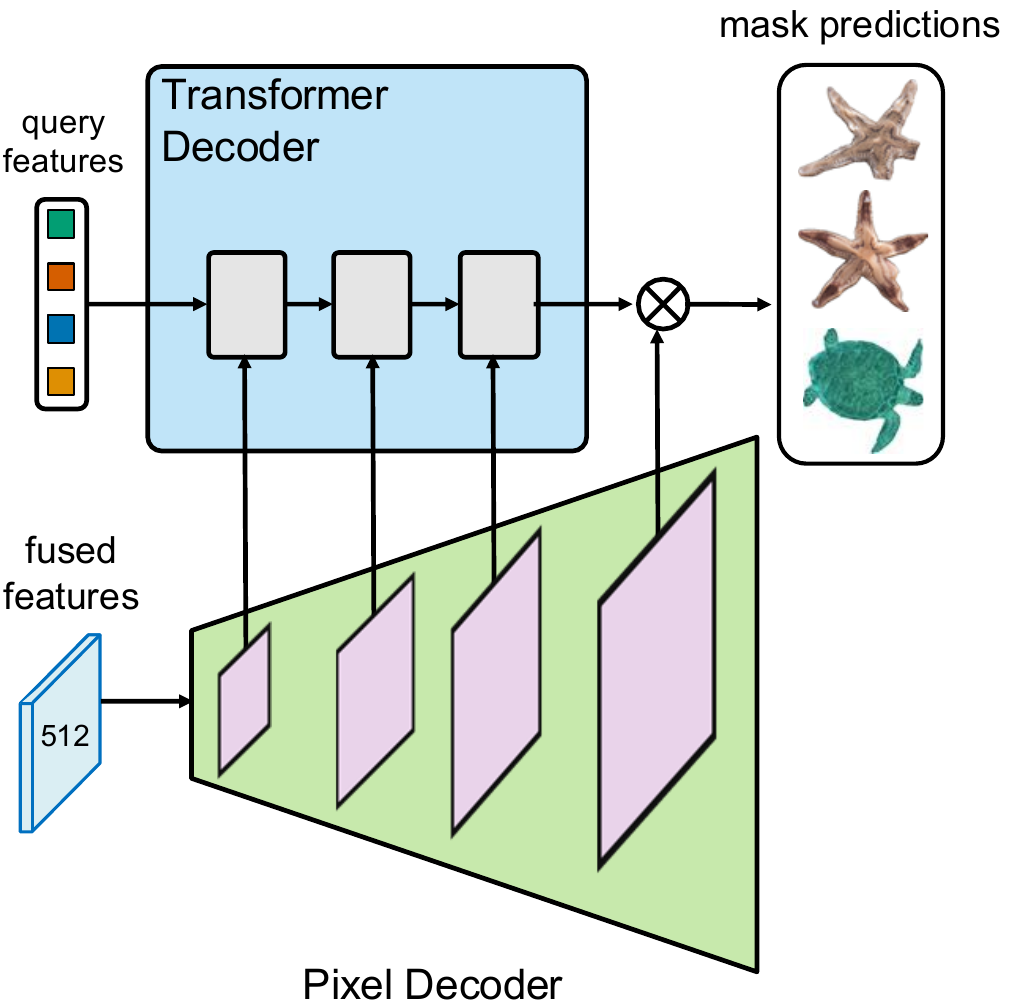}
    \vspace{1mm}
    \caption{Architecture of the mask generator.}
    \label{fig:decoder}
\end{figure}

\subsubsection{Mask Generator}
\label{sec:mask_generator}

We adopt the decoder in the mask-attention Transformer, the core component in the Mask2Former architecture~\citep{mask2former}, to realise our mask generator. The mask generator receives input as a fused feature vector from the MSFF module and produces outputs including $N$ class-agnostic binary masks $\{m^{pred}_i\}_{i=1}^N$ and their corresponding $N$ mask embedding features $\{z^{pred}_i\}_{i=1}^N$ for all possible objects in the input image. We illustrate the mask generator in~\Cref{fig:decoder}. 

The mask generator employs a pixel decoder that progressively increases the resolution of the fused features from the MSFF module and generates per-pixel high-resolution embeddings. This pixel decoder is designed meticulously, using multiple layers to capture fine-grained and broad contextual information. Following that, a Transformer's decoder processes the intermediate feature maps in the pixel encoder to handle object queries, which are initialised randomly but then learnt through training. To effectively process the intermediate feature maps in the pixel decoder, the mask generator guides each feature map at a scale to an individual layer in the Transformer's decoder. Consequently, each layer in the Transformer's decoder focuses on a feature map at a specific scale in the range of $\{1/32, 1/16, 1/8\}$. We observed that this strategy significantly enhances the ability of the mask generator to adeptly handle objects in various sizes. 

\begin{figure}
    \centering
    \includegraphics[width=0.99\linewidth]{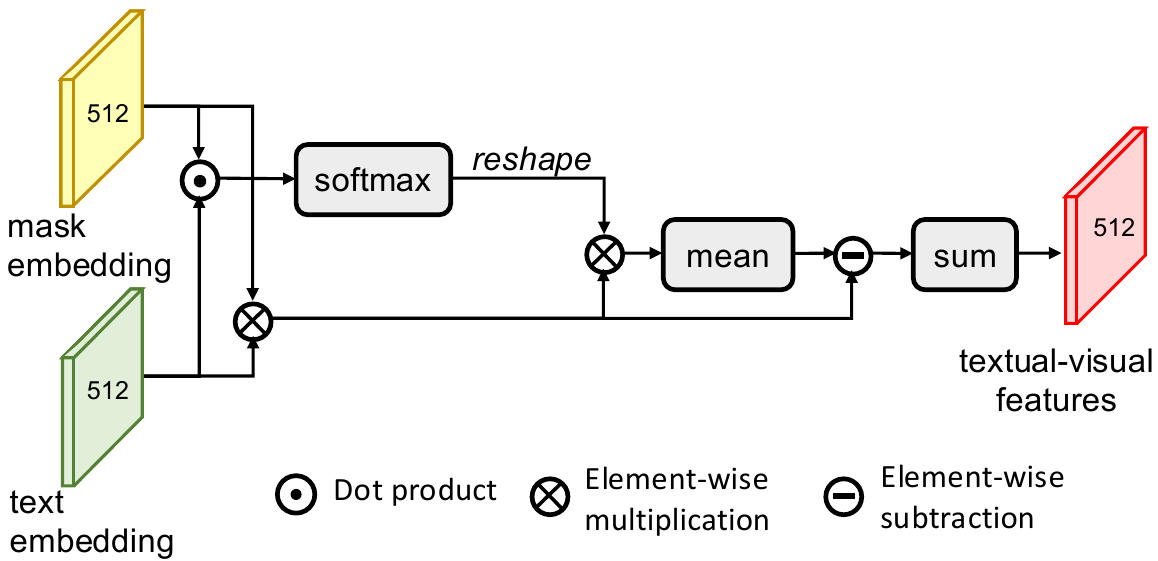}
    \vspace{1mm}
    \caption{Architecture of the textual-visual aggregation (TVA) module.}
    \label{fig:tva}
\end{figure}

\subsubsection{Textual-Visual Aggregation} 

The TVA module is designed to highlight object-relevant features to drive the object representation learning towards foreground objects, whose architecture is shown in~\Cref{fig:tva}. We later show that experimental results validated its effectiveness.

The TVA module in our proposed pipeline operates as follows. Like the Mask R-CNN~\citep{maskrcnn}, we crop corresponding features from the MSFF module and perform mask pooling for each object mask returned by the mask generator. This step results in mask embeddings (i.e., embeddings are determined by masks). We then compute the interactions between these mask embeddings and the text embeddings produced by the CLIP. Nevertheless, instead of directly using a dot product to calculate the interaction between two embeddings as in CLIP~\citep{CLIP}, we apply a softmax operator to the dot product of the embeddings to weight features, then apply mean-normalisation to remove irrelevant features before aggregating them by a channel-wise summation. Removing irrelevant features helps to mitigate the problem of noisy activations, making the learning process lean towards features relevant to the object categories specified in the input text prompt. 

\Cref{fig:teaser} visualises learnt textual features by our method on several challenging cases. As shown, the learnt textual-visual features on camouflaged objects can be well identified and located, although the objects blend into cluttered backgrounds. This is evident in the ability of our method to learn distinguished object-vs-background features.

\subsection{Camouflaged Instance Normalisation}
\label{sec:camourlaged_instance_normalisation}

\begin{figure}
    \centering
    \includegraphics[width=0.99\linewidth]{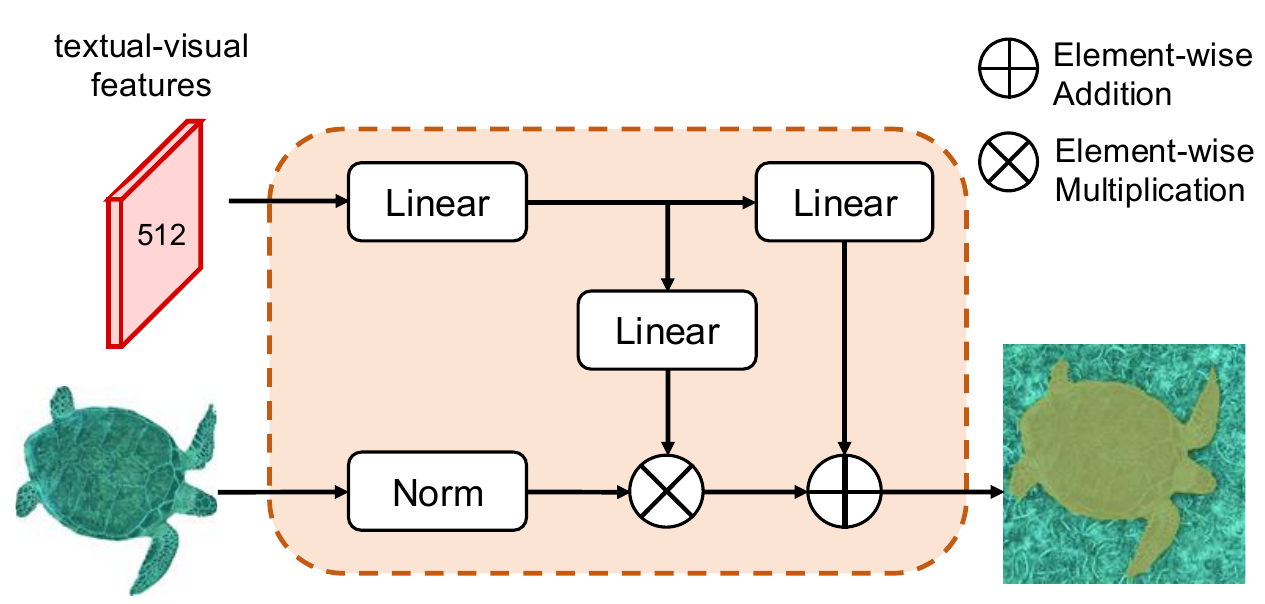}
    \vspace{1mm}
    \caption{Architecture of the camouflaged instance normalisation (CIN) module.}
    \label{fig:cin}
\end{figure}

Inspired by the adaptive instance selection network~\citep{huang2017adain,pei2022osformer}, we develop a CIN module to achieve final masks for the target objects. We present the architecture of the CIN module in~\Cref{fig:cin}.

The CIN module takes inputs as a textual-visual feature map from the TVA module and an object mask from the mask generator. A linear layer first projects the textual-visual feature map into a higher-dimensional space. Next, affine weights and biases are attained by applying two subsequent linear layers to the result of the first linear layer. The affine weights and biases are then combined, together with the input mask from the mask generator, to predict a final instance mask for the object specified in the input mask. Since the CIS task is category-agnostic, we use a confidence score for the existence of a camouflaged object at each location, rather than a classification score in generic instance segmentation. 

\subsection{Training}

We train the entire pipeline of our method by optimising the loss functions (binary mask, cross-entropy, dice losses) used in the mask generator and the CIN module in a supervised fashion.

Specifically, we adopt a binary cross-entropy loss as our binary mask loss $\mathcal{L}_{bce}$ and a dice loss $\mathcal{L}_{dice}$ ~\citep{dice2016} for supervising binary mask predictions in the mask generator. The dice loss is used to remedy class imbalance. 

We carry out the training of the CIN module using the conventional close-vocabulary training approach. Suppose that we can access to the ground-truth category label for each object mask during the training phase. For each mask embedding $z^{pred}_i$ produced by the mask generator, let $y^{cate}_i \in \mathbf{C}_{\text{train}}$ be the corresponding ground-truth category of $z^{pred}_i$. We invoke the text encoder $\mathcal{T}$ in the pre-trained CLIP model to encode the names of all categories in $\mathbf{C}_{\text{train}}$. This results in a set of text embeddings
$\mathcal{T}\left(\mathbf{C}_{\text{train}}\right) = \left\{\mathcal{T}\left(c_1\right), \ldots, \mathcal{T}\left(c_{|\mathbf{C}_{\text{train}}|}\right)\right\}$ where $c_k \in \mathbf{C}_{\text{train}}$ represents a category name. 

The loss for embedding classification (i.e., associating mask embeddings $m^{pred}_i$ with their categories $y^{cate}_i$) is calculated as:
\begin{align}
&\mathcal{L}_{\mathrm{ce}} = \notag \\ 
&\frac{1}{N}\sum_{i=1}^N \text{CE}\left(\text{Softmax}\left(\frac{z^{pred}_i  \mathcal{T}\left(\mathbf{C}_{\text{train}}\right)}{\tau}\right), y^{cate}_i\right)
\end{align}
where $\tau$ is a learnable temperature parameter and $\text{CE}$ is the cross-entropy loss for the classification of each training embedding.

The total loss for the training of our pipeline is finally defined as,
\begin{align}
    \label{eq:final_loss}
    \mathcal{L} = \alpha \mathcal{L}_{\mathrm{bce}} + \mathcal{L}_{\mathrm{dice}} + \mathcal{L}_{\mathrm{ce}}
\end{align}
where $\alpha$ is a hyper-parameter, we empirically set to 0.4.

In line with the work by~\citep{mask2former}, we apply the Hungarian matching algorithm~\citep{hungarian} to match predicted masks with ground-truth masks and compute the loss between the matching pairs. 

%%===========================================================================================%%

\section{Experiments}
\label{sec:exp}

\subsection{Datasets}

Following previous studies~\citep{zheng2021zero,xu2023odise,MaskCLIP,vpd}, we used the instance segmentation part of the MS-COCO dataset~\citep{coco} with 80 object categories to pre-train our model. We then fine-tuned the model on 3,040 images from the training set of the COD10K-v3 dataset~\citep{fan2022concealed}. Pre-training the model on the MS-COCO dataset aims to learn general knowledge about objects in the wild, while fine-tuning the model on the COD10K-v3 dataset adapts the model to camouflaged objects. We empirically found that this strategy significantly boosts up the performance of our method.

We tested our method and others on two benchmark camouflaged object datasets: the test set of the COD10K-v3 (including 2,026 images) and the NC4K~\citep{yunqiu_cod21} (including 4,121 images). The NC4K dataset contains only test images. The training sets (for both pre-training and fine-tuning) and the test sets (for both the COD10K-v3 and NC4K) share only six common object categories (out of 80 and 69 object categories from the MS-COCO and COD10K-v3/NC4K, respectively). This setting, i.e., cross-dataset training-testing, has been used widely in the evaluation of the generalisation ability of CIS models. It reflects the practicality of CIS, thus ensuring the reliability of evaluations.

We also evaluated our method on generic open-vocabulary datasets, including the ADE20K~\citep{ade20k} and Cityscapes~\citep{cityscapes}. For the ADE20K dataset, we used the validation set of the short version~\citep{ade20k-short} covering 150 object categories and 2,000 images. The Cityscapes dataset contains a total of 19 classes, which are divided into 11 ``stuff'' and 8 ``thing'' classes. We conducted evaluations on the validation set of the Cityscapes, including 500 images. Note that we pre-trained our method on the MS-COCO dataset and then directly evaluated the method on these open-vocabulary datasets without fine-tuning. 

\subsection{Implementation Details}

We implemented our method in Pytorch and built it on the Detectron2 framework~\citep{wu2019detectron2}. We trained our method for 90k iterations with a batch size of 64 on 4 NVIDIA A40 GPUs. All training images were resized to $512 \times 512$-pixels. Random jitters in the range $[0.1,2.0]$ were applied to the training images. We froze both the SD (v1.3) and CLIP models during training. We adopted the Adam optimiser~\citep{adamw} with the learning rate $\gamma$ set to $10^{-4}$ and weight decay of 0.05. We used a step learning rate scheduler and reduced the learning rate by a factor of 10 at 81k and 86k iterations. 

The training took around 4.3 days to complete. Due to class imbalance in the COD10K-v3 dataset, we manually removed some extremely rare classes, \eg, classes with less than five instances. In addition, we applied the \texttt{RepeatFactorTrainingSampler} from the Detectron2 framework, to allow a sample to appear more times than others based on its repeat factor.

\begin{table*}[!t]
\centering
\caption{Comparison of our method with existing instance segmentation methods on the test set of the COD10K-v3 and the NC4K datasets. 
Methods of the ``closed-set supervised learning approach'' are trained on the training set of the COD10K-v3 dataset. Methods of the ``open-vocab text-to-image approach'' are pre-trained on the MS-COCO dataset. 
We denote ``Ours'' and ``Ours (task-specific)'' for two variants of our method without and with fine-tuning on the training set of the COD10K-v3 dataset. ``Params'' denotes the number of \textit{trainable} / total parameters. 
The best results are \textbf{bold}, and the second best results are \underline{underline}.
}
\vspace{2mm}
\setlength{\tabcolsep}{5pt}
\renewcommand{\arraystretch}{1.2}
\resizebox{0.99\textwidth}{!}{%
\begin{tabular}{cl|ccc|ccc|c}
\toprule
\multicolumn{2}{c}{\multirow{2}{*}{\textbf{Method}}} & \multicolumn{3}{|c|}{\textbf{COD10K-v3 Test}} & \multicolumn{3}{c|}{\textbf{NC4K}} & \multirow{2}{*}{\textbf{\begin{tabular}[c]{@{}c@{}}Params \\ (Millions)\end{tabular}}} \\ \cmidrule(lr){3-8}
 &  & AP & AP50 & AP75 & AP & AP50 & AP75 &  \\ \midrule \midrule
\multirow{17}{*}{\begin{tabular}[c]{@{}l@{}}closed-set \\ supervised \\ learning \end{tabular}} & Mask R-CNN~\citep{maskrcnn} & 25.0 & 55.5 & 20.4 & 27.7 & 58.6 & 22.7 & 43.9/43.9 \\
 & MS R-CNN~\citep{msrcnn} & 30.1 & 57.2 & 28.7 & 31 & 58.7 & 29.4 & 60.0/60.6 \\
 & Cascade R-CNN~\citep{cascadercnn} & 25.3 & 56.1 & 21.3 & 29.5 & 60.8 & 24.8 & 71.7/71.7 \\
 & HTC~\citep{htc} & 28.1 & 56.3 & 25.1 & 29.8 & 59.0 & 26.6 & 76.9/76.9 \\
 & YOLACT~\citep{yolact} & 24.3 & 53.3 & 19.7 & 32.1 & 65.3 & 27.9 & 35.3/35.3 \\
 & BlendMask~\citep{blendmask} & 28.2 & 56.4 & 25.2 & 27.7 & 56.7 & 24.2 & 35.8/35.8 \\
 & SOLOv2~\citep{wang2020solov2} & 32.5 & 63.2 & 29.9 & 34.4 & 65.9 & 31.9 & 46.2/46.2 \\
 & Condlnst~\citep{condinst} & 30.6 & 63.6 & 26.1 & 33.4 & 67.4 & 29.4 & 34.1/34.1 \\
 & Querylnst~\citep{queryinst} & 28.5 & 60.1 & 23.1 & 33.0 & 66.7 & 29.4 & 172.5/172.5 \\
 & SOTR~\citep{guo2021sotr} & 27.9 & 58.7 & 24.1 & 29.3 & 61.0 & 25.6 & 63.1/63.1 \\
 & MaskFormer~\citep{maskformer} & 38.2 & 65.1 & 37.9 & 44.6 & 71.9 & 45.8 & 45.0/45.0 \\
 & Mask2Former~\citep{mask2former} & 39.4 & 67.7 & 38.5 & 45.8 & 73.6 & 47.5 & 43.9/43.9 \\
 & Mask Transfiner~\citep{masktransfiner} & 28.7 & 56.3 & 26.4 & 29.4 & 56.7 & 27.2 & 44.3/44.3 \\
 & OSFormer~\citep{pei2022osformer} & 41.0 & 71.1 & 40.8 & 42.5 & 72.5 & 42.3 & 46.6/46.6 \\
 & DCNet~\citep{dcnet} & \textbf{45.3} & 70.7 & \textbf{47.5} & \underline{52.8} & \textbf{77.1} & \textbf{56.5} & 53.4/53.4 \\ 
 & MSPNet~\citep{li2024multi} & 39.7 & 69.8 & 39.8 & 41.8 & 71.8 & 42.3 & 48.1/48.1 \\ 
 & UQFormer~\citep{dong2024unified} & \underline{45.2} & \underline{71.6} & 46.6 & 47.2 & 74.2 & 49.2 & 37.5/37.5 \\ 
 & CamoFA~\citep{CamoFA} & 43.5 & \textbf{74.9} & 42.7 & 45.0 & 75.7 & 44.3 & - \\ 
 & \cellcolor{gray!25}Ours (task-specific) & \cellcolor{gray!25}45.1 & \cellcolor{gray!25}71.1 & \cellcolor{gray!25}\underline{47.4} & \cellcolor{gray!25}\textbf{52.9} & \cellcolor{gray!25}\underline{76.8} & \cellcolor{gray!25}\underline{55.9} & \cellcolor{gray!25}28.7/1522.7 \\ 
 \midrule  
 \multirow{6}{*}{\begin{tabular}[c]{@{}l@{}}open-vocab \\ VLM \\ (\textit{w/o finetuning}) \end{tabular}} 
 & MaskCLIP~\citep{MaskCLIP} & 3.3 & 5.9 & 4.1 & 6.3 & 5.6 & 6.5 & 542.0/542.0 \\
 & MasQCLIP~\citep{xu2023masqclip} & 4.1 & 7.7 & 5.8 & 8.0 & 7.6 & 8.4 & 375.2/357.2 \\
 & X-Decoder~\citep{xDecoder} & 7.7 & 12.9 & 7.5 & 3.9 & 8.1 & 3.4 & 38.3/38.3 \\
 & SEEM~\citep{zou2023segment} & 6.6 & 10.8 & 6.5 & 9.2 & 12.7 & 9.9 & 415.3/415.3 \\
 & OpenSeeD~\citep{OpenSeeD} & 6.1 & 10.4 & 5.9 & 9.3 & 14.5 & 9.8 & 116.2/116.2 \\
 & TPNet~\citep{he2024text} & 18.3 & \underline{41.8} & 14.3 & 21.4 & \textbf{48.3} & 16.6 & 71.78/71.78 \\ 
\cmidrule(lr){2-9}
 \multirow{2}{*}{\begin{tabular}[c]{@{}l@{}}open-vocab T2I \\ (\textit{w/o finetuning}) \end{tabular}} 
 & ODISE~\citep{xu2023odise} & \underline{21.1} & 37.8 & \underline{20.5} & \underline{22.9} & 37.2 & \underline{21.4} & 28.1/1522.1 \\
 & \cellcolor{gray!25}Ours & \cellcolor{gray!25}\textbf{23.9} & \cellcolor{gray!25}\textbf{44.3} & \cellcolor{gray!25}\textbf{23.1} & \cellcolor{gray!25}\textbf{24.8} & \cellcolor{gray!25}\underline{44.2} & \cellcolor{gray!25}\textbf{23.9} & \cellcolor{gray!25}28.7/1522.7 \\
 \bottomrule
\end{tabular}%
}
\label{tab:cis}
% \vspace{-3mm}
\end{table*}

\subsection{Results}

We evaluated our method and existing CIS methods using the average precision (AP) values measured at different intersection-over-union (IOU) thresholds. In particular, we calculated the overall AP in 
the range $[50\%,95\%]$ for the IOU thresholds (i.e., for a threshold within the above range, a predicted instance is considered as true positive if there exists a true instance in the ground-truth such that their IOU is equal or greater than that threshold). We also measured detailed AP for the IOU thresholds of $50\%$ (AP50) and $75\%$ (AP75).

\subsubsection{Camouflaged Object Datasets}

We report the performance of our method on camouflaged object datasets (COD10K-v3 and NC4K) in~\Cref{tab:cis} (last row). Recall that, following the conventional setting in CIS, e.g.,~\citep{zheng2021zero,xu2023odise,MaskCLIP,vpd}, we pre-trained our model on the MS-COCO dataset and then fine-tuned it on the training set of the COD10K-v3 dataset. To show the effectiveness of this strategy, we experimented with a variant of our method by skipping the fine-tuning phase. In particular, we pre-trained our method on the MS-COCO dataset and then evaluated it directly on the test set of the COD10K-v3 and the NC4K datasets. We show the performance of this strategy in the last row, denoted as ``Ours'', in~\Cref{tab:cis}. Experimental results show that fine-tuning the method on a camouflaged object dataset, denoted as ``Ours (task-specific)'', significantly improves its performance on all evaluation metrics.

We compare our method with existing instance segmentation methods on the CIS task in~\Cref{tab:cis}. We group existing methods into two groups. We name the first group ``closed-set supervised learning approach''. The methods of this approach follow the traditional fashion, which supervises an instance segmentation model on a training set and tests the model on a test set. This approach's training and test sets are in the same domain and include imagery data only. Most existing instance segmentation methods in the field can be customised to enable CIS using this approach. In our experiments, the methods of the first group are trained on the training set of the COD10K-v3 dataset. The second group, called the ``open-vocab approach,'' includes methods using the vision and language model (VLM) and text-to-image diffusion techniques with open-vocabulary.

As shown in~\Cref{tab:cis}, our method with full setting (pre-training and fine-tuning), denoted as ``Ours (task-specific)'', significantly outperforms ODISE on all evaluation metrics, making a new state-of-the-art for OVCIS. Our method also performs on par with recent methods (DCNet~\citep{dcnet}, MSPNet~\citep{li2024multi}, UQFormer~\citep{dong2024unified}, and CamoFA~\citep{CamoFA}). Nevertheless, compared with recent methods, our method requires much fewer \textit{trainable} parameters (see the last column in~\Cref{tab:cis}).~\Cref{tab:cis} also compares all the methods in terms of the number of parameters used.

In summary, with regard to both segmentation accuracy and memory usage, our method is more advanced, compared with existing ones. Recall that only six object categories are shared between the MS-COCO dataset (with 80 object categories) and the COD10K-v3/NC4K dataset (with 69 object categories). This challenge shows the ability of our method in handling open-vocabulary tasks. 

\begin{figure*}
    \centering
    \includegraphics[width=0.99\linewidth]{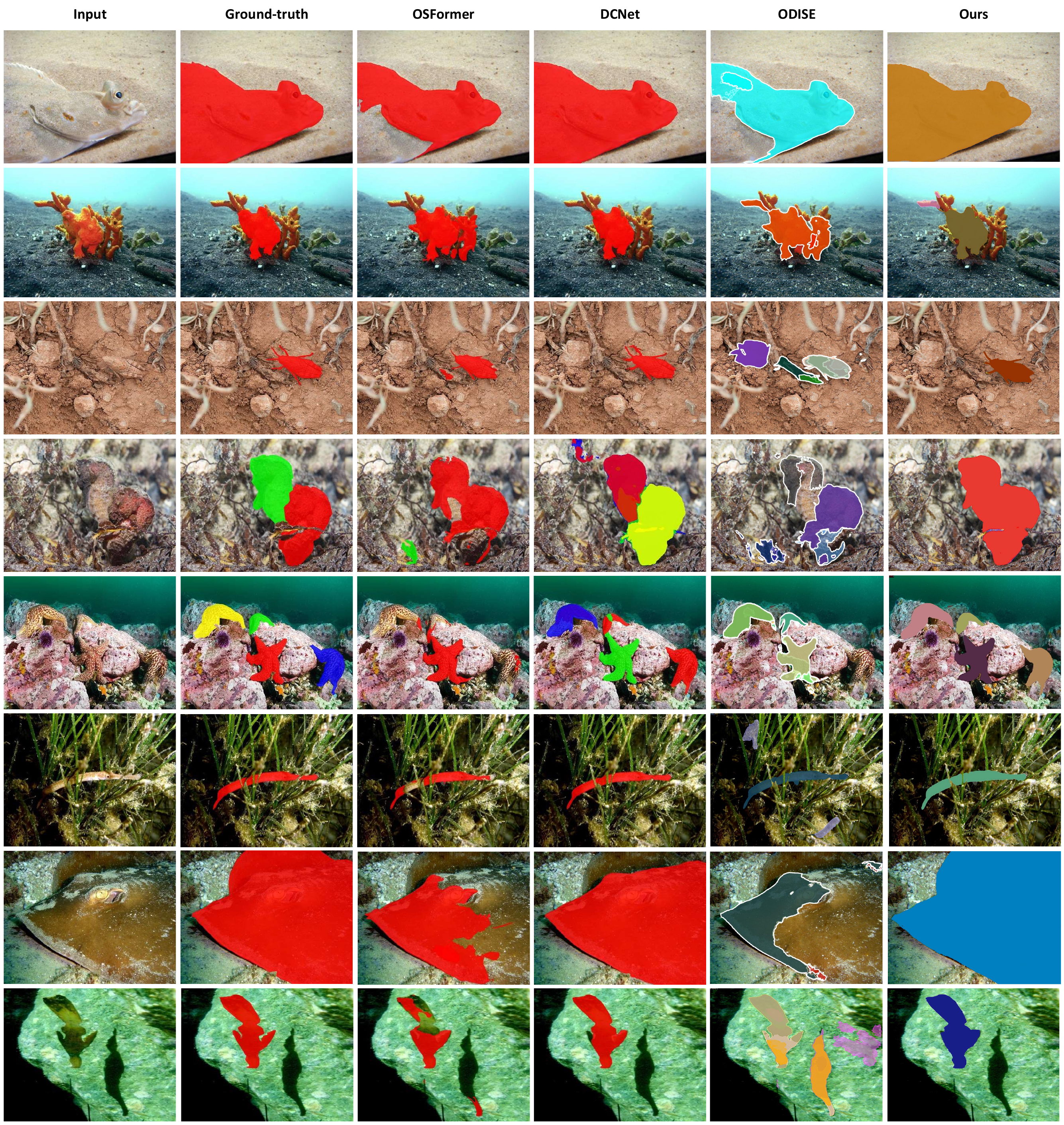}
    \vspace{2mm}
    \caption{Qualitative comparison of our method with existing methods on the COD10K-v3 and NC4K datasets. This figure is best viewed in colour. }
    \label{fig:cod_vis}
\end{figure*}

We visualise several results of our methods and existing ones in~\Cref{fig:cod_vis}. As shown, our method excels at pixel-level instance segmentation, accurately delineating camouflaged objects along their blurry boundaries in cluttered backgrounds. The results also demonstrate our proficiency in segmenting multiple instances.

\begin{figure*}[ht]
  \centering
  \includegraphics[width=0.99\linewidth]{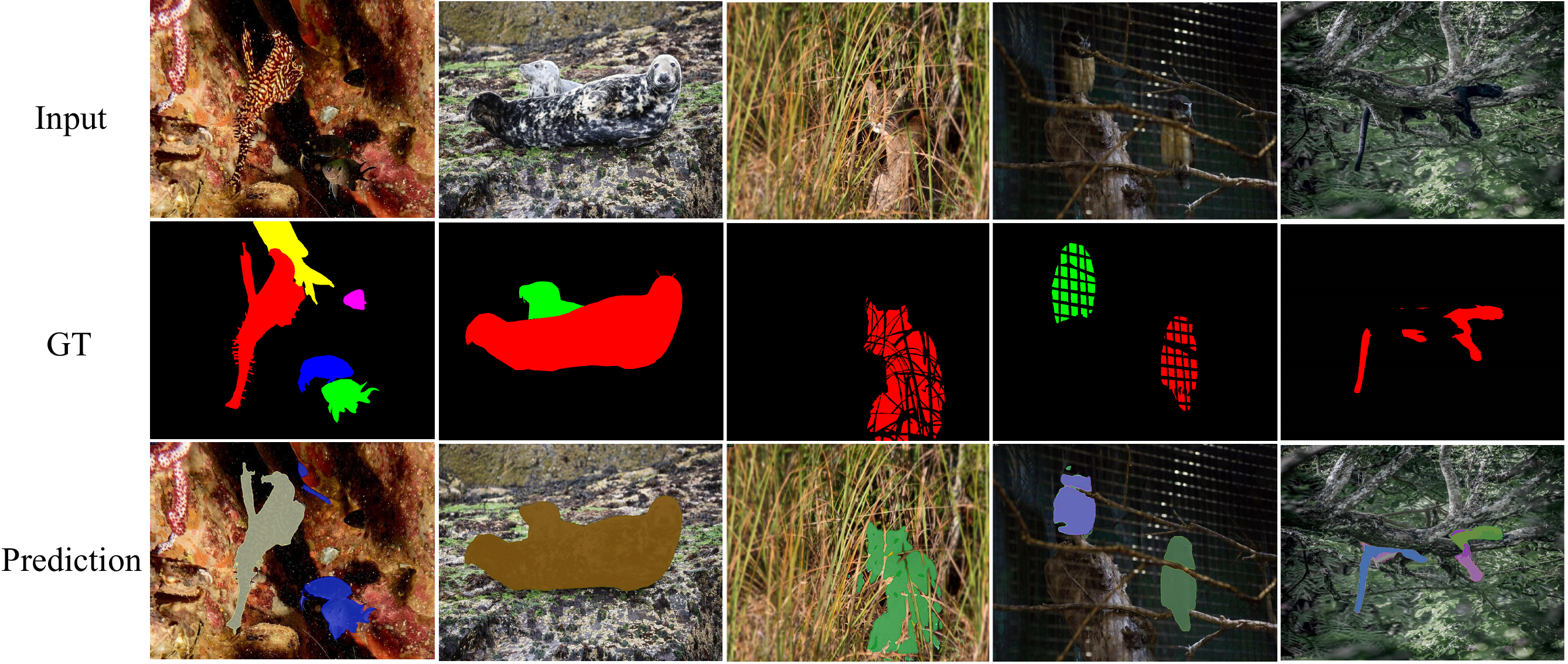}
  \vspace{2mm}
  \caption{Failure cases of our method on the COD10K-v3 dataset. In the first and second columns, our method fails to separate instances of nearby and similar objects, such as the yellow fish and two sea lions. Our method can detect and segment camouflaged objects in the third and fourth columns but with slightly less accurate boundaries. In the last column, our method struggles with the significant spatial separation of the black panther's body parts, leading to misclassification of the entire object. This figure is best viewed in colour.}
  \label{fig:failure}
  \vspace{-6mm}
\end{figure*}

In addition, \Cref{fig:failure} illustrates failure cases of our method. We found that our method would be ineffective in distinguishing and separating an object that shares very similar characteristics with others or consists of fragmented parts. However, such circumstances would also be challenging for human beings as well. 

\subsubsection{Generic Open-Vocabulary Datasets}

To showcase the versatility and generality of our method in various application domains (other than camouflaged objects), we evaluated our method on the ADE20K~\citep{ade20k} and Cityscapes datasets~\citep{cityscapes}, two widely used open-vocabulary benchmark datasets. Note that these datasets are not designed for camouflaged object detection and segmentation. We summarise the performance of our method and existing open-vocabulary instance segmentation methods on these two datasets in~\Cref{tab:generic-datasets}.

Our method ranks second on both the ADE20K and Cityscapes datasets. Nevertheless, compared with the first ranked method, i.e., OpenSeeD~\citep{OpenSeeD}, our method uses approximately four times fewer trainable parameters than OpenSeeD, while scarifying less than 1\% and 8\% of the overall AP on the ADE20K and Cityscapes datasets, respectively.

\begin{table*}[!t]
\centering
\setlength{\tabcolsep}{6pt}
\renewcommand{\arraystretch}{1.2}
\small
\caption{Comparison of our method with existing open-vocabulary instance segmentation methods on the ADE20K and Cityscapes datasets. We measure the accuracy of the segmentation using the AP. ``-'' denotes no-report performance. The best results are \textbf{bold}, and the second best results are \underline{underline}. In the last two columns, we also report the number of trainable and total parameters used in the methods.}
\vspace{2mm}
\resizebox{\textwidth}{!}{%
\begin{tabular}{l|Hc|c|c|c}
\toprule
\textbf{Method} & \textbf{GFLOPs} &  \textbf{ADE20K} & \textbf{Cityscapes} & \textbf{Trainable Params (M)} & \textbf{Total Params (M)}\\ \midrule \midrule
MaskCLIP~\citep{MaskCLIP} & 542 & 6.1 & - & 354.1 (100\%) & 354.1 \\
ODISE~\citep{xu2023odise} & - & 13.9 & - & 28.1 (1.85\%) & 1522.1 \\
X-Decoder~\citep{xDecoder} & - & 13.1 & 24.9 & 38.3 (100\%) & 38.3 \\
OpenSeeD~\citep{OpenSeeD} & - & \textbf{15.0} & \textbf{33.2} & 116.2 (100\%) & 116.2 \\
\rowcolor{gray!25} Ours & xxx & \underline{14.1} & \underline{25.6} & 28.7 (1.88\%) & 1522.7 \\ \bottomrule
\end{tabular}%
}
\label{tab:generic-datasets}
\end{table*}

\subsection{Ablation Studies}
\label{sec:ablation}

In this section, we present ablation studies to validate different aspects of the design and implementation of our method. First, we investigated the impact of prompt engineering and prompt templates on OVCIS tasks. Second, we validated the technical modules developed in our method to make it specialised to CIS tasks.

\subsubsection{Prompt Engineering for OVCIS}

For open-vocabulary-based studies, an object category can be specified by multiple alternative text descriptions. For instance, the ``cat'' category can be described as ``cat'', ``cats'', ``kitty'', or ``kitties''. To improve the diversity of open-vocabulary in text prompts, we applied the identical prompt engineering method introduced by~\citep{OpenSeg} to assemble a list of synonyms, subcategories, and plurals for the categories. Given a text prompt, the category is chosen as the one with the highest probability from an ensembling list of multiple alternative queries. We observed that the prompt engineering technique is simple yet effective in improving the segmentation accuracy of our method.~\Cref{tab:prompt} shows the impact of applying prompt engineering to CIS.

\begin{table}[t]
\centering
\small
\setlength{\tabcolsep}{7pt}
\renewcommand{\arraystretch}{1.2}
\caption{Ablation study on applying \textbf{prompt engineering} to improve OVCIS. Results are tested on the COD10K-v3 dataset.}
\label{tab:prompt}
% \vspace{4mm}
% \resizebox{0.48\textwidth}{!}{%
\begin{tabular}{c|lll}
\toprule
\textbf{Prompt} & \textbf{AP} & \textbf{AP50} & \textbf{AP75} \\ \midrule \midrule
\xmark & 22.8 & 43.1 & 22.1 \\
\cmark & 23.4 \textcolor{teal}{+0.6} & 43.8 \textcolor{teal}{+0.7} & 22.6 \textcolor{teal}{+0.5} \\ 
\bottomrule
\end{tabular}%
% }
\end{table}

\subsubsection{Prompt templates for OVCIS}

Inspired from~\cite{OVCOS_ECCV2024}, we apply the prompt template set, which considers task attributes and shows better performance in~\Cref{tab:template}. We can see that using the prompt template can affect the influence of different templates on semantic embedding, which inspires further explorations for more effective prompt engineering.

\begin{table*}[t]
\centering
% \small
\setlength{\tabcolsep}{7pt}
\renewcommand{\arraystretch}{1.2}
\caption{Ablation study on applying \textbf{prompt templates} to improve OVCIS. Results are tested on the COD10K-v3 dataset. }
\label{tab:template}
% \vspace{2mm}
% \resizebox{0.48\textwidth}{!}{%
\begin{tabular}{l|l}
\toprule
\textbf{Task-related templates} & \textbf{AP} \\
\midrule \midrule
``A photo of \textless class\textgreater." & 23.4 \\
\midrule
Using multiple templates:  &  \multirow{7}{*}{23.9 \textcolor{teal}{+0.5}}\\
\qquad \quad \ding{43} ``A photo of the camouflaged \textless class\textgreater." &  \\
\qquad \quad \ding{43} ``A photo of the concealed \textless class\textgreater." &  \\
\qquad \quad \ding{43} ``A photo of the \textless class\textgreater\ camouflaged in the background." &  \\
\qquad \quad \ding{43} ``A photo of the \textless class\textgreater\ concealed in the background." &  \\
\qquad \quad \ding{43} ``A photo of the \textless class\textgreater\ camouflaged to blend in with its surroundings." &  \\
\qquad \quad \ding{43} ``A photo of the \textless class\textgreater\ concealed to blend in with its surroundings." &  \\
\bottomrule
\end{tabular}
% }
\end{table*}

\begin{table*}[t]
\centering
\small
\caption{Ablation study on the effectiveness of the proposed technical modules to CIS. Results are tested on the COD10K-v3 dataset by using the AP metric.}
\vspace{2mm}
\label{tab:ablations}
\resizebox{\textwidth}{!}{%
\begin{tabular}{l|ll}
\toprule
\textbf{Variant}                                     & \textbf{Ours} & \begin{tabular}[c]{@{}l@{}}\textbf{Ours} \\ \textbf{(task-specific)} \end{tabular} \\ \midrule \midrule
no text (text embeddings = 0)                         & 12.2 {\scriptsize \textcolor{maroon}{-7.1}}  & 31.4 {\scriptsize \textcolor{maroon}{-13.5}}                \\
skip MSFF module (only the last layer of the diffusion U-Net is used)   & 18.4 {\scriptsize \textcolor{maroon}{-0.9}} & 40.5 {\scriptsize \textcolor{maroon}{-4.4}}                \\
skip MSFF module (concatenation of all multiscale features)   & 18.1 {\scriptsize \textcolor{maroon}{-1.2}} & 39.8 {\scriptsize \textcolor{maroon}{-5.1}}                \\
skip CIN module (directly use the TVA's output for instance classification) & 17.6 {\scriptsize \textcolor{maroon}{-1.7}} & 37.7 {\scriptsize \textcolor{maroon}{-7.2}}                \\
skip TVA module (element-wise dot product of mask embedding and text embedding) & 18.8 {\scriptsize \textcolor{maroon}{-0.5}} & 42.7 {\scriptsize \textcolor{maroon}{-2.2}}                \\ \midrule
Full setting                                            & \textbf{19.3} & \textbf{44.9}       \\ \bottomrule
\end{tabular}%
}
\end{table*}

\subsubsection{CIS-Specialised Modules}

\begin{figure*}[t]
\centering
\includegraphics[width=0.99\linewidth]{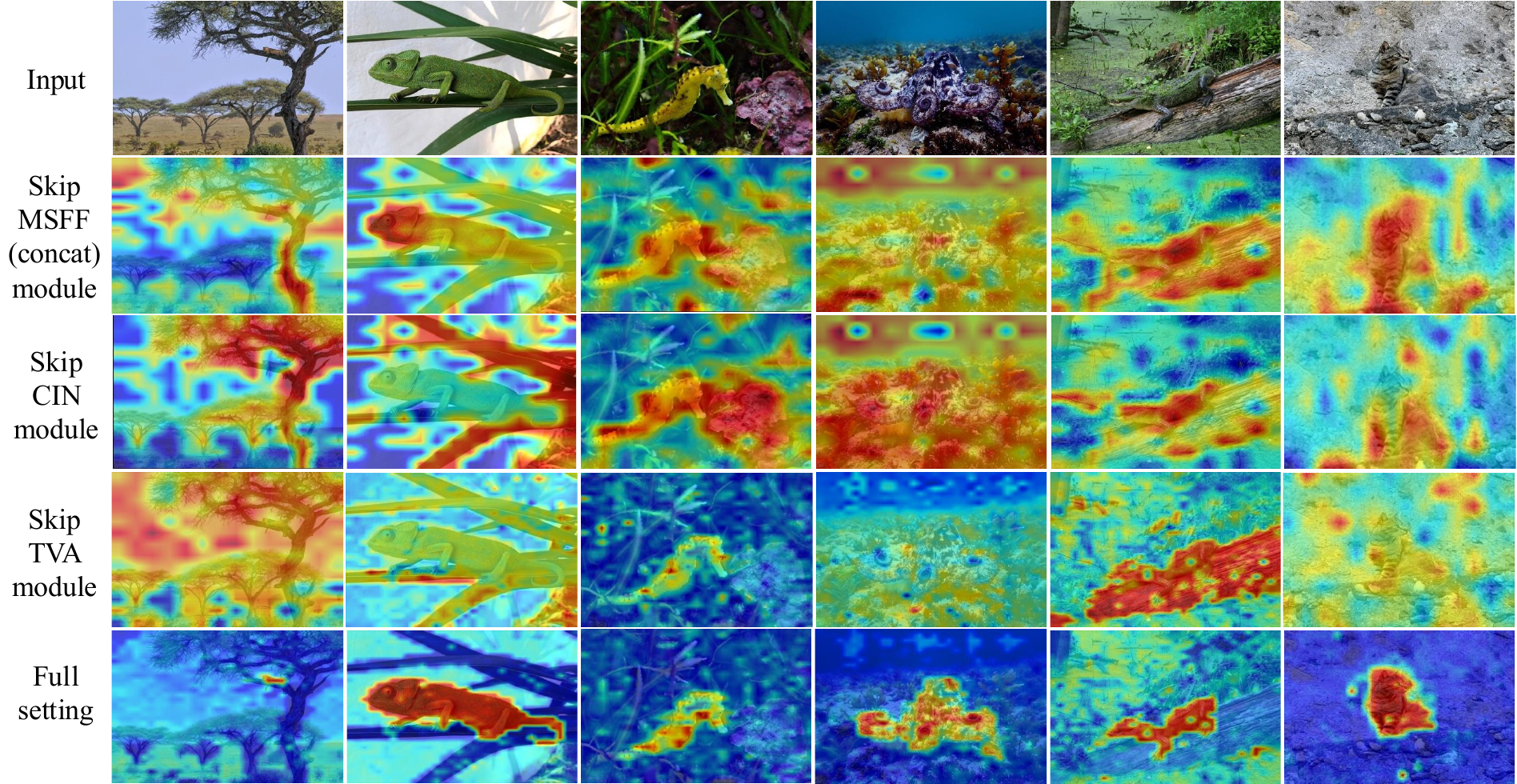}
\vspace{2mm}
\caption{\textbf{Qualitative intermediate outputs for module ablations.} The attention map (interim result) is a heat map of an object instance where foreground pixels are highlighted in red and background pixels are represented in blue. These intermediate outputs explain the quantitative gains in~\Cref{tab:ablations}: the skip MSFF module (concat), the skip CIN module, the skip TVA module, and the full setting. This figure is best viewed in colour.}
\label{fig:ablation_vis}
\end{figure*}

We developed several technical modules in our method to make it specialised to CIS. We refer the reader to~\Cref{fig:camo_pipeline} for a recall on how the modules are configured in our pipeline. To confirm the importance of those modules, we experimented with different variants of our method, each variant is made by alteration and/or omission of a module. We pre-trained the variants on the MS-COCO dataset for 30k iterations, then tested them on the test set of the COD10K-v3 dataset. We present the results of this ablation study in~\Cref{tab:ablations} and visualise the impact of the different modules in~\Cref{fig:ablation_vis}. 

We validated the importance of the use of text in our method (in the 1st row of~\Cref{tab:ablations}). This was implemented by setting the text embeddings used in the method to zeros. We observed a significant drop in the performance of this variant, resulting in the lowest AP (12.2). This indicates that text embeddings play a crucial role as they provide essential contextual or semantic information that helps to identify camouflages.  

% We propose the MSFF module to fuse image-guided features learnt by the diffusion model at multiple scales. We proved the necessity of this module by directly feeding the last layer of the diffusion U-Net to the mask generator. Experimental results show that this variant incurs a performance loss. However, compared with the full setting, which fuses all the layers from both the encoder and decoder of the diffusion U-Net, the last layer of the diffusion U-Net seems to carry substantial information for the CIS task. 
We propose the MSFF module to fuse image-guided features learnt by the diffusion model at multiple scales. We validated the design of this module by comparing it with the standard fusion approach that concatenates the multiscale features from the encoder with the last layer of the decoder of the diffusion U-Net. The experimental results (in the 2nd and 3rd row of~\Cref{tab:ablations}) show that the standard fusion approach incurs a performance loss. Moreover, compared with the full setting, which fuses all layers of both the encoder and decoder of the diffusion U-Net, the last layer of the diffusion U-Net appears to carry substantial information for the CIS task.

We develop the CIN module to further enhance the representations of camouflaged objects, such as prediction and classification. To validate the CIN module, we removed it from our pipeline by directly passing the output from the TVA module to mask prediction and classification. We found that, by omitting the CIN module, the AP of the pipeline decreases dramatically (from 19.3 to 17.6), as shown in the 4th row of~\Cref{tab:ablations}. 

We devise the TVA module to aggregate textual and visual features in an instance-oriented manner, i.e., textual and visual features are aggregated alongside instance masks and consolidated against the background via feature weighting. To validate this module, we simplified its operation by applying an element-wise dot product on the input mask embeddings and text embeddings. We observed that, compared with other modules, the TVA module is less critical, which is evident by the low performance drop when simplification is applied to its architecture (see the 5th row of~\Cref{tab:ablations}).

\subsection{Additional Analysis}
\label{sec:additional}

\begin{figure}
    \centering
    \includegraphics[width=0.99\linewidth]{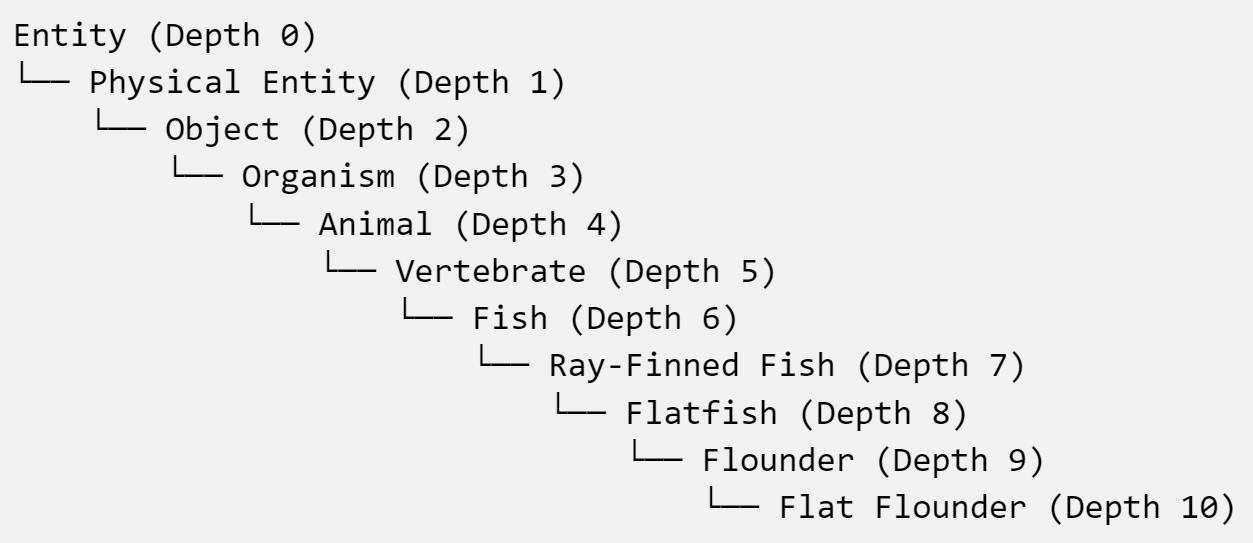}
    % \vspace{2mm}
    \caption{Sample of hierarchical structure of ``Summer Flounder'' with the hypernym's depth = 10.}
    \label{fig:hierarchical_structure}
\end{figure}

In our method, we utilised CLIP~\citep{CLIP} (text and image encoders) to extract textual and visual features. We showcase CLIP's capability in~\Cref{tab:classification}, where we evaluate CLIP in performing zero-shot classification of camouflaged objects on different datasets (COD10K-v3, NC4K, CAMO~\citep{CAMO}). Specifically, we applied the~\href{https://www.nltk.org/howto/wordnet.html}{NLTK's WordNet} to extract the animal type from each image's caption generated by ClipCap~\citep{mokady2021clipcap} and check if the animal type and corresponding ground-truth category share the same hierarchical semantic relation (depth of the hypernym = 10)~\citep{fu2014learning}. In detail, the depth value helps in understanding the position and specificity of a concept within a hierarchical structure (the higher the depth, the more specific the concept). The example of the ``Summer Flounder'' is shown in~\Cref{fig:hierarchical_structure}.

\begin{table}[!t]
% \begin{wraptable}{r}{0.34\textwidth}
    % \small
    \centering
    % \vspace{-1mm}
    %\hspace{-8mm} % Negative horizontal space to nudge the table left
    \caption{Zero-shot image classification using CLIP on camouflaged datasets.}
    \vspace{2mm}
    \label{tab:classification}
    \setlength{\tabcolsep}{6pt}
    \renewcommand{\arraystretch}{1.3}
    % \resizebox{0.24\textwidth}{!}{%
    \begin{tabular}{c|ccc}
        \toprule
         & \textbf{COD10K-v3} & \textbf{NC4K}  & \textbf{CAMO} \\ \midrule \midrule
        \textbf{Accuracy (\%)} & 48.06 & 45.69 & 46.48 \\ \bottomrule
    \end{tabular}
    % }
    % \vspace{-6mm}
% \end{wraptable}
\end{table}

In addition, we conducted a \textbf{``prompt coarsening'' ablation} by systematically replacing fine-grained category names with their WordNet hypernyms, same as in our CLIP analysis above (\eg, cat $\rightarrow$ feline $\rightarrow$ mammal $\rightarrow$ animal) and reporting performance degradation. Because ``animal'' is semantically related but less discriminative, we expect decreased class separability among subclasses, which this stress test will quantify. To make this evaluation meaningful under hierarchical substitutions, we will report both standard AP (exact-label) and a semantics-aware metric (e.g., Open AP~\citep{zhou2025rethinking}), which explicitly accounts for semantic similarity between predicted and ground-truth names as advocated in prior work on open-vocabulary evaluation. We reported the results in Table~\ref{tab:coarser_prompts}. As shown, the standard AP exhibits a significant decrease, whereas Open AP indicates only a slight decline. This is mainly due to the stability of localization in Open AP, while fine-grained naming necessitates specific prompts. This behavior aligns with the expected performance of a text-conditioned open-vocabulary system.

% {\scriptsize \textcolor{maroon}{-7.1}}
\begin{table*}[t]
\centering
\caption{Results on OVCIS evaluated on ``coarser prompts'' by vanilla and Open AP on the test set of the COD10K-v3 and the NC4K datasets.}
\label{tab:coarser_prompts}
\vspace{2mm}
\setlength{\tabcolsep}{5pt}
\renewcommand{\arraystretch}{1.2}
\resizebox{0.75\textwidth}{!}{%
\begin{tabular}{@{}c|c|ccc|ccc@{}}
\toprule
\multirow{2}{*}{\textbf{Coarser Prompts}} & \multirow{2}{*}{\textbf{Metric}} & \multicolumn{3}{c|}{\textbf{COD10K-v3 Test}} & \multicolumn{3}{c}{\textbf{NC4K}} \\ 
\cmidrule(l){3-8} 
& & \textbf{AP} & \textbf{AP50} & \textbf{AP75} & \textbf{AP} & \textbf{AP50} & \textbf{AP75} \\ 
\midrule \midrule
\xmark & Vanilla AP & 23.9 & 44.3 & 23.1 & 24.8 & 44.2 & 23.9 \\
\cmark & Vanilla AP & 22.7 & 42.1 & 21.9 & 23.6 & 41.9 & 22.7 \\
\cmark & Open AP    & 23.3 & 43.2 & 22.5 & 24.2 & 43.1 & 23.3 \\ 
\bottomrule
\end{tabular}%
}
\end{table*}

We also evaluated our work in the COD setting, where only binary masks are considered. Specifically, we experimented with our method on benchmark COD datasets, including CAMO, Chameleon, and COD10K-v2. We report the results of this experiment in~\Cref{tab:cod}. The results demonstrate the superiority of our method over existing COD baselines. In detail, while Camouflous~\citep{Khan_2024_WACV} reports low MAE values on CAMO (0.043 MAE) and COD10K-v2 (0.021 MAE), our model achieves the highest $F$, $S$, and $E$ on these datasets, including leading $F$ scores on CAMO (0.847 $F$) and COD10K-v2 (0.807 $F$). Compared with the models C2F-Net~\citep{sun2021c2fnet} and BCNet~\citep{BCNet}, which emphasize global context fusion and boundary-aware refinement, our method strikes a balance between semantic precision and contextual depth, thereby producing more robust segmentation results. Notably, on Chameleon, our method slightly trails Camouflous in MAE (0.119 vs. 0.021) but still delivers the highest $E$ (0.959), indicating stronger overall object integrity. These results validate the effectiveness of our task-specific design in COD.

\begin{table*}[t]
\caption{Comparison of our method with existing closed-set supervised learning camouflaged detection (binary segmentation) methods on the test set of the CAMO, Chameleon, and COD10K-v2 datasets. We adopt the results from~\citep{jamali2025context}.}
\label{tab:cod}
\setlength{\tabcolsep}{2pt}
\renewcommand{\arraystretch}{1.3}
\resizebox{\textwidth}{!}{%
\begin{tabular}{@{}l|cccc|cccc|cccc@{}}
\toprule
\multicolumn{1}{c|}{\multirow{2}{*}{\textbf{Method}}} & \multicolumn{4}{c|}{\textbf{CAMO}}                                                  & \multicolumn{4}{c|}{\textbf{Chameleon}}                                             & \multicolumn{4}{c}{\textbf{COD10K-v2}}                                              \\ \cmidrule(l){2-13} 
\multicolumn{1}{c|}{}                                 & \textbf{MAE ↓} & \textbf{$F$ ↑} & \textbf{$S$ ↑} & \textbf{$E$ ↑} & \textbf{MAE ↓} & \textbf{$F$ ↑} & \textbf{$S$ ↑} & \textbf{$E$ ↑} & \textbf{MAE ↓} & \textbf{$F$ ↑} & \textbf{$S$ ↑} & \textbf{$E$ ↑} \\ 
\midrule \midrule
C2F-Net~\citep{sun2021c2fnet}                             & 0.091          & 0.647                & 0.796                & 0.828                & 0.034          & 0.782                & 0.828                & 0.950                & 0.043          & 0.629                & 0.775                & 0.872                \\
DiCANet~\citep{DiCANet}                             & 0.068          & 0.790                & 0.830                & 0.886                & 0.028          & 0.776                & 0.853                & 0.914                & 0.032          & 0.676                & 0.802                & 0.890                \\
PCFNet~\citep{song2023pixel}                             & 0.053          & 0.840                & 0.844                & 0.913                & 0.023          & 0.876                & 0.912                & 0.957                & 0.027          & 0.751                & 0.838                & 0.924                \\
BCNet~\citep{BCNet}                       & 0.069          & 0.761                & 0.802                & 0.865                & 0.029          & 0.802                & 0.839                & 0.944                & 0.033          & 0.704                & 0.827                & 0.894                \\
CamoMFCF~\citep{app14062494}                               & 0.080          & 0.727                & 0.796                & 0.854                & 0.032          & 0.805                & 0.838                & 0.935                & 0.036          & 0.686                & 0.813                & 0.890                \\
Camouflous~\citep{Khan_2024_WACV}                         & \textbf{0.043} & \underline{0.842}                & \underline{0.873}                & \underline{0.926}                & \underline{0.021}          & \textbf{0.866}       & \textbf{0.902}       & \underline{0.958}                & \textbf{0.021} & \underline{0.802}                & \textbf{0.873}       & \textbf{0.935}       \\
Ours (task-specific)                                  & \underline{0.044}          & \textbf{0.847}       & \textbf{0.878}       & \textbf{0.931}       & \textbf{0.119} & \underline{0.865}                & \underline{0.901}                & \textbf{0.959}       & \underline{0.020}          & \textbf{0.807}       & \textbf{0.873}       & \underline{0.933}                \\ \bottomrule
\end{tabular}
}
\end{table*}

% \subsection{Potential Applications}
% \label{sec:application}

% potential applications

%%===========================================================================================%%

\section{Conclusion}
\label{sec:conclusion}

This work advances the computer vision research for open-vocabulary camouflaged instance segmentation (OVCIS) by leveraging text-to-image diffusion and text-image transfer techniques. To this end, we propose a method that effectively integrates textual information learnt from open-vocabulary into the visual domain to enrich the representations of camouflaged objects. We evaluate our method and compare it with existing methods in both CIS and generic open-vocabulary segmentation on benchmark datasets. Experimental results show the effectiveness and advantages of our method over existing baselines in both tasks.

\vspace{0.3cm}
\noindent\textbf{Limitations and Future Works.} Despite proven strengths, the proposed method has limitations. While the learnt knowledge from natural language can effectively distinguish an object from its background when visual cues are insufficient due to camouflages, it may not be helpful to separate touching/overlapping instances. Additionally, the method struggles with segmenting occluded objects. Under severe occlusions, a camouflaged object can be over-segmented into non-semantic fragments, leading to misclassification of the object. Enhancing object representations with background-aware features from open-vocabulary (i.e., by using text prompts including both foreground and background information, e.g., ``a lizard is on a tree'') may help to address the aforementioned issues. We consider this research direction to be our future work.

\vspace{0.3cm}
\noindent\textbf{Broader Impact.} 

Our study directly contributes to advance research on wildlife monitoring, ecological interactions, and evolutionary understanding related to camouflage in nature~\citep{troscianko2017quantifying,norouzzadeh2018automatically,beery2018recognition,simoes2023deepwild}. To the best of our knowledge, our work is the first open-vocabulary approach to camouflaged instance segmentation, offering advanced features such as zero-shot performance ability and multimodal enabling, improving the practicality of computer vision-based ecological studies. In addition, our work can significantly influence future developments in other fields, including, for instance, safety and security applications (e.g., military reconnaissance~\citep{Liu_neuro_2023}) and medical diagnostics (e.g., camouflaged colon polyp segmentation~\citep{Wang_computer_2024}).

\section*{Acknowledgement} This research is supported by an internal grant from HKUST (R9429), the National Research Foundation, Singapore, and DSO National Laboratories under the AI Singapore Programme (AISG Award No: AISG2-GC-2023-008), Career Development Fund (CDF) of Agency for Science, Technology and Research (A*STAR) (No.: C233312028), National Research Foundation, Singapore and Infocomm Media Development Authority under its Trust Tech Funding Initiative (No. DTC-RGC-04), a MAAP Discovery funding (2022-2025) from Deakin University and the Science Foundation Ireland under the SFI Frontiers for the Future Programme (22/FFP-P/11522). This work is partially done during Tuan-Anh Vu's research attachment at CFAR \& IHPC, A*STAR, Singapore.

\vspace{0.3cm}
\noindent\textbf{Availability of data and materials.} All datasets (MS-COCO dataset~\citep{coco}, COD10K-v3~\citep{fan2022concealed}, NC4K~\citep{yunqiu_cod21}, CAMO~\citep{CAMO}, ADE20K~\citep{ade20k}, and Cityscapes~\citep{cityscapes}) used in our manuscript are available online on their websites. All related materials (models, codes, \etc) will be available online upon acceptance.

%%===========================================================================================%%

% \begin{appendices}

% \section{Experimental Setup}
% \subsection{Datasets}

% \end{appendices}

%%===========================================================================================%%
%% If you are submitting to one of the Nature Portfolio journals, using the eJP submission   %%
%% system, please include the references within the manuscript file itself. You may do this  %%
%% by copying the reference list from your .bbl file, paste it into the main manuscript .tex %%
%% file, and delete the associated \verb+\bibliography+ commands.                            %%
%%===========================================================================================%%

\bibliography{refs}% common bib file

\begin{thebibliography}{}
\renewcommand{\doi}[1]{\url{https://doi.org/#1}}
\bibcommenthead

\bibitem [\protect \citeauthoryear {%
Baranchuk%
\ \protect \BOthers {.}}{%
Baranchuk%
\ \protect \BOthers {.}}{%
{\protect \APACyear {2022}}%
}]{%
baranchuk2022ddpmseg}
\APACinsertmetastar {%
baranchuk2022ddpmseg}%
\begin{APACrefauthors}%
Baranchuk, D.%
, Voynov, A.%
, Rubachev, I.%
, Khrulkov, V.%
\BCBL {} Babenko, A.%
\end{APACrefauthors}%
\unskip\
\newblock
\APACrefYearMonthDay{2022}{}{}.
\newblock
{\BBOQ}\APACrefatitle {Label-Efficient Semantic Segmentation with Diffusion Models} {Label-efficient semantic segmentation with diffusion models}.{\BBCQ}
\newblock
 \APACrefbtitle {Proceedings of the {I}nternational {C}onference on {L}earning {R}epresentations.} {Proceedings of the {I}nternational {C}onference on {L}earning {R}epresentations.}
\PrintBackRefs{\CurrentBib}

\bibitem [\protect \citeauthoryear {%
Beery%
\ \protect \BOthers {.}}{%
Beery%
\ \protect \BOthers {.}}{%
{\protect \APACyear {2018}}%
}]{%
beery2018recognition}
\APACinsertmetastar {%
beery2018recognition}%
\begin{APACrefauthors}%
Beery, S.%
, Van~Horn, G.%
\BCBL {} Perona, P.%
\end{APACrefauthors}%
\unskip\
\newblock
\APACrefYearMonthDay{2018}{}{}.
\newblock
{\BBOQ}\APACrefatitle {Recognition in terra incognita} {Recognition in terra incognita}.{\BBCQ}
\newblock
 \APACrefbtitle {ECCV} {Eccv}\ (\BPGS\ 456--473).
\PrintBackRefs{\CurrentBib}

\bibitem [\protect \citeauthoryear {%
Bolya%
\ \protect \BOthers {.}}{%
Bolya%
\ \protect \BOthers {.}}{%
{\protect \APACyear {2019}}%
}]{%
yolact}
\APACinsertmetastar {%
yolact}%
\begin{APACrefauthors}%
Bolya, D.%
, Zhou, C.%
, Xiao, F.%
\BCBL {} Lee, Y.J.%
\end{APACrefauthors}%
\unskip\
\newblock
\APACrefYearMonthDay{2019}{}{}.
\newblock
{\BBOQ}\APACrefatitle {Yolact: Real-time instance segmentation} {Yolact: Real-time instance segmentation}.{\BBCQ}
\newblock
 \APACrefbtitle {Proceedings of the {IEEE/CVF} {I}nternational {C}onference on {C}omputer {V}ision} {Proceedings of the {IEEE/CVF} {I}nternational {C}onference on {C}omputer {V}ision}\ (\BPGS\ 9157--9166).
\PrintBackRefs{\CurrentBib}

\bibitem [\protect \citeauthoryear {%
Cai%
\ \BBA {} Nuno%
}{%
Cai%
\ \BBA {} Nuno%
}{%
{\protect \APACyear {2019}}%
}]{%
cascadercnn}
\APACinsertmetastar {%
cascadercnn}%
\begin{APACrefauthors}%
Cai, Z.%
\BCBT {}\ \BBA {} Nuno, V.%
\end{APACrefauthors}%
\unskip\
\newblock
\APACrefYearMonthDay{2019}{}{}.
\newblock
{\BBOQ}\APACrefatitle {Cascade R-CNN: High quality object detection and instance segmentation} {Cascade r-cnn: High quality object detection and instance segmentation}.{\BBCQ}
\newblock
\APACjournalVolNumPages{{IEEE} {T}ransactions on {P}attern {A}nalysis and {M}achine {I}ntelligence}{43}{5}{1483--1498,}
\newblock

\newblock

\PrintBackRefs{\CurrentBib}

\bibitem [\protect \citeauthoryear {%
H.~Chen%
\ \protect \BOthers {.}}{%
H.~Chen%
\ \protect \BOthers {.}}{%
{\protect \APACyear {2020}}%
}]{%
blendmask}
\APACinsertmetastar {%
blendmask}%
\begin{APACrefauthors}%
Chen, H.%
, Sun, K.%
, Tian, Z.%
, Shen, C.%
, Huang, Y.%
\BCBL {} Yan, Y.%
\end{APACrefauthors}%
\unskip\
\newblock
\APACrefYearMonthDay{2020}{}{}.
\newblock
{\BBOQ}\APACrefatitle {Blendmask: Top-down meets bottom-up for instance segmentation} {Blendmask: Top-down meets bottom-up for instance segmentation}.{\BBCQ}
\newblock
 \APACrefbtitle {Proceedings of the {IEEE/CVF} {C}onference on {C}omputer {V}ision and {P}attern {R}ecognition} {Proceedings of the {IEEE/CVF} {C}onference on {C}omputer {V}ision and {P}attern {R}ecognition}\ (\BPGS\ 8573--8581).
\PrintBackRefs{\CurrentBib}

\bibitem [\protect \citeauthoryear {%
K.~Chen%
\ \protect \BOthers {.}}{%
K.~Chen%
\ \protect \BOthers {.}}{%
{\protect \APACyear {2019}}%
}]{%
htc}
\APACinsertmetastar {%
htc}%
\begin{APACrefauthors}%
Chen, K.%
, Pang, J.%
, Wang, J.%
, Xiong, Y.%
, Li, X.%
, Sun, S.%
\BDBL {}others%
\end{APACrefauthors}%
\unskip\
\newblock
\APACrefYearMonthDay{2019}{}{}.
\newblock
{\BBOQ}\APACrefatitle {Hybrid task cascade for instance segmentation} {Hybrid task cascade for instance segmentation}.{\BBCQ}
\newblock
 \APACrefbtitle {Proceedings of the {IEEE/CVF} {C}onference on {C}omputer {V}ision and {P}attern {R}ecognition} {Proceedings of the {IEEE/CVF} {C}onference on {C}omputer {V}ision and {P}attern {R}ecognition}\ (\BPGS\ 4974--4983).
\PrintBackRefs{\CurrentBib}

\bibitem [\protect \citeauthoryear {%
Cheng%
\ \protect \BOthers {.}}{%
Cheng%
\ \protect \BOthers {.}}{%
{\protect \APACyear {2022}}%
}]{%
mask2former}
\APACinsertmetastar {%
mask2former}%
\begin{APACrefauthors}%
Cheng, B.%
, Misra, I.%
, Schwing, A.G.%
, Kirillov, A.%
\BCBL {} Girdhar, R.%
\end{APACrefauthors}%
\unskip\
\newblock
\APACrefYearMonthDay{2022}{}{}.
\newblock
{\BBOQ}\APACrefatitle {Masked-attention mask transformer for universal image segmentation} {Masked-attention mask transformer for universal image segmentation}.{\BBCQ}
\newblock
 \APACrefbtitle {Proceedings of the {IEEE/CVF} {C}onference on {C}omputer {V}ision and {P}attern {R}ecognition} {Proceedings of the {IEEE/CVF} {C}onference on {C}omputer {V}ision and {P}attern {R}ecognition}\ (\BPGS\ 1290--1299).
\PrintBackRefs{\CurrentBib}

\bibitem [\protect \citeauthoryear {%
Cheng%
\ \protect \BOthers {.}}{%
Cheng%
\ \protect \BOthers {.}}{%
{\protect \APACyear {2021}}%
}]{%
maskformer}
\APACinsertmetastar {%
maskformer}%
\begin{APACrefauthors}%
Cheng, B.%
, Schwing, A.%
\BCBL {} Kirillov, A.%
\end{APACrefauthors}%
\unskip\
\newblock
\APACrefYearMonthDay{2021}{}{}.
\newblock
{\BBOQ}\APACrefatitle {Per-pixel classification is not all you need for semantic segmentation} {Per-pixel classification is not all you need for semantic segmentation}.{\BBCQ}
\newblock
\APACjournalVolNumPages{{A}dvances in {N}eural {I}nformation {P}rocessing {S}ystems}{34}{}{17864--17875,}
\newblock

\newblock

\PrintBackRefs{\CurrentBib}

\bibitem [\protect \citeauthoryear {%
Cordts%
\ \protect \BOthers {.}}{%
Cordts%
\ \protect \BOthers {.}}{%
{\protect \APACyear {2016}}%
}]{%
cityscapes}
\APACinsertmetastar {%
cityscapes}%
\begin{APACrefauthors}%
Cordts, M.%
, Omran, M.%
, Ramos, S.%
, Rehfeld, T.%
, Enzweiler, M.%
, Benenson, R.%
\BDBL {}Schiele, B.%
\end{APACrefauthors}%
\unskip\
\newblock
\APACrefYearMonthDay{2016}{}{}.
\newblock
{\BBOQ}\APACrefatitle {The Cityscapes Dataset for Semantic Urban Scene Understanding} {The cityscapes dataset for semantic urban scene understanding}.{\BBCQ}
\newblock
 \APACrefbtitle {Proceedings of the {IEEE/CVF} {C}onference on {C}omputer {V}ision and {P}attern {R}ecognition} {Proceedings of the {IEEE/CVF} {C}onference on {C}omputer {V}ision and {P}attern {R}ecognition}\ (\BPGS\ 3213--3223).
\PrintBackRefs{\CurrentBib}

\bibitem [\protect \citeauthoryear {%
Desai%
\ \BBA {} Johnson%
}{%
Desai%
\ \BBA {} Johnson%
}{%
{\protect \APACyear {2021}}%
}]{%
Desai_CVPR_2021}
\APACinsertmetastar {%
Desai_CVPR_2021}%
\begin{APACrefauthors}%
Desai, K.%
\BCBT {}\ \BBA {} Johnson, J.%
\end{APACrefauthors}%
\unskip\
\newblock
\APACrefYearMonthDay{2021}{}{}.
\newblock
{\BBOQ}\APACrefatitle {{VirTex}: Learning Visual Representations from Textual Annotations} {{VirTex}: Learning visual representations from textual annotations}.{\BBCQ}
\newblock
 \APACrefbtitle {Proceedings of the {IEEE/CVF} {C}onference on {C}omputer {V}ision and {P}attern {R}ecognition} {Proceedings of the {IEEE/CVF} {C}onference on {C}omputer {V}ision and {P}attern {R}ecognition}\ (\BPGS\ 11162--11173).
\PrintBackRefs{\CurrentBib}

\bibitem [\protect \citeauthoryear {%
Dhariwal%
\ \BBA {} Nichol%
}{%
Dhariwal%
\ \BBA {} Nichol%
}{%
{\protect \APACyear {2021}}%
}]{%
dhariwal2021diffusion}
\APACinsertmetastar {%
dhariwal2021diffusion}%
\begin{APACrefauthors}%
Dhariwal, P.%
\BCBT {}\ \BBA {} Nichol, A.Q.%
\end{APACrefauthors}%
\unskip\
\newblock
\APACrefYearMonthDay{2021}{}{}.
\newblock
{\BBOQ}\APACrefatitle {Diffusion Models Beat GANs on Image Synthesis} {Diffusion models beat gans on image synthesis}.{\BBCQ}
\newblock
 \APACrefbtitle {Proceedings of the {A}dvances in {N}eural {I}nformation {P}rocessing {S}ystems} {Proceedings of the {A}dvances in {N}eural {I}nformation {P}rocessing {S}ystems}\ (\BPGS\ 8780--8794).
\PrintBackRefs{\CurrentBib}

\bibitem [\protect \citeauthoryear {%
Ding%
\ \protect \BOthers {.}}{%
Ding%
\ \protect \BOthers {.}}{%
{\protect \APACyear {2023}}%
}]{%
MaskCLIP}
\APACinsertmetastar {%
MaskCLIP}%
\begin{APACrefauthors}%
Ding, Z.%
, Wang, J.%
\BCBL {} Tu, Z.%
\end{APACrefauthors}%
\unskip\
\newblock
\APACrefYearMonthDay{2023}{}{}.
\newblock
{\BBOQ}\APACrefatitle {Open-Vocabulary Universal Image Segmentation with MaskCLIP} {Open-vocabulary universal image segmentation with maskclip}.{\BBCQ}
\newblock
 \APACrefbtitle {Proceedings of the {I}nternational {C}onference on {M}achine {L}earning.} {Proceedings of the {I}nternational {C}onference on {M}achine {L}earning.}
\PrintBackRefs{\CurrentBib}

\bibitem [\protect \citeauthoryear {%
Dong%
\ \protect \BOthers {.}}{%
Dong%
\ \protect \BOthers {.}}{%
{\protect \APACyear {2024}}%
}]{%
dong2024unified}
\APACinsertmetastar {%
dong2024unified}%
\begin{APACrefauthors}%
Dong, B.%
, Pei, J.%
, Gao, R.%
, Xiang, T\BHBI Z.%
, Wang, S.%
\BCBL {} Xiong, H.%
\end{APACrefauthors}%
\unskip\
\newblock
\APACrefYearMonthDay{2024}{}{}.
\newblock
{\BBOQ}\APACrefatitle {A unified query-based paradigm for camouflaged instance segmentation} {A unified query-based paradigm for camouflaged instance segmentation}.{\BBCQ}
\newblock
 \APACrefbtitle {Proceedings of the ACM International Conference on Multimedia} {Proceedings of the acm international conference on multimedia}\ (\BPGS\ 2131--2138).
\PrintBackRefs{\CurrentBib}

\bibitem [\protect \citeauthoryear {%
Du%
\ \protect \BOthers {.}}{%
Du%
\ \protect \BOthers {.}}{%
{\protect \APACyear {2022}}%
}]{%
du2022learning}
\APACinsertmetastar {%
du2022learning}%
\begin{APACrefauthors}%
Du, Y.%
, Wei, F.%
, Zhang, Z.%
, Shi, M.%
, Gao, Y.%
\BCBL {} Li, G.%
\end{APACrefauthors}%
\unskip\
\newblock
\APACrefYearMonthDay{2022}{}{}.
\newblock
{\BBOQ}\APACrefatitle {Learning to Prompt for Open-Vocabulary Object Detection with Vision-Language Model} {Learning to prompt for open-vocabulary object detection with vision-language model}.{\BBCQ}
\newblock
 \APACrefbtitle {Proceedings of the {IEEE/CVF} {C}onference on {C}omputer {V}ision and {P}attern {R}ecognition} {Proceedings of the {IEEE/CVF} {C}onference on {C}omputer {V}ision and {P}attern {R}ecognition}\ (\BPGS\ 14064--14073).
\PrintBackRefs{\CurrentBib}

\bibitem [\protect \citeauthoryear {%
Esser%
\ \protect \BOthers {.}}{%
Esser%
\ \protect \BOthers {.}}{%
{\protect \APACyear {2021}}%
}]{%
esser2021taming}
\APACinsertmetastar {%
esser2021taming}%
\begin{APACrefauthors}%
Esser, P.%
, Rombach, R.%
\BCBL {} Ommer, B.%
\end{APACrefauthors}%
\unskip\
\newblock
\APACrefYearMonthDay{2021}{}{}.
\newblock
{\BBOQ}\APACrefatitle {Taming Transformers for High-Resolution Image Synthesis} {Taming transformers for high-resolution image synthesis}.{\BBCQ}
\newblock
 \APACrefbtitle {Proceedings of the {IEEE/CVF} {C}onference on {C}omputer {V}ision and {P}attern {R}ecognition} {Proceedings of the {IEEE/CVF} {C}onference on {C}omputer {V}ision and {P}attern {R}ecognition}\ (\BPGS\ 12873--12883).
\PrintBackRefs{\CurrentBib}

\bibitem [\protect \citeauthoryear {%
Fan%
\ \protect \BOthers {.}}{%
Fan%
\ \protect \BOthers {.}}{%
{\protect \APACyear {2022}}%
}]{%
fan2022concealed}
\APACinsertmetastar {%
fan2022concealed}%
\begin{APACrefauthors}%
Fan, D\BHBI P.%
, Ji, G\BHBI P.%
, Cheng, M\BHBI M.%
\BCBL {} Shao, L.%
\end{APACrefauthors}%
\unskip\
\newblock
\APACrefYearMonthDay{2022}{}{}.
\newblock
{\BBOQ}\APACrefatitle {Concealed Object Detection} {Concealed object detection}.{\BBCQ}
\newblock
\APACjournalVolNumPages{{IEEE} {T}ransactions on {P}attern {A}nalysis and {M}achine {I}ntelligence}{}{}{6024–-6042,}
\newblock

\newblock

\PrintBackRefs{\CurrentBib}

\bibitem [\protect \citeauthoryear {%
Fan%
\ \protect \BOthers {.}}{%
Fan%
\ \protect \BOthers {.}}{%
{\protect \APACyear {2020}}%
}]{%
Fan_CVPR_2020}
\APACinsertmetastar {%
Fan_CVPR_2020}%
\begin{APACrefauthors}%
Fan, D\BHBI P.%
, Ji, G\BHBI P.%
, Sun, G.%
, Cheng, M\BHBI M.%
, Shen, J.%
\BCBL {} Shao, L.%
\end{APACrefauthors}%
\unskip\
\newblock
\APACrefYearMonthDay{2020}{}{}.
\newblock
{\BBOQ}\APACrefatitle {Camouflaged Object Detection} {Camouflaged object detection}.{\BBCQ}
\newblock
 \APACrefbtitle {Proceedings of the {IEEE/CVF} {C}onference on {C}omputer {V}ision and {P}attern {R}ecognition} {Proceedings of the {IEEE/CVF} {C}onference on {C}omputer {V}ision and {P}attern {R}ecognition}\ (\BPGS\ 2777--2787).
\PrintBackRefs{\CurrentBib}

\bibitem [\protect \citeauthoryear {%
Fang%
\ \protect \BOthers {.}}{%
Fang%
\ \protect \BOthers {.}}{%
{\protect \APACyear {2021}}%
}]{%
queryinst}
\APACinsertmetastar {%
queryinst}%
\begin{APACrefauthors}%
Fang, Y.%
, Yang, S.%
, Wang, X.%
, Li, Y.%
, Fang, C.%
, Shan, Y.%
\BDBL {}Liu, W.%
\end{APACrefauthors}%
\unskip\
\newblock
\APACrefYearMonthDay{2021}{}{}.
\newblock
{\BBOQ}\APACrefatitle {Instances as queries} {Instances as queries}.{\BBCQ}
\newblock
 \APACrefbtitle {Proceedings of the {IEEE/CVF} {I}nternational {C}onference on {C}omputer {V}ision} {Proceedings of the {IEEE/CVF} {I}nternational {C}onference on {C}omputer {V}ision}\ (\BPGS\ 6910--6919).
\PrintBackRefs{\CurrentBib}

\bibitem [\protect \citeauthoryear {%
Fleming%
\ \protect \BOthers {.}}{%
Fleming%
\ \protect \BOthers {.}}{%
{\protect \APACyear {2014}}%
}]{%
Fleming2014CameraTW}
\APACinsertmetastar {%
Fleming2014CameraTW}%
\begin{APACrefauthors}%
Fleming, P.J.S.%
, Meek, P.D.%
, Ballard, G.%
, Banks, P.B.%
, Claridge, A.W.%
, Sanderson, J.G.%
\BCBL {} Swann, D.E.%
\end{APACrefauthors}%
\unskip\
\newblock
\APACrefYear{2014}.
\newblock
\APACrefbtitle {Camera Trapping: Wildlife Management and Research} {Camera trapping: Wildlife management and research}.
\newblock
\APACaddressPublisher{}{{CSIRO} {P}ublishing}.
\PrintBackRefs{\CurrentBib}

\bibitem [\protect \citeauthoryear {%
Fu%
\ \protect \BOthers {.}}{%
Fu%
\ \protect \BOthers {.}}{%
{\protect \APACyear {2014}}%
}]{%
fu2014learning}
\APACinsertmetastar {%
fu2014learning}%
\begin{APACrefauthors}%
Fu, R.%
, Guo, J.%
, Qin, B.%
, Che, W.%
, Wang, H.%
\BCBL {} Liu, T.%
\end{APACrefauthors}%
\unskip\
\newblock
\APACrefYearMonthDay{2014}{}{}.
\newblock
{\BBOQ}\APACrefatitle {Learning semantic hierarchies via word embeddings} {Learning semantic hierarchies via word embeddings}.{\BBCQ}
\newblock
 \APACrefbtitle {Proceedings of the 52nd Annual Meeting of the Association for Computational Linguistics (Volume 1: Long Papers)} {Proceedings of the 52nd annual meeting of the association for computational linguistics (volume 1: Long papers)}\ (\BPGS\ 1199--1209).
\PrintBackRefs{\CurrentBib}

\bibitem [\protect \citeauthoryear {%
Gal%
\ \protect \BOthers {.}}{%
Gal%
\ \protect \BOthers {.}}{%
{\protect \APACyear {2023}}%
}]{%
gal2023designing}
\APACinsertmetastar {%
gal2023designing}%
\begin{APACrefauthors}%
Gal, R.%
, Arar, M.%
, Atzmon, Y.%
, Bermano, A.H.%
, Chechik, G.%
\BCBL {} Cohen-Or, D.%
\end{APACrefauthors}%
\unskip\
\newblock
\APACrefYearMonthDay{2023}{}{}.
\newblock
{\BBOQ}\APACrefatitle {Encoder-Based Domain Tuning for Fast Personalization of Text-to-Image Models} {Encoder-based domain tuning for fast personalization of text-to-image models}.{\BBCQ}
\newblock
\APACjournalVolNumPages{{ACM Transactions on Graphics}}{42}{4}{1--13,}
\newblock

\newblock

\PrintBackRefs{\CurrentBib}

\bibitem [\protect \citeauthoryear {%
Gao%
\ \protect \BOthers {.}}{%
Gao%
\ \protect \BOthers {.}}{%
{\protect \APACyear {2022}}%
}]{%
gao2022open}
\APACinsertmetastar {%
gao2022open}%
\begin{APACrefauthors}%
Gao, M.%
, Xing, C.%
, Niebles, J.C.%
, Li, J.%
, Xu, R.%
, Liu, W.%
\BCBL {} Xiong, C.%
\end{APACrefauthors}%
\unskip\
\newblock
\APACrefYearMonthDay{2022}{}{}.
\newblock
{\BBOQ}\APACrefatitle {Open vocabulary object detection with pseudo bounding-box labels} {Open vocabulary object detection with pseudo bounding-box labels}.{\BBCQ}
\newblock
 \APACrefbtitle {Proceedings of the {E}uropean {C}onference on {C}omputer {V}ision} {Proceedings of the {E}uropean {C}onference on {C}omputer {V}ision}\ (\BPGS\ 266--282).
\PrintBackRefs{\CurrentBib}

\bibitem [\protect \citeauthoryear {%
Ghiasi%
\ \protect \BOthers {.}}{%
Ghiasi%
\ \protect \BOthers {.}}{%
{\protect \APACyear {2022}}%
}]{%
OpenSeg}
\APACinsertmetastar {%
OpenSeg}%
\begin{APACrefauthors}%
Ghiasi, G.%
, Gu, X.%
, Cui, Y.%
\BCBL {} Lin, T.%
\end{APACrefauthors}%
\unskip\
\newblock
\APACrefYearMonthDay{2022}{}{}.
\newblock
{\BBOQ}\APACrefatitle {Scaling Open-Vocabulary Image Segmentation with Image-Level Labels} {Scaling open-vocabulary image segmentation with image-level labels}.{\BBCQ}
\newblock
 \APACrefbtitle {Proceedings of the {E}uropean {C}onference on {C}omputer {V}ision} {Proceedings of the {E}uropean {C}onference on {C}omputer {V}ision}\ (\BPGS\ 540--557).
\PrintBackRefs{\CurrentBib}

\bibitem [\protect \citeauthoryear {%
Gu%
\ \protect \BOthers {.}}{%
Gu%
\ \protect \BOthers {.}}{%
{\protect \APACyear {2022}}%
}]{%
ViLD}
\APACinsertmetastar {%
ViLD}%
\begin{APACrefauthors}%
Gu, X.%
, Lin, T.%
, Kuo, W.%
\BCBL {} Cui, Y.%
\end{APACrefauthors}%
\unskip\
\newblock
\APACrefYearMonthDay{2022}{}{}.
\newblock
{\BBOQ}\APACrefatitle {Open-vocabulary Object Detection via Vision and Language Knowledge Distillation} {Open-vocabulary object detection via vision and language knowledge distillation}.{\BBCQ}
\newblock
 \APACrefbtitle {Proceedings of the {I}nternational {C}onference on {L}earning {R}epresentations.} {Proceedings of the {I}nternational {C}onference on {L}earning {R}epresentations.}
\PrintBackRefs{\CurrentBib}

\bibitem [\protect \citeauthoryear {%
Guo%
\ \protect \BOthers {.}}{%
Guo%
\ \protect \BOthers {.}}{%
{\protect \APACyear {2021}}%
}]{%
guo2021sotr}
\APACinsertmetastar {%
guo2021sotr}%
\begin{APACrefauthors}%
Guo, R.%
, Niu, D.%
, Qu, L.%
\BCBL {} Li, Z.%
\end{APACrefauthors}%
\unskip\
\newblock
\APACrefYearMonthDay{2021}{}{}.
\newblock
{\BBOQ}\APACrefatitle {Sotr: Segmenting objects with transformers} {Sotr: Segmenting objects with transformers}.{\BBCQ}
\newblock
 \APACrefbtitle {Proceedings of the {IEEE/CVF} {I}nternational {C}onference on {C}omputer {V}ision} {Proceedings of the {IEEE/CVF} {I}nternational {C}onference on {C}omputer {V}ision}\ (\BPGS\ 7157--7166).
\PrintBackRefs{\CurrentBib}

\bibitem [\protect \citeauthoryear {%
C.~He%
\ \protect \BOthers {.}}{%
C.~He%
\ \protect \BOthers {.}}{%
{\protect \APACyear {2023}}%
}]{%
He_CVPR_2023}
\APACinsertmetastar {%
He_CVPR_2023}%
\begin{APACrefauthors}%
He, C.%
, Li, K.%
, Zhang, Y.%
, Tang, L.%
, Zhang, Y.%
, Guo, Z.%
\BCBL {} Li, X.%
\end{APACrefauthors}%
\unskip\
\newblock
\APACrefYearMonthDay{2023}{}{}.
\newblock
{\BBOQ}\APACrefatitle {Camouflaged Object Detection with Feature Decomposition and Edge Reconstruction} {Camouflaged object detection with feature decomposition and edge reconstruction}.{\BBCQ}
\newblock
 \APACrefbtitle {Proceedings of the {IEEE/CVF} {C}onference on {C}omputer {V}ision and {P}attern {R}ecognition} {Proceedings of the {IEEE/CVF} {C}onference on {C}omputer {V}ision and {P}attern {R}ecognition}\ (\BPGS\ 22046--22055).
\PrintBackRefs{\CurrentBib}

\bibitem [\protect \citeauthoryear {%
K.~He%
\ \protect \BOthers {.}}{%
K.~He%
\ \protect \BOthers {.}}{%
{\protect \APACyear {2017}}%
}]{%
maskrcnn}
\APACinsertmetastar {%
maskrcnn}%
\begin{APACrefauthors}%
He, K.%
, Gkioxari, G.%
, Doll{\'{a}}r, P.%
\BCBL {} Girshick, R.B.%
\end{APACrefauthors}%
\unskip\
\newblock
\APACrefYearMonthDay{2017}{}{}.
\newblock
{\BBOQ}\APACrefatitle {Mask {R-CNN}} {Mask {R-CNN}}.{\BBCQ}
\newblock
 \APACrefbtitle {Proceedings of the {IEEE/CVF {I}nternational {C}onference on {C}omputer {V}ision}} {Proceedings of the {IEEE/CVF {I}nternational {C}onference on {C}omputer {V}ision}}\ (\BPGS\ 2980--2988).
\PrintBackRefs{\CurrentBib}

\bibitem [\protect \citeauthoryear {%
Z.~He%
\ \protect \BOthers {.}}{%
Z.~He%
\ \protect \BOthers {.}}{%
{\protect \APACyear {2024}}%
}]{%
he2024text}
\APACinsertmetastar {%
he2024text}%
\begin{APACrefauthors}%
He, Z.%
, Xia, C.%
, Qiao, S.%
\BCBL {} Li, J.%
\end{APACrefauthors}%
\unskip\
\newblock
\APACrefYearMonthDay{2024}{}{}.
\newblock
{\BBOQ}\APACrefatitle {Text-prompt Camouflaged Instance Segmentation with Graduated Camouflage Learning} {Text-prompt camouflaged instance segmentation with graduated camouflage learning}.{\BBCQ}
\newblock
 \APACrefbtitle {Proceedings of the ACM International Conference on Multimedia} {Proceedings of the acm international conference on multimedia}\ (\BPGS\ 5584--5593).
\PrintBackRefs{\CurrentBib}

\bibitem [\protect \citeauthoryear {%
Hertz%
\ \protect \BOthers {.}}{%
Hertz%
\ \protect \BOthers {.}}{%
{\protect \APACyear {2023}}%
}]{%
hertz2023prompt}
\APACinsertmetastar {%
hertz2023prompt}%
\begin{APACrefauthors}%
Hertz, A.%
, Mokady, R.%
, Tenenbaum, J.%
, Aberman, K.%
, Pritch, Y.%
\BCBL {} Cohen-Or, D.%
\end{APACrefauthors}%
\unskip\
\newblock
\APACrefYearMonthDay{2023}{}{}.
\newblock
{\BBOQ}\APACrefatitle {Prompt-to-prompt image editing with cross attention control} {Prompt-to-prompt image editing with cross attention control}.{\BBCQ}
\newblock
 \APACrefbtitle {Proceedings of the {I}nternational {C}onference on {L}earning {R}epresentations.} {Proceedings of the {I}nternational {C}onference on {L}earning {R}epresentations.}
\PrintBackRefs{\CurrentBib}

\bibitem [\protect \citeauthoryear {%
Ho%
\ \protect \BOthers {.}}{%
Ho%
\ \protect \BOthers {.}}{%
{\protect \APACyear {2020}}%
}]{%
ho2020denoising}
\APACinsertmetastar {%
ho2020denoising}%
\begin{APACrefauthors}%
Ho, J.%
, Jain, A.%
\BCBL {} Abbeel, P.%
\end{APACrefauthors}%
\unskip\
\newblock
\APACrefYearMonthDay{2020}{}{}.
\newblock
{\BBOQ}\APACrefatitle {Denoising Diffusion Probabilistic Models} {Denoising diffusion probabilistic models}.{\BBCQ}
\newblock
 \APACrefbtitle {Proceedings of the {A}dvances in {N}eural {I}nformation {P}rocessing {S}ystems} {Proceedings of the {A}dvances in {N}eural {I}nformation {P}rocessing {S}ystems}\ (\BPGS\ 6840--6851).
\PrintBackRefs{\CurrentBib}

\bibitem [\protect \citeauthoryear {%
X.~Huang%
\ \BBA {} Belongie%
}{%
X.~Huang%
\ \BBA {} Belongie%
}{%
{\protect \APACyear {2017}}%
}]{%
huang2017adain}
\APACinsertmetastar {%
huang2017adain}%
\begin{APACrefauthors}%
Huang, X.%
\BCBT {}\ \BBA {} Belongie, S.J.%
\end{APACrefauthors}%
\unskip\
\newblock
\APACrefYearMonthDay{2017}{}{}.
\newblock
{\BBOQ}\APACrefatitle {Arbitrary Style Transfer in Real-Time with Adaptive Instance Normalization} {Arbitrary style transfer in real-time with adaptive instance normalization}.{\BBCQ}
\newblock
 \APACrefbtitle {Proceedings of the {IEEE/CVF} {I}nternational {C}onference on {C}omputer {V}ision} {Proceedings of the {IEEE/CVF} {I}nternational {C}onference on {C}omputer {V}ision}\ (\BPGS\ 1510--1519).
\PrintBackRefs{\CurrentBib}

\bibitem [\protect \citeauthoryear {%
Z.~Huang%
\ \protect \BOthers {.}}{%
Z.~Huang%
\ \protect \BOthers {.}}{%
{\protect \APACyear {2019}}%
}]{%
msrcnn}
\APACinsertmetastar {%
msrcnn}%
\begin{APACrefauthors}%
Huang, Z.%
, Huang, L.%
, Gong, Y.%
, Huang, C.%
\BCBL {} Wang, X.%
\end{APACrefauthors}%
\unskip\
\newblock
\APACrefYearMonthDay{2019}{}{}.
\newblock
{\BBOQ}\APACrefatitle {Mask scoring r-cnn} {Mask scoring r-cnn}.{\BBCQ}
\newblock
 \APACrefbtitle {Proceedings of the {IEEE/CVF} {C}onference on {C}omputer {V}ision and {P}attern {R}ecognition} {Proceedings of the {IEEE/CVF} {C}onference on {C}omputer {V}ision and {P}attern {R}ecognition}\ (\BPGS\ 6409--6418).
\PrintBackRefs{\CurrentBib}

\bibitem [\protect \citeauthoryear {%
Ike%
\ \protect \BOthers {.}}{%
Ike%
\ \protect \BOthers {.}}{%
{\protect \APACyear {2024}}%
}]{%
DiCANet}
\APACinsertmetastar {%
DiCANet}%
\begin{APACrefauthors}%
Ike, C.S.%
, Muhammad, N.%
, Bibi, N.%
, Alhazmi, S.%
\BCBL {} Eoghan, F.%
\end{APACrefauthors}%
\unskip\
\newblock
\APACrefYearMonthDay{2024}{}{}.
\newblock
{\BBOQ}\APACrefatitle {Discriminative context-aware network for camouflaged object detection} {Discriminative context-aware network for camouflaged object detection}.{\BBCQ}
\newblock
\APACjournalVolNumPages{Frontiers in Artificial Intelligence}{}{}{,}
\newblock
\begin{APACrefDOI} \doi{10.3389/frai.2024.1347898} \end{APACrefDOI}
\newblock

\newblock

\PrintBackRefs{\CurrentBib}

\bibitem [\protect \citeauthoryear {%
Jamali%
\ \protect \BOthers {.}}{%
Jamali%
\ \protect \BOthers {.}}{%
{\protect \APACyear {2025}}%
}]{%
jamali2025context}
\APACinsertmetastar {%
jamali2025context}%
\begin{APACrefauthors}%
Jamali, M.%
, Davidsson, P.%
, Khoshkangini, R.%
, Ljungqvist, M.G.%
\BCBL {} Mihailescu, R\BHBI C.%
\end{APACrefauthors}%
\unskip\
\newblock
\APACrefYearMonthDay{2025}{}{}.
\newblock
{\BBOQ}\APACrefatitle {Context in object detection: a systematic literature review} {Context in object detection: a systematic literature review}.{\BBCQ}
\newblock
\APACjournalVolNumPages{Artificial Intelligence Review}{}{}{,}
\newblock

\newblock

\PrintBackRefs{\CurrentBib}

\bibitem [\protect \citeauthoryear {%
Jia%
\ \protect \BOthers {.}}{%
Jia%
\ \protect \BOthers {.}}{%
{\protect \APACyear {2021}}%
}]{%
ALIGN}
\APACinsertmetastar {%
ALIGN}%
\begin{APACrefauthors}%
Jia, C.%
, Yang, Y.%
, Xia, Y.%
, Chen, Y.%
, Parekh, Z.%
, Pham, H.%
\BDBL {}Duerig, T.%
\end{APACrefauthors}%
\unskip\
\newblock
\APACrefYearMonthDay{2021}{}{}.
\newblock
{\BBOQ}\APACrefatitle {Scaling up visual and vision-language representation learning with noisy text supervision} {Scaling up visual and vision-language representation learning with noisy text supervision}.{\BBCQ}
\newblock
 \APACrefbtitle {Proceedings of the {I}nternational {C}onference on {M}achine {L}earning} {Proceedings of the {I}nternational {C}onference on {M}achine {L}earning}\ (\BPGS\ 4904--4916).
\PrintBackRefs{\CurrentBib}

\bibitem [\protect \citeauthoryear {%
Karras%
\ \protect \BOthers {.}}{%
Karras%
\ \protect \BOthers {.}}{%
{\protect \APACyear {2020}}%
}]{%
karras2020styleganv2}
\APACinsertmetastar {%
karras2020styleganv2}%
\begin{APACrefauthors}%
Karras, T.%
, Laine, S.%
, Aittala, M.%
, Hellsten, J.%
, Lehtinen, J.%
\BCBL {} Aila, T.%
\end{APACrefauthors}%
\unskip\
\newblock
\APACrefYearMonthDay{2020}{}{}.
\newblock
{\BBOQ}\APACrefatitle {Analyzing and Improving the Image Quality of StyleGAN} {Analyzing and improving the image quality of stylegan}.{\BBCQ}
\newblock
 \APACrefbtitle {Proceedings of the {IEEE/CVF} {C}onference on {C}omputer {V}ision and {P}attern {R}ecognition} {Proceedings of the {IEEE/CVF} {C}onference on {C}omputer {V}ision and {P}attern {R}ecognition}\ (\BPGS\ 8107--8116).
\PrintBackRefs{\CurrentBib}

\bibitem [\protect \citeauthoryear {%
Ke%
\ \protect \BOthers {.}}{%
Ke%
\ \protect \BOthers {.}}{%
{\protect \APACyear {2022}}%
}]{%
masktransfiner}
\APACinsertmetastar {%
masktransfiner}%
\begin{APACrefauthors}%
Ke, L.%
, Danelljan, M.%
, Li, X.%
, Tai, Y\BHBI W.%
, Tang, C\BHBI K.%
\BCBL {} Yu, F.%
\end{APACrefauthors}%
\unskip\
\newblock
\APACrefYearMonthDay{2022}{}{}.
\newblock
{\BBOQ}\APACrefatitle {Mask transfiner for high-quality instance segmentation} {Mask transfiner for high-quality instance segmentation}.{\BBCQ}
\newblock
 \APACrefbtitle {Proceedings of the {IEEE/CVF} {C}onference on {C}omputer {V}ision and {P}attern {R}ecognition} {Proceedings of the {IEEE/CVF} {C}onference on {C}omputer {V}ision and {P}attern {R}ecognition}\ (\BPGS\ 4412--4421).
\PrintBackRefs{\CurrentBib}

\bibitem [\protect \citeauthoryear {%
Khan%
\ \protect \BOthers {.}}{%
Khan%
\ \protect \BOthers {.}}{%
{\protect \APACyear {2024}}%
}]{%
Khan_2024_WACV}
\APACinsertmetastar {%
Khan_2024_WACV}%
\begin{APACrefauthors}%
Khan, A.%
, Khan, M.%
, Gueaieb, W.%
, El~Saddik, A.%
, De~Masi, G.%
\BCBL {} Karray, F.%
\end{APACrefauthors}%
\unskip\
\newblock
\APACrefYearMonthDay{2024}{January}{}.
\newblock
{\BBOQ}\APACrefatitle {CamoFocus: Enhancing Camouflage Object Detection With Split-Feature Focal Modulation and Context Refinement} {Camofocus: Enhancing camouflage object detection with split-feature focal modulation and context refinement}.{\BBCQ}
\newblock
 \APACrefbtitle {Proceedings of the IEEE/CVF Winter Conference on Applications of Computer Vision ({WACV})} {Proceedings of the ieee/cvf winter conference on applications of computer vision ({WACV})}\ (\BPGS\ 1434--1443).
\PrintBackRefs{\CurrentBib}

\bibitem [\protect \citeauthoryear {%
Kuhn%
}{%
Kuhn%
}{%
{\protect \APACyear {1955}}%
}]{%
hungarian}
\APACinsertmetastar {%
hungarian}%
\begin{APACrefauthors}%
Kuhn, H.W.%
\end{APACrefauthors}%
\unskip\
\newblock
\APACrefYearMonthDay{1955}{March}{}.
\newblock
{\BBOQ}\APACrefatitle {{The Hungarian Method for the Assignment Problem}} {{The Hungarian Method for the Assignment Problem}}.{\BBCQ}
\newblock
\APACjournalVolNumPages{{N}aval {R}esearch {L}ogistics {Q}uarterly}{2}{1}{83--97,}
\newblock

\newblock

\PrintBackRefs{\CurrentBib}

\bibitem [\protect \citeauthoryear {%
Kumari%
\ \protect \BOthers {.}}{%
Kumari%
\ \protect \BOthers {.}}{%
{\protect \APACyear {2023}}%
}]{%
kumari2023multi}
\APACinsertmetastar {%
kumari2023multi}%
\begin{APACrefauthors}%
Kumari, N.%
, Zhang, B.%
, Zhang, R.%
, Shechtman, E.%
\BCBL {} Zhu, J\BHBI Y.%
\end{APACrefauthors}%
\unskip\
\newblock
\APACrefYearMonthDay{2023}{}{}.
\newblock
{\BBOQ}\APACrefatitle {Multi-Concept Customization of Text-to-Image Diffusion} {Multi-concept customization of text-to-image diffusion}.{\BBCQ}
\newblock
 \APACrefbtitle {Proceedings of the {IEEE/CVF} {C}onference on {C}omputer {V}ision and {P}attern {R}ecognition} {Proceedings of the {IEEE/CVF} {C}onference on {C}omputer {V}ision and {P}attern {R}ecognition}\ (\BPGS\ 1931--1941).
\PrintBackRefs{\CurrentBib}

\bibitem [\protect \citeauthoryear {%
Kuo%
\ \protect \BOthers {.}}{%
Kuo%
\ \protect \BOthers {.}}{%
{\protect \APACyear {2023}}%
}]{%
kuo2023f}
\APACinsertmetastar {%
kuo2023f}%
\begin{APACrefauthors}%
Kuo, W.%
, Cui, Y.%
, Gu, X.%
, Piergiovanni, A.%
\BCBL {} Angelova, A.%
\end{APACrefauthors}%
\unskip\
\newblock
\APACrefYearMonthDay{2023}{}{}.
\newblock
{\BBOQ}\APACrefatitle {F-VLM: Open-Vocabulary Object Detection upon Frozen Vision and Language Models} {F-vlm: Open-vocabulary object detection upon frozen vision and language models}.{\BBCQ}
\newblock
 \APACrefbtitle {Proceedings of the {I}nternational {C}onference on {L}earning {R}epresentations.} {Proceedings of the {I}nternational {C}onference on {L}earning {R}epresentations.}
\PrintBackRefs{\CurrentBib}

\bibitem [\protect \citeauthoryear {%
M\BHBI Q.~Le%
\ \protect \BOthers {.}}{%
M\BHBI Q.~Le%
\ \protect \BOthers {.}}{%
{\protect \APACyear {2025}}%
}]{%
CamoFA}
\APACinsertmetastar {%
CamoFA}%
\begin{APACrefauthors}%
Le, M\BHBI Q.%
, Tran, M\BHBI T.%
, Le, T\BHBI N.%
, Nguyen, T.V.%
\BCBL {} Do, T\BHBI T.%
\end{APACrefauthors}%
\unskip\
\newblock
\APACrefYearMonthDay{2025}{}{}.
\newblock
{\BBOQ}\APACrefatitle {{ CamoFA: A Learnable Fourier-Based Augmentation for Camouflage Segmentation }} {{ CamoFA: A Learnable Fourier-Based Augmentation for Camouflage Segmentation }}.{\BBCQ}
\newblock
 \APACrefbtitle {2025 IEEE/CVF Winter Conference on Applications of Computer Vision (WACV)} {2025 ieee/cvf winter conference on applications of computer vision (wacv)}\ (\BPGS\ 3427--3436).
\PrintBackRefs{\CurrentBib}

\bibitem [\protect \citeauthoryear {%
T\BHBI N.~Le%
\ \protect \BOthers {.}}{%
T\BHBI N.~Le%
\ \protect \BOthers {.}}{%
{\protect \APACyear {2019}}%
}]{%
CAMO}
\APACinsertmetastar {%
CAMO}%
\begin{APACrefauthors}%
Le, T\BHBI N.%
, Nguyen, T.V.%
, Nie, Z.%
, Tran, M\BHBI T.%
\BCBL {} Sugimoto, A.%
\end{APACrefauthors}%
\unskip\
\newblock
\APACrefYearMonthDay{2019}{}{}.
\newblock
{\BBOQ}\APACrefatitle {Anabranch network for camouflaged object segmentation} {Anabranch network for camouflaged object segmentation}.{\BBCQ}
\newblock
\APACjournalVolNumPages{Computer Vision and Image Understanding}{184}{}{45--56,}
\newblock

\newblock

\PrintBackRefs{\CurrentBib}

\bibitem [\protect \citeauthoryear {%
C.~Li%
\ \protect \BOthers {.}}{%
C.~Li%
\ \protect \BOthers {.}}{%
{\protect \APACyear {2024}}%
}]{%
li2024multi}
\APACinsertmetastar {%
li2024multi}%
\begin{APACrefauthors}%
Li, C.%
, Jiao, G.%
, Yue, G.%
, He, R.%
\BCBL {} Huang, J.%
\end{APACrefauthors}%
\unskip\
\newblock
\APACrefYearMonthDay{2024}{}{}.
\newblock
{\BBOQ}\APACrefatitle {Multi-scale pooling learning for camouflaged instance segmentation} {Multi-scale pooling learning for camouflaged instance segmentation}.{\BBCQ}
\newblock
\APACjournalVolNumPages{Applied Intelligence}{54}{5}{4062--4076,}
\newblock

\newblock

\PrintBackRefs{\CurrentBib}

\bibitem [\protect \citeauthoryear {%
D.~Li%
\ \protect \BOthers {.}}{%
D.~Li%
\ \protect \BOthers {.}}{%
{\protect \APACyear {2022}}%
}]{%
li2022bigdatasetgan}
\APACinsertmetastar {%
li2022bigdatasetgan}%
\begin{APACrefauthors}%
Li, D.%
, Ling, H.%
, Kim, S.W.%
, Kreis, K.%
, Fidler, S.%
\BCBL {} Torralba, A.%
\end{APACrefauthors}%
\unskip\
\newblock
\APACrefYearMonthDay{2022}{}{}.
\newblock
{\BBOQ}\APACrefatitle {BigDatasetGAN: Synthesizing ImageNet with Pixel-wise Annotations} {Bigdatasetgan: Synthesizing imagenet with pixel-wise annotations}.{\BBCQ}
\newblock
 \APACrefbtitle {Proceedings of the {IEEE/CVF} {C}onference on {C}omputer {V}ision and {P}attern {R}ecognition} {Proceedings of the {IEEE/CVF} {C}onference on {C}omputer {V}ision and {P}attern {R}ecognition}\ (\BPGS\ 21298--21308).
\PrintBackRefs{\CurrentBib}

\bibitem [\protect \citeauthoryear {%
Lin%
\ \protect \BOthers {.}}{%
Lin%
\ \protect \BOthers {.}}{%
{\protect \APACyear {2014}}%
}]{%
coco}
\APACinsertmetastar {%
coco}%
\begin{APACrefauthors}%
Lin, T.%
, Maire, M.%
, Belongie, S.J.%
, Hays, J.%
, Perona, P.%
, Ramanan, D.%
\BDBL {}Zitnick, C.L.%
\end{APACrefauthors}%
\unskip\
\newblock
\APACrefYearMonthDay{2014}{}{}.
\newblock
{\BBOQ}\APACrefatitle {Microsoft {COCO:} Common Objects in Context} {Microsoft {COCO:} common objects in context}.{\BBCQ}
\newblock
 \APACrefbtitle {Proceedings of the {E}uropean {C}onference on {C}omputer {V}ision} {Proceedings of the {E}uropean {C}onference on {C}omputer {V}ision}\ (\BPGS\ 740--755).
\PrintBackRefs{\CurrentBib}

\bibitem [\protect \citeauthoryear {%
Liu%
\ \BBA {} Di%
}{%
Liu%
\ \BBA {} Di%
}{%
{\protect \APACyear {2023}}%
}]{%
Liu_neuro_2023}
\APACinsertmetastar {%
Liu_neuro_2023}%
\begin{APACrefauthors}%
Liu, M.%
\BCBT {}\ \BBA {} Di, X.%
\end{APACrefauthors}%
\unskip\
\newblock
\APACrefYearMonthDay{2023}{}{}.
\newblock
{\BBOQ}\APACrefatitle {Extraordinary {MHNet}: Military high-level camouflage object detection network and dataset} {Extraordinary {MHNet}: Military high-level camouflage object detection network and dataset}.{\BBCQ}
\newblock
\APACjournalVolNumPages{Neurocomputing}{549}{}{126466,}
\newblock

\newblock

\PrintBackRefs{\CurrentBib}

\bibitem [\protect \citeauthoryear {%
Loshchilov%
\ \BBA {} Hutter%
}{%
Loshchilov%
\ \BBA {} Hutter%
}{%
{\protect \APACyear {2019}}%
}]{%
adamw}
\APACinsertmetastar {%
adamw}%
\begin{APACrefauthors}%
Loshchilov, I.%
\BCBT {}\ \BBA {} Hutter, F.%
\end{APACrefauthors}%
\unskip\
\newblock
\APACrefYearMonthDay{2019}{}{}.
\newblock
{\BBOQ}\APACrefatitle {Decoupled Weight Decay Regularization} {Decoupled weight decay regularization}.{\BBCQ}
\newblock
 \APACrefbtitle {Proceedings of the {I}nternational {C}onference on {L}earning {R}epresentations.} {Proceedings of the {I}nternational {C}onference on {L}earning {R}epresentations.}
\PrintBackRefs{\CurrentBib}

\bibitem [\protect \citeauthoryear {%
Luo%
\ \protect \BOthers {.}}{%
Luo%
\ \protect \BOthers {.}}{%
{\protect \APACyear {2023}}%
}]{%
dcnet}
\APACinsertmetastar {%
dcnet}%
\begin{APACrefauthors}%
Luo, N.%
, Pan, Y.%
, Sun, R.%
, Zhang, T.%
, Xiong, Z.%
\BCBL {} Wu, F.%
\end{APACrefauthors}%
\unskip\
\newblock
\APACrefYearMonthDay{2023}{}{}.
\newblock
{\BBOQ}\APACrefatitle {Camouflaged Instance Segmentation via Explicit De-Camouflaging} {Camouflaged instance segmentation via explicit de-camouflaging}.{\BBCQ}
\newblock
 \APACrefbtitle {Proceedings of the {IEEE/CVF} {C}onference on {C}omputer {V}ision and {P}attern {R}ecognition} {Proceedings of the {IEEE/CVF} {C}onference on {C}omputer {V}ision and {P}attern {R}ecognition}\ (\BPGS\ 17918--17927).
\PrintBackRefs{\CurrentBib}

\bibitem [\protect \citeauthoryear {%
Lyu%
\ \protect \BOthers {.}}{%
Lyu%
\ \protect \BOthers {.}}{%
{\protect \APACyear {2021}}%
}]{%
yunqiu_cod21}
\APACinsertmetastar {%
yunqiu_cod21}%
\begin{APACrefauthors}%
Lyu, Y.%
, Zhang, J.%
, Dai, Y.%
, Li, A.%
, Liu, B.%
, Barnes, N.%
\BCBL {} Fan, D\BHBI P.%
\end{APACrefauthors}%
\unskip\
\newblock
\APACrefYearMonthDay{2021}{}{}.
\newblock
{\BBOQ}\APACrefatitle {Simultaneously Localize, Segment and Rank the Camouflaged Objects} {Simultaneously localize, segment and rank the camouflaged objects}.{\BBCQ}
\newblock
 \APACrefbtitle {Proceedings of the {IEEE/CVF} {C}onference on {C}omputer {V}ision and {P}attern {R}ecognition} {Proceedings of the {IEEE/CVF} {C}onference on {C}omputer {V}ision and {P}attern {R}ecognition}\ (\BPGS\ 11586--11596).
\PrintBackRefs{\CurrentBib}

\bibitem [\protect \citeauthoryear {%
Milletari%
\ \protect \BOthers {.}}{%
Milletari%
\ \protect \BOthers {.}}{%
{\protect \APACyear {2016}}%
}]{%
dice2016}
\APACinsertmetastar {%
dice2016}%
\begin{APACrefauthors}%
Milletari, F.%
, Navab, N.%
\BCBL {} Ahmadi, S\BHBI A.%
\end{APACrefauthors}%
\unskip\
\newblock
\APACrefYearMonthDay{2016}{}{}.
\newblock
{\BBOQ}\APACrefatitle {{V-Net}: Fully Convolutional Neural Networks for Volumetric Medical Image Segmentation} {{V-Net}: Fully convolutional neural networks for volumetric medical image segmentation}.{\BBCQ}
\newblock
 \APACrefbtitle {Proceedings of the {I}nternational {C}onference on {3D} {V}ision} {Proceedings of the {I}nternational {C}onference on {3D} {V}ision}\ (\BPGS\ 565--571).
\PrintBackRefs{\CurrentBib}

\bibitem [\protect \citeauthoryear {%
Minderer%
\ \protect \BOthers {.}}{%
Minderer%
\ \protect \BOthers {.}}{%
{\protect \APACyear {2022}}%
}]{%
minderer2022simple}
\APACinsertmetastar {%
minderer2022simple}%
\begin{APACrefauthors}%
Minderer, M.%
, Gritsenko, A.A.%
, Stone, A.%
, Neumann, M.%
, Weissenborn, D.%
, Dosovitskiy, A.%
\BDBL {}Houlsby, N.%
\end{APACrefauthors}%
\unskip\
\newblock
\APACrefYearMonthDay{2022}{}{}.
\newblock
{\BBOQ}\APACrefatitle {Simple open-vocabulary object detection with vision transformers} {Simple open-vocabulary object detection with vision transformers}.{\BBCQ}
\newblock
 \APACrefbtitle {Proceedings of the {E}uropean {C}onference on {C}omputer {V}ision} {Proceedings of the {E}uropean {C}onference on {C}omputer {V}ision}\ (\BPGS\ 728--755).
\PrintBackRefs{\CurrentBib}

\bibitem [\protect \citeauthoryear {%
Mokady%
\ \protect \BOthers {.}}{%
Mokady%
\ \protect \BOthers {.}}{%
{\protect \APACyear {2021}}%
}]{%
mokady2021clipcap}
\APACinsertmetastar {%
mokady2021clipcap}%
\begin{APACrefauthors}%
Mokady, R.%
, Hertz, A.%
\BCBL {} Bermano, A.H.%
\end{APACrefauthors}%
\unskip\
\newblock
\APACrefYearMonthDay{2021}{}{}.
\newblock
{\BBOQ}\APACrefatitle {ClipCap: CLIP Prefix for Image Captioning} {Clipcap: Clip prefix for image captioning}.{\BBCQ}
\newblock
\APACjournalVolNumPages{arXiv preprint arXiv:2111.09734}{}{}{,}
\newblock

\newblock

\PrintBackRefs{\CurrentBib}

\bibitem [\protect \citeauthoryear {%
Nguyen%
\ \protect \BOthers {.}}{%
Nguyen%
\ \protect \BOthers {.}}{%
{\protect \APACyear {2023}}%
}]{%
Nguyen_MTAP_2023}
\APACinsertmetastar {%
Nguyen_MTAP_2023}%
\begin{APACrefauthors}%
Nguyen, T.T.T.%
, Eichholtzer, A.C.%
, Driscoll, D.A.%
, Semianiw, N.I.%
, Corva, D.M.%
, Kouzani, A.Z.%
\BDBL {}Nguyen, D.T.%
\end{APACrefauthors}%
\unskip\
\newblock
\APACrefYearMonthDay{2023}{}{}.
\newblock
{\BBOQ}\APACrefatitle {SAWIT: A small-sized animal wild image dataset with annotations} {Sawit: A small-sized animal wild image dataset with annotations}.{\BBCQ}
\newblock
\APACjournalVolNumPages{{Multimedia Tools and Applications}}{}{}{1--26,}
\newblock

\newblock

\PrintBackRefs{\CurrentBib}

\bibitem [\protect \citeauthoryear {%
Nichol%
\ \protect \BOthers {.}}{%
Nichol%
\ \protect \BOthers {.}}{%
{\protect \APACyear {2022}}%
}]{%
nichol22glide}
\APACinsertmetastar {%
nichol22glide}%
\begin{APACrefauthors}%
Nichol, A.Q.%
, Dhariwal, P.%
, Ramesh, A.%
, Shyam, P.%
, Mishkin, P.%
, McGrew, B.%
\BDBL {}Chen, M.%
\end{APACrefauthors}%
\unskip\
\newblock
\APACrefYearMonthDay{2022}{}{}.
\newblock
{\BBOQ}\APACrefatitle {{GLIDE:} Towards Photorealistic Image Generation and Editing with Text-Guided Diffusion Models} {{GLIDE:} towards photorealistic image generation and editing with text-guided diffusion models}.{\BBCQ}
\newblock
 \APACrefbtitle {Proceedings of the {I}nternational {C}onference on {M}achine {L}earning} {Proceedings of the {I}nternational {C}onference on {M}achine {L}earning}\ (\BPGS\ 16784--16804).
\PrintBackRefs{\CurrentBib}

\bibitem [\protect \citeauthoryear {%
Norouzzadeh%
\ \protect \BOthers {.}}{%
Norouzzadeh%
\ \protect \BOthers {.}}{%
{\protect \APACyear {2018}}%
}]{%
norouzzadeh2018automatically}
\APACinsertmetastar {%
norouzzadeh2018automatically}%
\begin{APACrefauthors}%
Norouzzadeh, M.S.%
, Nguyen, A.%
, Kosmala, M.%
, Swanson, A.%
, Palmer, M.S.%
, Packer, C.%
\BCBL {} Clune, J.%
\end{APACrefauthors}%
\unskip\
\newblock
\APACrefYearMonthDay{2018}{}{}.
\newblock
{\BBOQ}\APACrefatitle {Automatically identifying, counting, and describing wild animals in camera-trap images with deep learning} {Automatically identifying, counting, and describing wild animals in camera-trap images with deep learning}.{\BBCQ}
\newblock
\APACjournalVolNumPages{PNAS}{115}{25}{E5716--E5725,}
\newblock

\newblock

\PrintBackRefs{\CurrentBib}

\bibitem [\protect \citeauthoryear {%
Pang%
\ \protect \BOthers {.}}{%
Pang%
\ \protect \BOthers {.}}{%
{\protect \APACyear {2024}}%
}]{%
OVCOS_ECCV2024}
\APACinsertmetastar {%
OVCOS_ECCV2024}%
\begin{APACrefauthors}%
Pang, Y.%
, Zhao, X.%
, Zuo, J.%
, Zhang, L.%
\BCBL {} Lu, H.%
\end{APACrefauthors}%
\unskip\
\newblock
\APACrefYearMonthDay{2024}{}{}.
\newblock
{\BBOQ}\APACrefatitle {Open-Vocabulary Camouflaged Object Segmentation} {Open-vocabulary camouflaged object segmentation}.{\BBCQ}
\newblock
 \APACrefbtitle {Proceedings of the {E}uropean {C}onference on {C}omputer {V}ision (ECCV).} {Proceedings of the {E}uropean {C}onference on {C}omputer {V}ision (eccv).}
\PrintBackRefs{\CurrentBib}

\bibitem [\protect \citeauthoryear {%
Parmar%
\ \protect \BOthers {.}}{%
Parmar%
\ \protect \BOthers {.}}{%
{\protect \APACyear {2023}}%
}]{%
parmar2023zero}
\APACinsertmetastar {%
parmar2023zero}%
\begin{APACrefauthors}%
Parmar, G.%
, Kumar~Singh, K.%
, Zhang, R.%
, Li, Y.%
, Lu, J.%
\BCBL {} Zhu, J\BHBI Y.%
\end{APACrefauthors}%
\unskip\
\newblock
\APACrefYearMonthDay{2023}{}{}.
\newblock
{\BBOQ}\APACrefatitle {Zero-Shot Image-to-Image Translation} {Zero-shot image-to-image translation}.{\BBCQ}
\newblock
 \APACrefbtitle {Proceedings of the {ACM SIGGRAPH}} {Proceedings of the {ACM SIGGRAPH}}\ (\BPGS\ 1--11).
\PrintBackRefs{\CurrentBib}

\bibitem [\protect \citeauthoryear {%
Pei%
\ \protect \BOthers {.}}{%
Pei%
\ \protect \BOthers {.}}{%
{\protect \APACyear {2022}}%
}]{%
pei2022osformer}
\APACinsertmetastar {%
pei2022osformer}%
\begin{APACrefauthors}%
Pei, J.%
, Cheng, T.%
, Fan, D\BHBI P.%
, Tang, H.%
, Chen, C.%
\BCBL {} Van~Gool, L.%
\end{APACrefauthors}%
\unskip\
\newblock
\APACrefYearMonthDay{2022}{}{}.
\newblock
{\BBOQ}\APACrefatitle {OSFormer: One-Stage Camouflaged Instance Segmentation with Transformers} {Osformer: One-stage camouflaged instance segmentation with transformers}.{\BBCQ}
\newblock
 \APACrefbtitle {Proceedings of the {E}uropean {C}onference on {C}omputer {V}ision} {Proceedings of the {E}uropean {C}onference on {C}omputer {V}ision}\ (\BPGS\ 19--37).
\PrintBackRefs{\CurrentBib}

\bibitem [\protect \citeauthoryear {%
Radford%
\ \protect \BOthers {.}}{%
Radford%
\ \protect \BOthers {.}}{%
{\protect \APACyear {2021}}%
}]{%
CLIP}
\APACinsertmetastar {%
CLIP}%
\begin{APACrefauthors}%
Radford, A.%
, Kim, J.W.%
, Hallacy, C.%
, Ramesh, A.%
, Goh, G.%
, Agarwal, S.%
\BDBL {}Sutskever, I.%
\end{APACrefauthors}%
\unskip\
\newblock
\APACrefYearMonthDay{2021}{}{}.
\newblock
{\BBOQ}\APACrefatitle {Learning Transferable Visual Models From Natural Language Supervision} {Learning transferable visual models from natural language supervision}.{\BBCQ}
\newblock
 \APACrefbtitle {Proceedings of the {I}nternational {C}onference on {M}achine {L}earning} {Proceedings of the {I}nternational {C}onference on {M}achine {L}earning}\ (\BPGS\ 8748--8763).
\PrintBackRefs{\CurrentBib}

\bibitem [\protect \citeauthoryear {%
Raffel%
\ \protect \BOthers {.}}{%
Raffel%
\ \protect \BOthers {.}}{%
{\protect \APACyear {2020}}%
}]{%
colin2022t5}
\APACinsertmetastar {%
colin2022t5}%
\begin{APACrefauthors}%
Raffel, C.%
, Shazeer, N.%
, Roberts, A.%
, Lee, K.%
, Narang, S.%
, Matena, M.%
\BDBL {}Liu, P.J.%
\end{APACrefauthors}%
\unskip\
\newblock
\APACrefYearMonthDay{2020}{}{}.
\newblock
{\BBOQ}\APACrefatitle {Exploring the Limits of Transfer Learning with a Unified Text-to-Text Transformer} {Exploring the limits of transfer learning with a unified text-to-text transformer}.{\BBCQ}
\newblock
\APACjournalVolNumPages{Journal of {M}achine {L}earning {R}esearch}{21}{}{1--67,}
\newblock

\newblock

\PrintBackRefs{\CurrentBib}

\bibitem [\protect \citeauthoryear {%
Ramesh%
\ \protect \BOthers {.}}{%
Ramesh%
\ \protect \BOthers {.}}{%
{\protect \APACyear {2022}}%
}]{%
ramesh2022hierarchical}
\APACinsertmetastar {%
ramesh2022hierarchical}%
\begin{APACrefauthors}%
Ramesh, A.%
, Dhariwal, P.%
, Nichol, A.%
, Chu, C.%
\BCBL {} Chen, M.%
\end{APACrefauthors}%
\unskip\
\newblock
\APACrefYearMonthDay{2022}{}{}.
\newblock
{\BBOQ}\APACrefatitle {Hierarchical text-conditional image generation with clip latents} {Hierarchical text-conditional image generation with clip latents}.{\BBCQ}
\newblock
\APACjournalVolNumPages{ar{X}iv preprint ar{X}iv:2204.06125}{}{}{1--27,}
\newblock

\newblock

\PrintBackRefs{\CurrentBib}

\bibitem [\protect \citeauthoryear {%
Rasheed%
\ \protect \BOthers {.}}{%
Rasheed%
\ \protect \BOthers {.}}{%
{\protect \APACyear {2022}}%
}]{%
rasheed2022bridging}
\APACinsertmetastar {%
rasheed2022bridging}%
\begin{APACrefauthors}%
Rasheed, H.A.%
, Maaz, M.%
, Khattak, M.U.%
, Khan, S.H.%
\BCBL {} Khan, F.S.%
\end{APACrefauthors}%
\unskip\
\newblock
\APACrefYearMonthDay{2022}{}{}.
\newblock
{\BBOQ}\APACrefatitle {Bridging the Gap between Object and Image-level Representations for Open-Vocabulary Detection} {Bridging the gap between object and image-level representations for open-vocabulary detection}.{\BBCQ}
\newblock
 \APACrefbtitle {Proceedings of the {A}dvances in {N}eural {I}nformation {P}rocessing {S}ystems} {Proceedings of the {A}dvances in {N}eural {I}nformation {P}rocessing {S}ystems}\ (\BPGS\ 33781--33794).
\PrintBackRefs{\CurrentBib}

\bibitem [\protect \citeauthoryear {%
Ren%
\ \protect \BOthers {.}}{%
Ren%
\ \protect \BOthers {.}}{%
{\protect \APACyear {2015}}%
}]{%
FasterRCNN}
\APACinsertmetastar {%
FasterRCNN}%
\begin{APACrefauthors}%
Ren, S.%
, He, K.%
, Girshick, R.B.%
\BCBL {} Sun, J.%
\end{APACrefauthors}%
\unskip\
\newblock
\APACrefYearMonthDay{2015}{}{}.
\newblock
{\BBOQ}\APACrefatitle {Faster {R-CNN:} Towards Real-Time Object Detection with Region Proposal Networks} {Faster {R-CNN:} towards real-time object detection with region proposal networks}.{\BBCQ}
\newblock
 \APACrefbtitle {{A}dvances in {N}eural {I}nformation {P}rocessing {S}ystems} {{A}dvances in {N}eural {I}nformation {P}rocessing {S}ystems}\ (\BPGS\ 91--99).
\PrintBackRefs{\CurrentBib}

\bibitem [\protect \citeauthoryear {%
Rewatbowornwong%
\ \protect \BOthers {.}}{%
Rewatbowornwong%
\ \protect \BOthers {.}}{%
{\protect \APACyear {2023}}%
}]{%
tritrong2022repurposing}
\APACinsertmetastar {%
tritrong2022repurposing}%
\begin{APACrefauthors}%
Rewatbowornwong, P.%
, Tritrong, N.%
\BCBL {} Suwajanakorn, S.%
\end{APACrefauthors}%
\unskip\
\newblock
\APACrefYearMonthDay{2023}{}{}.
\newblock
{\BBOQ}\APACrefatitle {Repurposing GANs for One-Shot Semantic Part Segmentation} {Repurposing gans for one-shot semantic part segmentation}.{\BBCQ}
\newblock
\APACjournalVolNumPages{{IEEE} {T}ransactions on {P}attern {A}nalysis and {M}achine {I}ntelligence}{45}{4}{5114--5125,}
\newblock

\newblock

\PrintBackRefs{\CurrentBib}

\bibitem [\protect \citeauthoryear {%
Robin%
\ \protect \BOthers {.}}{%
Robin%
\ \protect \BOthers {.}}{%
{\protect \APACyear {2022}}%
}]{%
rombach2022high}
\APACinsertmetastar {%
rombach2022high}%
\begin{APACrefauthors}%
Robin, R.%
, Andreas, B.%
, Dominik, L.%
, Patrick, E.%
\BCBL {} Bj{\"o}rn, O.%
\end{APACrefauthors}%
\unskip\
\newblock
\APACrefYearMonthDay{2022}{}{}.
\newblock
{\BBOQ}\APACrefatitle {High-resolution image synthesis with latent diffusion models} {High-resolution image synthesis with latent diffusion models}.{\BBCQ}
\newblock
 \APACrefbtitle {Proceedings of the {IEEE/CVF} {C}onference on {C}omputer {V}ision and {P}attern {R}ecognition} {Proceedings of the {IEEE/CVF} {C}onference on {C}omputer {V}ision and {P}attern {R}ecognition}\ (\BPGS\ 10674--10685).
\PrintBackRefs{\CurrentBib}

\bibitem [\protect \citeauthoryear {%
Saharia%
\ \protect \BOthers {.}}{%
Saharia%
\ \protect \BOthers {.}}{%
{\protect \APACyear {2022}}%
}]{%
saharia2022photorealistic}
\APACinsertmetastar {%
saharia2022photorealistic}%
\begin{APACrefauthors}%
Saharia, C.%
, Chan, W.%
, Saxena, S.%
, Li, L.%
, Whang, J.%
, Denton, E.%
\BDBL {}others%
\end{APACrefauthors}%
\unskip\
\newblock
\APACrefYearMonthDay{2022}{}{}.
\newblock
{\BBOQ}\APACrefatitle {Photorealistic text-to-image diffusion models with deep language understanding} {Photorealistic text-to-image diffusion models with deep language understanding}.{\BBCQ}
\newblock
 \APACrefbtitle {Proceedings of the {A}dvances in {N}eural {I}nformation {P}rocessing {S}ystems} {Proceedings of the {A}dvances in {N}eural {I}nformation {P}rocessing {S}ystems}\ (\BPGS\ 36479--36494).
\PrintBackRefs{\CurrentBib}

\bibitem [\protect \citeauthoryear {%
Sariyildiz%
\ \protect \BOthers {.}}{%
Sariyildiz%
\ \protect \BOthers {.}}{%
{\protect \APACyear {2020}}%
}]{%
Sariyildiz_ECCV_2020}
\APACinsertmetastar {%
Sariyildiz_ECCV_2020}%
\begin{APACrefauthors}%
Sariyildiz, M.B.%
, Perez, J.%
\BCBL {} Larlus, D.%
\end{APACrefauthors}%
\unskip\
\newblock
\APACrefYearMonthDay{2020}{}{}.
\newblock
{\BBOQ}\APACrefatitle {Learning Visual Representations with Caption Annotations} {Learning visual representations with caption annotations}.{\BBCQ}
\newblock
 \APACrefbtitle {Proceedings of the {E}uropean {C}onference on {C}omputer {V}ision (ECCV)} {Proceedings of the {E}uropean {C}onference on {C}omputer {V}ision (eccv)}\ (\BPGS\ 1--17).
\PrintBackRefs{\CurrentBib}

\bibitem [\protect \citeauthoryear {%
Schuhmann%
\ \protect \BOthers {.}}{%
Schuhmann%
\ \protect \BOthers {.}}{%
{\protect \APACyear {2022}}%
}]{%
schuhmann2022laion}
\APACinsertmetastar {%
schuhmann2022laion}%
\begin{APACrefauthors}%
Schuhmann, C.%
, Beaumont, R.%
, Vencu, R.%
, Gordon, C.%
, Wightman, R.%
, Cherti, M.%
\BDBL {}others%
\end{APACrefauthors}%
\unskip\
\newblock
\APACrefYearMonthDay{2022}{}{}.
\newblock
{\BBOQ}\APACrefatitle {{LAION}-5B: An open large-scale dataset for training next generation image-text models} {{LAION}-5b: An open large-scale dataset for training next generation image-text models}.{\BBCQ}
\newblock
\APACjournalVolNumPages{{A}dvances in {N}eural {I}nformation {P}rocessing {S}ystems}{}{}{25278--25294,}
\newblock

\newblock

\PrintBackRefs{\CurrentBib}

\bibitem [\protect \citeauthoryear {%
Sim{\~o}es%
\ \protect \BOthers {.}}{%
Sim{\~o}es%
\ \protect \BOthers {.}}{%
{\protect \APACyear {2023}}%
}]{%
simoes2023deepwild}
\APACinsertmetastar {%
simoes2023deepwild}%
\begin{APACrefauthors}%
Sim{\~o}es, F.%
, Bouveyron, C.%
\BCBL {} Precioso, F.%
\end{APACrefauthors}%
\unskip\
\newblock
\APACrefYearMonthDay{2023}{}{}.
\newblock
{\BBOQ}\APACrefatitle {DeepWILD: Wildlife Identification, Localisation and estimation on camera trap videos using Deep learning} {Deepwild: Wildlife identification, localisation and estimation on camera trap videos using deep learning}.{\BBCQ}
\newblock
\APACjournalVolNumPages{Ecological Informatics}{75}{}{102095,}
\newblock

\newblock

\PrintBackRefs{\CurrentBib}

\bibitem [\protect \citeauthoryear {%
J.~Song%
\ \protect \BOthers {.}}{%
J.~Song%
\ \protect \BOthers {.}}{%
{\protect \APACyear {2021}}%
}]{%
song2021denoising}
\APACinsertmetastar {%
song2021denoising}%
\begin{APACrefauthors}%
Song, J.%
, Meng, C.%
\BCBL {} Ermon, S.%
\end{APACrefauthors}%
\unskip\
\newblock
\APACrefYearMonthDay{2021}{}{}.
\newblock
{\BBOQ}\APACrefatitle {Denoising Diffusion Implicit Models} {Denoising diffusion implicit models}.{\BBCQ}
\newblock
 \APACrefbtitle {Proceedings of the {I}nternational {C}onference on {L}earning {R}epresentations.} {Proceedings of the {I}nternational {C}onference on {L}earning {R}epresentations.}
\PrintBackRefs{\CurrentBib}

\bibitem [\protect \citeauthoryear {%
Z.~Song%
\ \protect \BOthers {.}}{%
Z.~Song%
\ \protect \BOthers {.}}{%
{\protect \APACyear {2023}}%
}]{%
song2023pixel}
\APACinsertmetastar {%
song2023pixel}%
\begin{APACrefauthors}%
Song, Z.%
, Kang, X.%
, Wei, X.%
\BCBL {} Li, S.%
\end{APACrefauthors}%
\unskip\
\newblock
\APACrefYearMonthDay{2023}{}{}.
\newblock
{\BBOQ}\APACrefatitle {Pixel-centric context perception network for camouflaged object detection} {Pixel-centric context perception network for camouflaged object detection}.{\BBCQ}
\newblock
\APACjournalVolNumPages{IEEE Transactions on Neural Networks and Learning Systems}{}{}{,}
\newblock

\newblock

\PrintBackRefs{\CurrentBib}

\bibitem [\protect \citeauthoryear {%
G.~Sun%
\ \protect \BOthers {.}}{%
G.~Sun%
\ \protect \BOthers {.}}{%
{\protect \APACyear {2023}}%
}]{%
sun2023ioc}
\APACinsertmetastar {%
sun2023ioc}%
\begin{APACrefauthors}%
Sun, G.%
, An, Z.%
, Liu, Y.%
, Liu, C.%
, Sakaridis, C.%
, Fan, D\BHBI P.%
\BCBL {} Van~Gool, L.%
\end{APACrefauthors}%
\unskip\
\newblock
\APACrefYearMonthDay{2023}{}{}.
\newblock
{\BBOQ}\APACrefatitle {Indiscernible Object Counting in Underwater Scenes} {Indiscernible object counting in underwater scenes}.{\BBCQ}
\newblock
 \APACrefbtitle {Proceedings of the {IEEE/CVF} {C}onference on {C}omputer {V}ision and {P}attern {R}ecognition} {Proceedings of the {IEEE/CVF} {C}onference on {C}omputer {V}ision and {P}attern {R}ecognition}\ (\BPGS\ 13791--13801).
\PrintBackRefs{\CurrentBib}

\bibitem [\protect \citeauthoryear {%
Y.~Sun%
\ \protect \BOthers {.}}{%
Y.~Sun%
\ \protect \BOthers {.}}{%
{\protect \APACyear {2021}}%
}]{%
sun2021c2fnet}
\APACinsertmetastar {%
sun2021c2fnet}%
\begin{APACrefauthors}%
Sun, Y.%
, Chen, G.%
, Zhou, T.%
, Zhang, Y.%
\BCBL {} Liu, N.%
\end{APACrefauthors}%
\unskip\
\newblock
\APACrefYearMonthDay{2021}{}{}.
\newblock
{\BBOQ}\APACrefatitle {Context-aware Cross-level Fusion Network for Camouflaged Object Detection} {Context-aware cross-level fusion network for camouflaged object detection}.{\BBCQ}
\newblock
 \APACrefbtitle {IJCAI} {Ijcai}\ (\BPGS\ 1025--1031).
\PrintBackRefs{\CurrentBib}

\bibitem [\protect \citeauthoryear {%
Tian%
\ \protect \BOthers {.}}{%
Tian%
\ \protect \BOthers {.}}{%
{\protect \APACyear {2020}}%
}]{%
condinst}
\APACinsertmetastar {%
condinst}%
\begin{APACrefauthors}%
Tian, Z.%
, Shen, C.%
\BCBL {} Chen, H.%
\end{APACrefauthors}%
\unskip\
\newblock
\APACrefYearMonthDay{2020}{}{}.
\newblock
{\BBOQ}\APACrefatitle {Conditional convolutions for instance segmentation} {Conditional convolutions for instance segmentation}.{\BBCQ}
\newblock
 \APACrefbtitle {Proceedings of the {E}uropean {C}onference on {C}omputer {V}ision} {Proceedings of the {E}uropean {C}onference on {C}omputer {V}ision}\ (\BPGS\ 282--298).
\PrintBackRefs{\CurrentBib}

\bibitem [\protect \citeauthoryear {%
Troscianko%
\ \protect \BOthers {.}}{%
Troscianko%
\ \protect \BOthers {.}}{%
{\protect \APACyear {2017}}%
}]{%
troscianko2017quantifying}
\APACinsertmetastar {%
troscianko2017quantifying}%
\begin{APACrefauthors}%
Troscianko, J.%
, Skelhorn, J.%
\BCBL {} Stevens, M.%
\end{APACrefauthors}%
\unskip\
\newblock
\APACrefYearMonthDay{2017}{}{}.
\newblock
{\BBOQ}\APACrefatitle {Quantifying camouflage: how to predict detectability from appearance} {Quantifying camouflage: how to predict detectability from appearance}.{\BBCQ}
\newblock
\APACjournalVolNumPages{BMC Evolutionary Biology}{17}{}{1--13,}
\newblock

\newblock

\PrintBackRefs{\CurrentBib}

\bibitem [\protect \citeauthoryear {%
H.~Wang%
\ \protect \BOthers {.}}{%
H.~Wang%
\ \protect \BOthers {.}}{%
{\protect \APACyear {2024}}%
}]{%
Wang_computer_2024}
\APACinsertmetastar {%
Wang_computer_2024}%
\begin{APACrefauthors}%
Wang, H.%
, Hu, T.%
, Zhang, Y.%
, Zhang, H.%
, Qi, Y.%
, Wang, L.%
\BDBL {}Du, M.%
\end{APACrefauthors}%
\unskip\
\newblock
\APACrefYearMonthDay{2024}{}{}.
\newblock
{\BBOQ}\APACrefatitle {Unveiling camouflaged and partially occluded colorectal polyps: Introducing {CPSNet} for accurate colon polyp segmentation} {Unveiling camouflaged and partially occluded colorectal polyps: Introducing {CPSNet} for accurate colon polyp segmentation}.{\BBCQ}
\newblock
\APACjournalVolNumPages{Computers in Biology and Medicine}{171}{}{108186,}
\newblock

\newblock

\PrintBackRefs{\CurrentBib}

\bibitem [\protect \citeauthoryear {%
X.~Wang%
\ \protect \BOthers {.}}{%
X.~Wang%
\ \protect \BOthers {.}}{%
{\protect \APACyear {2020}}%
}]{%
wang2020solov2}
\APACinsertmetastar {%
wang2020solov2}%
\begin{APACrefauthors}%
Wang, X.%
, Zhang, R.%
, Kong, T.%
, Li, L.%
\BCBL {} Shen, C.%
\end{APACrefauthors}%
\unskip\
\newblock
\APACrefYearMonthDay{2020}{}{}.
\newblock
{\BBOQ}\APACrefatitle {Solov2: Dynamic and fast instance segmentation} {Solov2: Dynamic and fast instance segmentation}.{\BBCQ}
\newblock
\APACjournalVolNumPages{{A}dvances in {N}eural {I}nformation {P}rocessing {S}ystems}{33}{}{17721--17732,}
\newblock

\newblock

\PrintBackRefs{\CurrentBib}

\bibitem [\protect \citeauthoryear {%
Wen%
\ \protect \BOthers {.}}{%
Wen%
\ \protect \BOthers {.}}{%
{\protect \APACyear {2024}}%
}]{%
app14062494}
\APACinsertmetastar {%
app14062494}%
\begin{APACrefauthors}%
Wen, Y.%
, Ke, W.%
\BCBL {} Sheng, H.%
\end{APACrefauthors}%
\unskip\
\newblock
\APACrefYearMonthDay{2024}{}{}.
\newblock
{\BBOQ}\APACrefatitle {Camouflaged Object Detection Based on Deep Learning with Attention-Guided Edge Detection and Multi-Scale Context Fusion} {Camouflaged object detection based on deep learning with attention-guided edge detection and multi-scale context fusion}.{\BBCQ}
\newblock
\APACjournalVolNumPages{Applied Sciences}{}{}{,}
\newblock
\begin{APACrefDOI} \doi{10.3390/app14062494} \end{APACrefDOI}
\newblock

\newblock

\PrintBackRefs{\CurrentBib}

\bibitem [\protect \citeauthoryear {%
J.~Wu%
\ \protect \BOthers {.}}{%
J.~Wu%
\ \protect \BOthers {.}}{%
{\protect \APACyear {2024}}%
}]{%
OVSurvey}
\APACinsertmetastar {%
OVSurvey}%
\begin{APACrefauthors}%
Wu, J.%
, Li, X.%
, Xu, S.%
, Yuan, H.%
, Ding, H.%
, Yang, Y.%
\BDBL {}Tao, D.%
\end{APACrefauthors}%
\unskip\
\newblock
\APACrefYearMonthDay{2024}{jul}{}.
\newblock
{\BBOQ}\APACrefatitle {Towards Open Vocabulary Learning: A Survey} {Towards open vocabulary learning: A survey}.{\BBCQ}
\newblock
\APACjournalVolNumPages{IEEE Transactions on Pattern Analysis and Machine Intelligence}{46}{07}{5092--5113,}
\newblock

\newblock

\PrintBackRefs{\CurrentBib}

\bibitem [\protect \citeauthoryear {%
Y.~Wu%
\ \protect \BOthers {.}}{%
Y.~Wu%
\ \protect \BOthers {.}}{%
{\protect \APACyear {2019}}%
}]{%
wu2019detectron2}
\APACinsertmetastar {%
wu2019detectron2}%
\begin{APACrefauthors}%
Wu, Y.%
, Kirillov, A.%
, Massa, F.%
, Lo, W\BHBI Y.%
\BCBL {} Girshick, R.%
\end{APACrefauthors}%
\unskip\
\newblock
\APACrefYearMonthDay{2019}{}{}.
\newblock
\APACrefbtitle {Detectron2.} {Detectron2.}
\newblock
\APAChowpublished {\url{https://github.com/facebookresearch/detectron2}}.
\PrintBackRefs{\CurrentBib}

\bibitem [\protect \citeauthoryear {%
Xiao%
\ \protect \BOthers {.}}{%
Xiao%
\ \protect \BOthers {.}}{%
{\protect \APACyear {2023}}%
}]{%
BCNet}
\APACinsertmetastar {%
BCNet}%
\begin{APACrefauthors}%
Xiao, J.%
, Chen, T.%
, Hu, X.%
, Zhang, G.%
\BCBL {} Wang, S.%
\end{APACrefauthors}%
\unskip\
\newblock
\APACrefYearMonthDay{2023}{}{}.
\newblock
{\BBOQ}\APACrefatitle {Boundary-guided context-aware network for camouflaged object detection} {Boundary-guided context-aware network for camouflaged object detection}.{\BBCQ}
\newblock
\APACjournalVolNumPages{Neural Computing and Applications}{}{}{,}
\newblock

\newblock

\PrintBackRefs{\CurrentBib}

\bibitem [\protect \citeauthoryear {%
Xie%
\ \protect \BOthers {.}}{%
Xie%
\ \protect \BOthers {.}}{%
{\protect \APACyear {2021}}%
}]{%
xie2021trans}
\APACinsertmetastar {%
xie2021trans}%
\begin{APACrefauthors}%
Xie, E.%
, Wang, W.%
, Wang, W.%
, Sun, P.%
, Xu, H.%
, Liang, D.%
\BCBL {} Luo, P.%
\end{APACrefauthors}%
\unskip\
\newblock
\APACrefYearMonthDay{2021}{}{}.
\newblock
{\BBOQ}\APACrefatitle {Segmenting Transparent Objects in the Wild with Transformer} {Segmenting transparent objects in the wild with transformer}.{\BBCQ}
\newblock
 \APACrefbtitle {Proceedings of the {International Joint Conferences on Artificial Intelligence}} {Proceedings of the {International Joint Conferences on Artificial Intelligence}}\ (\BPGS\ 1194--1200).
\PrintBackRefs{\CurrentBib}

\bibitem [\protect \citeauthoryear {%
J.~Xu%
\ \protect \BOthers {.}}{%
J.~Xu%
\ \protect \BOthers {.}}{%
{\protect \APACyear {2023}}%
}]{%
xu2023odise}
\APACinsertmetastar {%
xu2023odise}%
\begin{APACrefauthors}%
Xu, J.%
, Liu, S.%
, Vahdat, A.%
, Byeon, W.%
, Wang, X.%
\BCBL {} Mello, S.D.%
\end{APACrefauthors}%
\unskip\
\newblock
\APACrefYearMonthDay{2023}{}{}.
\newblock
{\BBOQ}\APACrefatitle {Open-Vocabulary Panoptic Segmentation with Text-to-Image Diffusion Models} {Open-vocabulary panoptic segmentation with text-to-image diffusion models}.{\BBCQ}
\newblock
 \APACrefbtitle {Proceedings of the {IEEE/CVF} {C}onference on {C}omputer {V}ision and {P}attern {R}ecognition} {Proceedings of the {IEEE/CVF} {C}onference on {C}omputer {V}ision and {P}attern {R}ecognition}\ (\BPGS\ 2955--2966).
\PrintBackRefs{\CurrentBib}

\bibitem [\protect \citeauthoryear {%
X.~Xu%
\ \protect \BOthers {.}}{%
X.~Xu%
\ \protect \BOthers {.}}{%
{\protect \APACyear {2023}}%
}]{%
xu2023masqclip}
\APACinsertmetastar {%
xu2023masqclip}%
\begin{APACrefauthors}%
Xu, X.%
, Xiong, T.%
, Ding, Z.%
\BCBL {} Tu, Z.%
\end{APACrefauthors}%
\unskip\
\newblock
\APACrefYearMonthDay{2023}{October}{}.
\newblock
{\BBOQ}\APACrefatitle {MasQCLIP for Open-Vocabulary Universal Image Segmentation} {Masqclip for open-vocabulary universal image segmentation}.{\BBCQ}
\newblock
 \APACrefbtitle {Proceedings of the IEEE/CVF International Conference on Computer Vision (ICCV)} {Proceedings of the ieee/cvf international conference on computer vision (iccv)}\ (\BPGS\ 887--898).
\PrintBackRefs{\CurrentBib}

\bibitem [\protect \citeauthoryear {%
Yan%
\ \protect \BOthers {.}}{%
Yan%
\ \protect \BOthers {.}}{%
{\protect \APACyear {2021}}%
}]{%
9371667}
\APACinsertmetastar {%
9371667}%
\begin{APACrefauthors}%
Yan, J.%
, Le, T.%
, Nguyen, K.%
, Tran, M.%
, Do, T.%
\BCBL {} Nguyen, T.V.%
\end{APACrefauthors}%
\unskip\
\newblock
\APACrefYearMonthDay{2021}{}{}.
\newblock
{\BBOQ}\APACrefatitle {MirrorNet: Bio-Inspired Camouflaged Object Segmentation} {Mirrornet: Bio-inspired camouflaged object segmentation}.{\BBCQ}
\newblock
\APACjournalVolNumPages{{IEEE} {A}ccess}{9}{}{43290--43300,}
\newblock

\newblock

\PrintBackRefs{\CurrentBib}

\bibitem [\protect \citeauthoryear {%
Zang%
\ \protect \BOthers {.}}{%
Zang%
\ \protect \BOthers {.}}{%
{\protect \APACyear {2022}}%
}]{%
OV-DETR}
\APACinsertmetastar {%
OV-DETR}%
\begin{APACrefauthors}%
Zang, Y.%
, Li, W.%
, Zhou, K.%
, Huang, C.%
\BCBL {} Loy, C.C.%
\end{APACrefauthors}%
\unskip\
\newblock
\APACrefYearMonthDay{2022}{}{}.
\newblock
{\BBOQ}\APACrefatitle {Open-Vocabulary DETR with Conditional Matching} {Open-vocabulary detr with conditional matching}.{\BBCQ}
\newblock
 \APACrefbtitle {Proceedings of the {E}uropean {C}onference on {C}omputer {V}ision} {Proceedings of the {E}uropean {C}onference on {C}omputer {V}ision}\ (\BPGS\ 106--122).
\PrintBackRefs{\CurrentBib}

\bibitem [\protect \citeauthoryear {%
Zareian%
\ \protect \BOthers {.}}{%
Zareian%
\ \protect \BOthers {.}}{%
{\protect \APACyear {2021}}%
}]{%
zareian2021open}
\APACinsertmetastar {%
zareian2021open}%
\begin{APACrefauthors}%
Zareian, A.%
, Rosa, K.D.%
, Hu, D.H.%
\BCBL {} Chang, S.%
\end{APACrefauthors}%
\unskip\
\newblock
\APACrefYearMonthDay{2021}{}{}.
\newblock
{\BBOQ}\APACrefatitle {Open-vocabulary object detection using captions} {Open-vocabulary object detection using captions}.{\BBCQ}
\newblock
 \APACrefbtitle {Proceedings of the {IEEE/CVF} {C}onference on {C}omputer {V}ision and {P}attern {R}ecognition} {Proceedings of the {IEEE/CVF} {C}onference on {C}omputer {V}ision and {P}attern {R}ecognition}\ (\BPGS\ 14393--14402).
\PrintBackRefs{\CurrentBib}

\bibitem [\protect \citeauthoryear {%
H.~Zhang%
\ \protect \BOthers {.}}{%
H.~Zhang%
\ \protect \BOthers {.}}{%
{\protect \APACyear {2023}}%
}]{%
OpenSeeD}
\APACinsertmetastar {%
OpenSeeD}%
\begin{APACrefauthors}%
Zhang, H.%
, Li, F.%
, Zou, X.%
, Liu, S.%
, Li, C.%
, Yang, J.%
\BCBL {} Zhang, L.%
\end{APACrefauthors}%
\unskip\
\newblock
\APACrefYearMonthDay{2023}{October}{}.
\newblock
{\BBOQ}\APACrefatitle {A Simple Framework for Open-Vocabulary Segmentation and Detection} {A simple framework for open-vocabulary segmentation and detection}.{\BBCQ}
\newblock
 \APACrefbtitle {Proceedings of the {IEEE/CVF} {I}nternational {C}onference on {C}omputer {V}ision} {Proceedings of the {IEEE/CVF} {I}nternational {C}onference on {C}omputer {V}ision}\ (\BPGS\ 1020--1031).
\PrintBackRefs{\CurrentBib}

\bibitem [\protect \citeauthoryear {%
J.~Zhang%
\ \protect \BOthers {.}}{%
J.~Zhang%
\ \protect \BOthers {.}}{%
{\protect \APACyear {2023}}%
}]{%
zhang2023vision}
\APACinsertmetastar {%
zhang2023vision}%
\begin{APACrefauthors}%
Zhang, J.%
, Huang, J.%
, Jin, S.%
\BCBL {} Lu, S.%
\end{APACrefauthors}%
\unskip\
\newblock
\APACrefYearMonthDay{2023}{}{}.
\newblock
{\BBOQ}\APACrefatitle {Vision-language models for vision tasks: A survey} {Vision-language models for vision tasks: A survey}.{\BBCQ}
\newblock
\APACjournalVolNumPages{ar{X}iv preprint ar{X}iv:2304.00685}{}{}{1--23,}
\newblock

\newblock

\PrintBackRefs{\CurrentBib}

\bibitem [\protect \citeauthoryear {%
Y.~Zhang%
\ \protect \BOthers {.}}{%
Y.~Zhang%
\ \protect \BOthers {.}}{%
{\protect \APACyear {2022}}%
}]{%
Zhang_PMLR_2022}
\APACinsertmetastar {%
Zhang_PMLR_2022}%
\begin{APACrefauthors}%
Zhang, Y.%
, Jiang, H.%
, Miura, Y.%
, Manning, C.D.%
\BCBL {} Langlotz, C.P.%
\end{APACrefauthors}%
\unskip\
\newblock
\APACrefYearMonthDay{2022}{}{}.
\newblock
{\BBOQ}\APACrefatitle {Contrastive Learning of Medical Visual Representations from Paired Images and Text} {Contrastive learning of medical visual representations from paired images and text}.{\BBCQ}
\newblock
\APACjournalVolNumPages{Proceedings of Machine Learning Research}{182}{}{1--24,}
\newblock

\newblock

\PrintBackRefs{\CurrentBib}

\bibitem [\protect \citeauthoryear {%
Zhao%
\ \protect \BOthers {.}}{%
Zhao%
\ \protect \BOthers {.}}{%
{\protect \APACyear {2023}}%
}]{%
vpd}
\APACinsertmetastar {%
vpd}%
\begin{APACrefauthors}%
Zhao, W.%
, Rao, Y.%
, Liu, Z.%
, Liu, B.%
, Zhou, J.%
\BCBL {} Lu, J.%
\end{APACrefauthors}%
\unskip\
\newblock
\APACrefYearMonthDay{2023}{}{}.
\newblock
{\BBOQ}\APACrefatitle {Unleashing Text-to-Image Diffusion Models for Visual Perception} {Unleashing text-to-image diffusion models for visual perception}.{\BBCQ}
\newblock
 \APACrefbtitle {Proceedings of the {IEEE/CVF} {I}nternational {C}onference on {C}omputer {V}ision} {Proceedings of the {IEEE/CVF} {I}nternational {C}onference on {C}omputer {V}ision}\ (\BPGS\ 5729--5739).
\PrintBackRefs{\CurrentBib}

\bibitem [\protect \citeauthoryear {%
Zheng%
\ \protect \BOthers {.}}{%
Zheng%
\ \protect \BOthers {.}}{%
{\protect \APACyear {2021}}%
}]{%
zheng2021zero}
\APACinsertmetastar {%
zheng2021zero}%
\begin{APACrefauthors}%
Zheng, Y.%
, Wu, J.%
, Qin, Y.%
, Zhang, F.%
\BCBL {} Cui, L.%
\end{APACrefauthors}%
\unskip\
\newblock
\APACrefYearMonthDay{2021}{}{}.
\newblock
{\BBOQ}\APACrefatitle {Zero-Shot Instance Segmentation} {Zero-shot instance segmentation}.{\BBCQ}
\newblock
 \APACrefbtitle {Proceedings of the {IEEE/CVF} {C}onference on {C}omputer {V}ision and {P}attern {R}ecognition} {Proceedings of the {IEEE/CVF} {C}onference on {C}omputer {V}ision and {P}attern {R}ecognition}\ (\BPGS\ 2593--2602).
\PrintBackRefs{\CurrentBib}

\bibitem [\protect \citeauthoryear {%
Zhong%
\ \protect \BOthers {.}}{%
Zhong%
\ \protect \BOthers {.}}{%
{\protect \APACyear {2022}}%
}]{%
zhong2022regionclip}
\APACinsertmetastar {%
zhong2022regionclip}%
\begin{APACrefauthors}%
Zhong, Y.%
, Yang, J.%
, Zhang, P.%
, Li, C.%
, Codella, N.%
, Li, L.H.%
\BDBL {}Gao, J.%
\end{APACrefauthors}%
\unskip\
\newblock
\APACrefYearMonthDay{2022}{}{}.
\newblock
{\BBOQ}\APACrefatitle {RegionCLIP: Region-based Language-Image Pretraining} {Regionclip: Region-based language-image pretraining}.{\BBCQ}
\newblock
 \APACrefbtitle {Proceedings of the {IEEE/CVF} {C}onference on {C}omputer {V}ision and {P}attern {R}ecognition} {Proceedings of the {IEEE/CVF} {C}onference on {C}omputer {V}ision and {P}attern {R}ecognition}\ (\BPGS\ 16772--16782).
\PrintBackRefs{\CurrentBib}

\bibitem [\protect \citeauthoryear {%
B.~Zhou%
\ \protect \BOthers {.}}{%
B.~Zhou%
\ \protect \BOthers {.}}{%
{\protect \APACyear {2017}}%
}]{%
ade20k-short}
\APACinsertmetastar {%
ade20k-short}%
\begin{APACrefauthors}%
Zhou, B.%
, Zhao, H.%
, Puig, X.%
, Fidler, S.%
, Barriuso, A.%
\BCBL {} Torralba, A.%
\end{APACrefauthors}%
\unskip\
\newblock
\APACrefYearMonthDay{2017}{}{}.
\newblock
{\BBOQ}\APACrefatitle {Scene Parsing through ADE20K Dataset} {Scene parsing through ade20k dataset}.{\BBCQ}
\newblock
 \APACrefbtitle {Proceedings of the {IEEE/CVF} {C}onference on {C}omputer {V}ision and {P}attern {R}ecognition} {Proceedings of the {IEEE/CVF} {C}onference on {C}omputer {V}ision and {P}attern {R}ecognition}\ (\BPGS\ 5122--5130).
\PrintBackRefs{\CurrentBib}

\bibitem [\protect \citeauthoryear {%
B.~Zhou%
\ \protect \BOthers {.}}{%
B.~Zhou%
\ \protect \BOthers {.}}{%
{\protect \APACyear {2019}}%
}]{%
ade20k}
\APACinsertmetastar {%
ade20k}%
\begin{APACrefauthors}%
Zhou, B.%
, Zhao, H.%
, Puig, X.%
, Xiao, T.%
, Fidler, S.%
, Barriuso, A.%
\BCBL {} Torralba, A.%
\end{APACrefauthors}%
\unskip\
\newblock
\APACrefYearMonthDay{2019}{}{}.
\newblock
{\BBOQ}\APACrefatitle {Semantic Understanding of Scenes Through the ADE20K Dataset} {Semantic understanding of scenes through the ade20k dataset}.{\BBCQ}
\newblock
\APACjournalVolNumPages{{I}nternational {J}ournal of {C}omputer {V}ision}{127}{3}{302--321,}
\newblock

\newblock

\PrintBackRefs{\CurrentBib}

\bibitem [\protect \citeauthoryear {%
H.~Zhou%
\ \protect \BOthers {.}}{%
H.~Zhou%
\ \protect \BOthers {.}}{%
{\protect \APACyear {2025}}%
}]{%
zhou2025rethinking}
\APACinsertmetastar {%
zhou2025rethinking}%
\begin{APACrefauthors}%
Zhou, H.%
, Qi, L.%
, Shen, T.%
, Huang, H.%
, Yang, X.%
, Li, X.%
\BCBL {} Yang, M\BHBI H.%
\end{APACrefauthors}%
\unskip\
\newblock
\APACrefYearMonthDay{2025}{}{}.
\newblock
{\BBOQ}\APACrefatitle {Rethinking Evaluation Metrics of Open-Vocabulary Segmentation} {Rethinking evaluation metrics of open-vocabulary segmentation}.{\BBCQ}
\newblock
\APACjournalVolNumPages{IEEE Transactions on Pattern Analysis and Machine Intelligence}{}{}{,}
\newblock

\newblock

\PrintBackRefs{\CurrentBib}

\bibitem [\protect \citeauthoryear {%
X.~Zhou%
\ \protect \BOthers {.}}{%
X.~Zhou%
\ \protect \BOthers {.}}{%
{\protect \APACyear {2022}}%
}]{%
detic}
\APACinsertmetastar {%
detic}%
\begin{APACrefauthors}%
Zhou, X.%
, Girdhar, R.%
, Joulin, A.%
, Kr{\"{a}}henb{\"{u}}hl, P.%
\BCBL {} Misra, I.%
\end{APACrefauthors}%
\unskip\
\newblock
\APACrefYearMonthDay{2022}{}{}.
\newblock
{\BBOQ}\APACrefatitle {Detecting Twenty-Thousand Classes Using Image-Level Supervision} {Detecting twenty-thousand classes using image-level supervision}.{\BBCQ}
\newblock
 \APACrefbtitle {Proceedings of the {E}uropean {C}onference on {C}omputer {V}ision} {Proceedings of the {E}uropean {C}onference on {C}omputer {V}ision}\ (\BPGS\ 350--368).
\PrintBackRefs{\CurrentBib}

\bibitem [\protect \citeauthoryear {%
Zou%
, Dou%
\BCBL {}\ \protect \BOthers {.}}{%
Zou%
, Dou%
\BCBL {}\ \protect \BOthers {.}}{%
{\protect \APACyear {2023}}%
}]{%
xDecoder}
\APACinsertmetastar {%
xDecoder}%
\begin{APACrefauthors}%
Zou, X.%
, Dou, Z\BHBI Y.%
, Yang, J.%
, Gan, Z.%
, Li, L.%
, Li, C.%
\BDBL {}Gao, J.%
\end{APACrefauthors}%
\unskip\
\newblock
\APACrefYearMonthDay{2023}{June}{}.
\newblock
{\BBOQ}\APACrefatitle {Generalized Decoding for Pixel, Image, and Language} {Generalized decoding for pixel, image, and language}.{\BBCQ}
\newblock
 \APACrefbtitle {Proceedings of the {IEEE/CVF} {C}onference on {C}omputer {V}ision and {P}attern {R}ecognition} {Proceedings of the {IEEE/CVF} {C}onference on {C}omputer {V}ision and {P}attern {R}ecognition}\ (\BPGS\ 15116--15127).
\PrintBackRefs{\CurrentBib}

\bibitem [\protect \citeauthoryear {%
Zou%
, Yang%
\BCBL {}\ \protect \BOthers {.}}{%
Zou%
, Yang%
\BCBL {}\ \protect \BOthers {.}}{%
{\protect \APACyear {2023}}%
}]{%
zou2023segment}
\APACinsertmetastar {%
zou2023segment}%
\begin{APACrefauthors}%
Zou, X.%
, Yang, J.%
, Zhang, H.%
, Li, F.%
, Li, L.%
, Wang, J.%
\BDBL {}Lee, Y.J.%
\end{APACrefauthors}%
\unskip\
\newblock
\APACrefYearMonthDay{2023}{}{}.
\newblock
{\BBOQ}\APACrefatitle {Segment Everything Everywhere All at Once} {Segment everything everywhere all at once}.{\BBCQ}
\newblock
 \APACrefbtitle {Thirty-seventh Conference on Neural Information Processing Systems.} {Thirty-seventh conference on neural information processing systems.}
\PrintBackRefs{\CurrentBib}

\end{thebibliography}
%% if required, the content of .bbl file can be included here once bbl is generated
%%\input sn-article.bbl

\end{document}